\documentclass[runningheads]{llncs}

\usepackage{eccv}

\usepackage{eccvabbrv}
\usepackage{graphicx}
\usepackage{booktabs}

\usepackage[accsupp]{axessibility}  

\usepackage[pagebackref,breaklinks,colorlinks,citecolor=eccvblue]{hyperref}

\usepackage{orcidlink}

\begin{document}

\newcommand{\ourwork}{UltraImageGen}
\newcommand{\yuyao}[1]{\textcolor{black}{#1}}
\title{\ourwork:~Efficient Ultra-High-Resolution Image Generation with Hierarchical Local Attention} 

\titlerunning{UltraImageGen}

\author{Yuyao Zhang \orcidlink{0009-0007-2180-3637}
\and
Yu-Wing Tai\orcidlink{0000-0002-3148-0380}}

\authorrunning{Y. Zhang and Y.-W. Tai}

\institute{Dartmouth College, USA \\ \vspace{1em}\textit{\url{https://github.com/PeterYYZhang/UltraImageGen/}}}

\maketitle
\begin{figure}[h]
    \centering
    \vspace{-0.3in}
    \includegraphics[width=\textwidth]{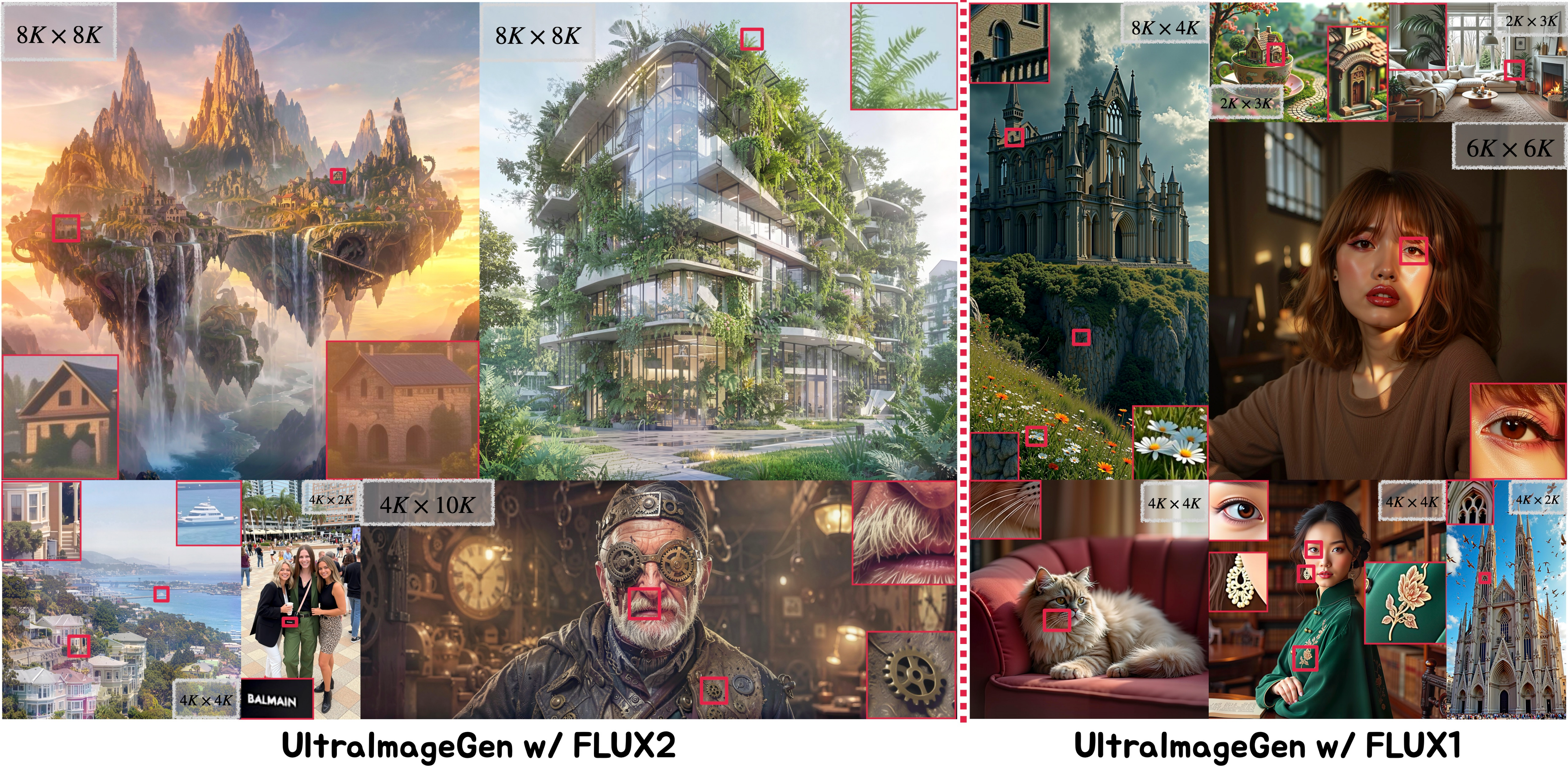}\\
    \caption{Ultra-high-resolution images generated by \ourwork\ with different base models at $8K \times 8K$, $4K \times 10K$, $8K \times 4K$, $6K \times 6K$, $4K \times 4K$, $4K \times 2K$, $2K \times 3K$. Zoomed-in regions highlight the fine-grained details (See supplementary materials for original PDF). Notice that the pretrained FLUX 1 and FLUX 2 are only capable of generating sub 2MP (< $1K\times2K$) and sub 4MP (< $2K\times2K$) images.}
    \label{fig:teaser}
    \vspace{-0.35in}
\end{figure}
\begin{abstract}
Ultra-high-resolution image generation is increasingly vital for applications requiring fine-grained textures and global structural fidelity, yet state-of-the-art text-to-image diffusion models such as FLUX and SD3 remain confined to sub 2MP (< $1K\times2K$) resolutions due to the quadratic complexity of attention mechanisms and the scarcity of high-quality high-resolution training data. We present \textbf{\ourwork}, a novel framework that introduces hierarchical local attention with low-resolution global guidance, enabling efficient, scalable, and semantically coherent image synthesis at ultra-high resolutions. Specifically, high-resolution latents are divided into hardware aligned fixed-size local windows to reduce attention complexity from quadratic to near-linear, while a low-resolution latent with scaled positional embeddings injects global semantics as an anchor. A lightweight LoRA adaptation bridges global and local pathways during denoising, ensuring consistency across structure and detail. To maximize efficiency and achieve scalable ultra-high-resolution generation, we repermute token sequence in window-first order, so that the GPU-friendly dense local blocks in attention calculation equals to the fixed-size local window in 2D regardless of resolution. Together~\ourwork~reliably scales pretrained models to resolutions higher than $8K$ with more than $10\times$ speed up and significantly lower memory usage. Extensive experiments demonstrate that~\ourwork~achieves superior quality while maintaining computational efficiency, establishing a practical paradigm for advancing ultra-high-resolution image generation. Our~\href{https://peteryyzhang.github.io/ERAG/}{project page} and~\href{https://github.com/PeterYYZhang/UltraImageGen/}{code} are available here.
\end{abstract}


\section{Introduction}

Ultra-high-resolution image generation is becoming increasingly crucial for both creative and practical applications. From digital art and advertising to scientific visualization and virtual environments, users demand outputs with fine-grained textures, faithful structures, and seamless details at $4K$ resolution or higher. However, despite the rapid progress of text-to-image (T2I) diffusion models~\cite{ding2021cogview,rombach2022high,chen2023pixart,podell2023sdxl,peebles2023scalable,xue2023raphael,li2024hunyuan,chen2024pixart,li2024playground,blackforestlabs2024flux1dev,cai2025hidream, xie2024sana, xie2025sana}, most state-of-the-art systems remain confined to resolutions below $1K \times 1K$. This limitation fundamentally restricts their usability in scenarios where ultra-high fidelity is paramount. 

The root of this challenge lies in two interdependent bottlenecks. First, scaling model capacity and datasets to higher resolutions requires tremendous resources, as high-quality $2K$–$8K$ training data are scarce and expensive to curate. Second, attention-based diffusion architectures suffer from quadratic growth in token complexity with respect to image resolution, making naive scaling computationally prohibitive. For example, generating a $4K$ image would require tens of thousands of tokens, leading to impractical memory and runtime costs. As a result, existing approaches either resort to costly retraining on synthetic high-resolution datasets~\cite{hoogeboom2023simple,liu2024linfusion,ren2024ultrapixel,teng2023relay,zheng2024any,zhang2025diffusion}, or adopt training-free upscaling methods~\cite{guo2024make,qiu2024freescale,du2024max,liu2024hiprompt,wu2025megafusion,shi2025resmaster,kim2025diffusehigh,huang2024fouriscale,bu2025hiflow} that often compromise efficiency and stability. Consequently, the field still lacks a solution that delivers \emph{both} ultra-high-resolution fidelity and computational scalability. 

In this work, we introduce \textbf{\ourwork}, a novel framework that rethinks ultra-high-resolution synthesis through the lens of efficiency and scalability. Inspired by the way artists construct large-scale murals, we decompose the generation task hierarchically: local windows capture fine textures through lightweight attention, while a low-resolution guidance image preserves global semantic structure. Technically, \ourwork\ introduces three innovations. First, a \emph{hierarchical local attention} mechanism restricts computation to fixed-size windows, reducing quadratic cost to near-linear scaling while yielding over $10GB$ memory savings and more than $10\times$ faster inference compared to dense attention. Second, a \emph{global guidance pathway} employs a low-resolution latent with scaled RoPE positional anchors to maintain long-range dependencies and ensure semantic coherence across windows. Third, a \emph{parameter-efficient joint denoising framework} integrates global and local pathways via LoRA-adapted projections trained only on $256\times256$–$1K\times1K$ data, enabling direct scaling to $4K$ or higher without requiring any native high-resolution training. These design choices translate into two key outcomes: (1) \emph{scalable fidelity}, achieved by hierarchically fusing global composition with local detail to extend beyond pretrained resolutions, and (2) \emph{computational efficiency}, enabled by reducing computation, memory, and runtime while reusing pretrained weights with lightweight adaptation.

Extensive experiments validate these contributions: our work reliably scales to $4K \times 4K$ and higher resolution synthesis with commodity training resolutions, achieving sharper textures, richer details, and stronger global coherence than existing approaches. Quantitatively, it surpasses dense attention baselines with over $10\times$ faster inference and lower memory usage, and it matches or outperforms state-of-the-art methods trained with native 4K data across FID, IS, and CLIP Score. Together, these results show \ourwork\ as a practical and effective solution for advancing ultra-high-resolution text-to-image generation.


We summarize our contributions as follows:
\begin{itemize}
    \item \yuyao{We introduce a new paradigm for ultra-high-resolution image generation, which combines window-first local attention and low-resolution global anchors to achieve resolution agnostic synthesis far beyond pretraining limits, without requiring native high-resolution training data.}
    \item Our framework entirely eliminates the need for high-resolution training data, instead leveraging positionally aligned low-resolution guidance to preserve global coherence. 
    \item We demonstrate substantial efficiency improvements ($>10\times$ speedup in $8K$, less memory usage) while achieving state-of-the-art quality at $4K, 8K$ and higher, offering a practical and generalizable solution for real-world ultra-high-resolution generation.
\end{itemize}

\section{Related Work}
\noindent\textbf{Text-to-Image Generation.} Recent years have witnessed rapid progress in text-to-image (T2I) generation~\cite{ding2021cogview,rombach2022high,chen2023pixart,podell2023sdxl,peebles2023scalable,xue2023raphael,li2024hunyuan,chen2024pixart,li2024playground,blackforestlabs2024flux1dev,cai2025hidream, xie2024sana, xie2025sana, gao2025seedream,flux-2-2025, zhang2025layercraft}, where diffusion models now achieve near-photorealistic synthesis at resolutions up to $1K \times 1K$. The dominant paradigm relies on cross-attention mechanisms and DiT architecture~\cite{peebles2023scalable}, as seen in the PixArt series~\cite{chen2023pixart, chen2024pixart,chen2024pixartsigma}, or multi-modal diffusion transformers (MMDiT) such as FLUX.1.0-dev~\cite{blackforestlabs2024flux1dev}. These architectures have established strong foundations for controllable and high-quality T2I synthesis. However, despite their success, existing methods struggle to scale beyond pretrained resolutions due to quadratic attention costs and the lack of large-scale high-resolution data. Our work builds directly on this line of research, but departs from the prevailing direction by focusing on resolution scalability and computational efficiency rather than retraining larger models.

\vspace{2mm}
\noindent\textbf{High-resolution image synthesis.} Real-world applications increasingly demand resolutions of $4K$ or higher, sparking methods in pushing beyond the $1K \times 1K$ barrier. Several works, including PixArt-$\Sigma$ and SANA 1.5~\cite{xie2025sana}, have achieved $4K$ synthesis through extensive high-resolution pretraining, while others~\cite{hoogeboom2023simple,liu2024linfusion,ren2024ultrapixel,xie2023difffit,teng2023relay, zheng2024any, zhang2025diffusion, yu2025ultra,ye2025ultraflux} pursue fine-tuning or training-from-scratch approaches using curated datasets. For example,~\cite{zhang2025diffusion} used wavelet supervision to enhance detail clarity, and~\cite{yu2025ultra} proposed lightweight fine-tuning for adapting to higher resolutions. Although effective, these methods remain constrained by the scarcity of high-quality high-resolution data and substantial GPU requirements, and since the lack of high quality data as lower resolutions, the performance sometimes degrades and sensitive to aspect ratio~\cite{ye2025ultraflux,zhang2025diffusion}. More recent training-free strategies~\cite{guo2024make,qiu2024freescale,du2024max,liu2024hiprompt,wu2025megafusion,shi2025resmaster,kim2025diffusehigh,huang2024fouriscale,bu2025hiflow, zhang2025hidiffusion, zhangfrecas} avoid data collection by leveraging pretrained models directly. 
\yuyao{While promising, these approaches often inherit significant runtime and memory overhead, limiting accessibility for broader use. Other recent methods~\cite{zhang2025hidiffusion, du2024max,zhangfrecas, bu2025hiflow, he2025ragsr, dong2025can, kwon2026reviving, hu2026ultragen} further explore training-free or architecture-specific high-resolution synthesis.~\cite{zhang2025hidiffusion, du2024max} regularizes the denoising trajectory, while~\cite{zhangfrecas, bu2025hiflow} adopts a coarse-to-fine strategy to synthesize high-frequency content, but they either struggle with long-range consistency or require careful stage alignment. RAGSR~\cite{he2025ragsr} focuses on single-image super-resolution with region-text alignment, but its image self-attention still scales with full-resolution tokens. $\Delta$ConvFusion~\cite{dong2025can} replaces self-attention with convolutional modules through distillation, and FCDM~\cite{kwon2026reviving} trains an attention-free diffusion backbone from scratch at low resolutions. UltraGen~\cite{hu2026ultragen} targets video generation with a dual-branch hierarchical architecture, and~\cite{zhang2026hieredit} focuses on ultra-high-resolution image editing. In contrast,~\ourwork~directly reuses pretrained backbones and combines hierarchical local-window attention with scaled-RoPE LR anchors, enabling stable and efficient ultra-high-resolution generation without native HR training data, multi-stage repair, or new architectures.}

\vspace{2mm}
\noindent\textbf{Attention Acceleration.} As image and video resolutions increase, the quadratic complexity of attention becomes the dominant bottleneck. Works including the SANA series~\cite{xie2024sana,xie2025sana, zhu2025dig} leverage linear attention to reduce complexity. While such methods achieve satisfactory performance, the non-injective property and loss of attention spikiness~\cite{han2024bridging,meng2025polaformer, zhang2402hedgehog} inherent in linear attention lead to confusion and inconsistent local details in real-world scenarios. For softmax attention, system-level optimizations such as Flash Attention~\cite{dao2022flashattention, dao2023flashattention, shah2024flashattention} exploit GPU features for faster execution, while quantization~\cite{zhang2025sageattention,zhang2024sageattention2} and sparsity-based designs~\cite{deng2024attention, liu2022dynamic} reduce computational load. Architectural innovations, such as LongFormer~\cite{beltagy2020longformer} and SwinFormer~\cite{liu2021swin}, employ local attention patterns, and more recent works~\cite{lai2025flexprefill, zhang2025spargeattn, xu2025xattention, xi2025sparse,yang2025sparse,yuan2024ditfastattn,zhang2025ditfastattnv2, zhang2025sta, liu2024clear, ren2025grat} propose block sparsification or compression strategies for diffusion transformers. Although these techniques yield noticeable acceleration, the gains remain insufficient for ultra-high-resolution synthesis, and compression often risks degrading fine-grained fidelity. Methods like~\cite{liu2024clear, ren2025grat, zhang2025sta} aim to achieve faster generation while preserving local image quality using local neighbor attention. However, none of them fully maintain long-range dependencies: CLEAR~\cite{liu2024clear} and GRAT~\cite{ren2025grat} require increasingly large local windows/grouping as resolution increases, while STA~\cite{zhang2025sta} sacrifices some long-range consistency, also~\cite{liu2024clear,zhang2025sta} do not support ultra-high-resolution generation without other techniques like SDEdit~\cite{meng2022sdedit}. In contrast, our method explicitly maintains long-range consistency at low resolution through hierarchical local-window attention, achieving stronger acceleration ($>4\times$) and better detail preservation when scaling to 4K and higher, without relying on excessively large local windows or compromising global structure. 

\begin{figure}[t]
    \centering
    \includegraphics[width=\linewidth]{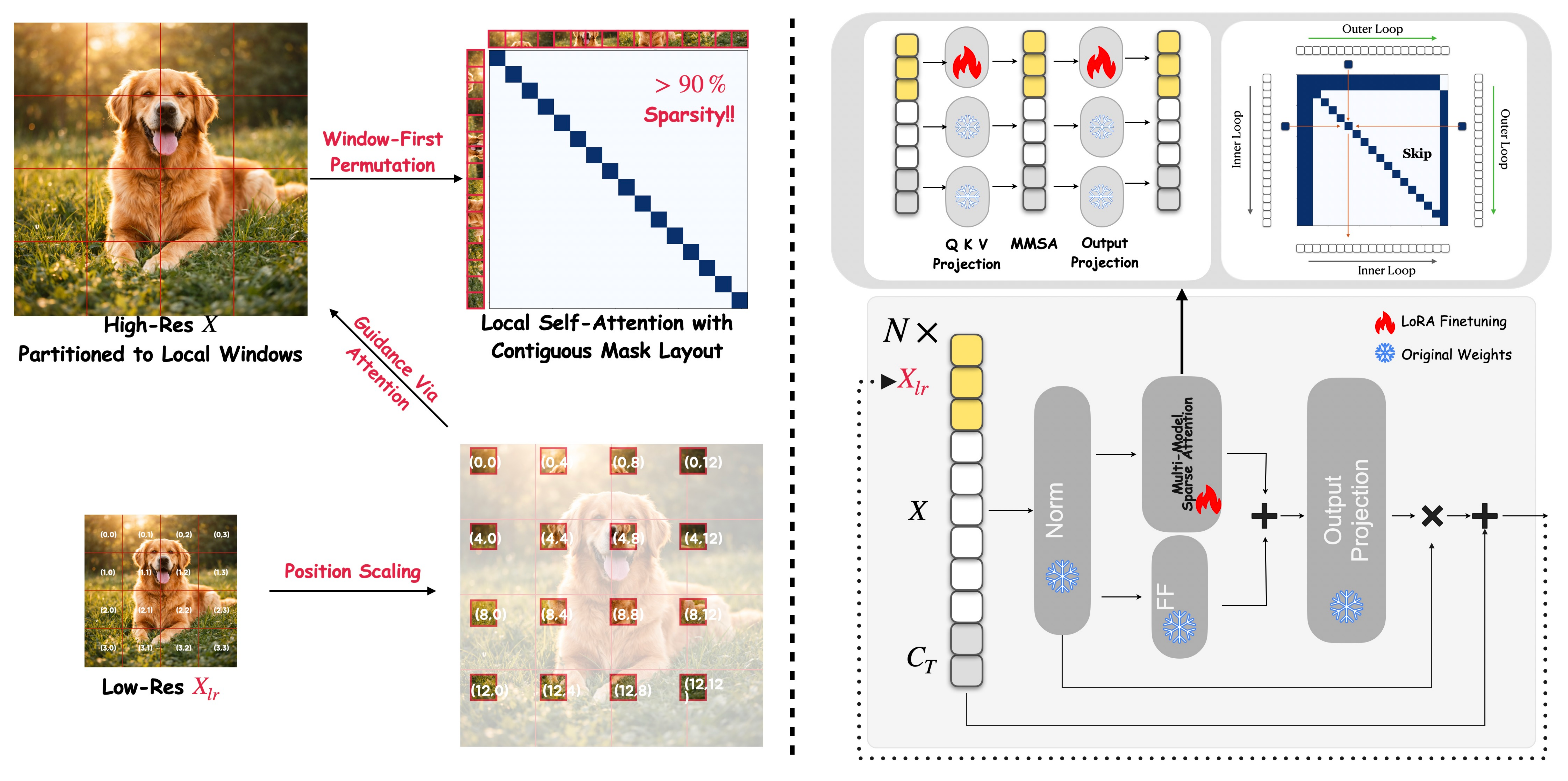}\\
    \vspace{-0.1in}
    \caption{Schematic of \ourwork's attention block modifications. The left column illustrates that high-resolution image latents $X$ are partitioned into local windows (red) and flattened in window-first order that each window attends to itself. Simultaneously, a low-resolution guidance latent $X_{lr}$ (yellow) provides global context to each window via position scaling. The right column shows the joint-denoising process and the attention kernel, that $X, X_{lr}$ are processed together with LoRA applied on the $X_{lr}$ part. The attention mask (upper right)  enforces local and guidance-specific interactions and tile skipping in attention calculation to enable efficient and coherent ultra-high-resolution generation. No additional high-resolution training data are needed.
    }
     \vspace{-0.15in}
    \label{fig:pipeline}
\end{figure}

\vspace{-0.05in}
\section{Method}
\vspace{-0.05in}
Figure~\ref{fig:pipeline} illustrates the overall framework of \ourwork, highlighting the key modifications introduced in the attention blocks to support ultra-high-resolution generation. High-resolution image latents $X$ are partitioned into local windows (red grids), where tokens attend only within their window to capture fine-grained detail efficiently. A low-resolution guidance latent $X_{lr}$ (yellow), enhanced with scaled positional anchors, injects global semantics and preserves long-range consistency across windows. The attention mask is designed to enforce both local and guidance-specific interactions while also enabling tile-skipping to avoid unnecessary computation. Together, these mechanisms allow \ourwork\ to generate coherent and detail-rich $4K$ images with near-linear complexity and without requiring any native high-resolution training data. This schematic provides the foundation for the following technical components.

\vspace{2mm}
\noindent\textbf{The Multi-Modal Diffusion Transformer (MMDiT) Preliminaries}. The MMDiT employed in state-of-the-art models~\cite{blackforestlabs2024flux1dev, cai2025hidream, gao2025seedream, flux-2-2025}, represents the current benchmark architecture for text-to-image generation. MMDiT processes two distinct token modalities: a noisy image token sequence $X \in \mathbb{R}^{N\times d}$ and a text token sequence $C_T \in \mathbb{R}^{M\times d}$. MMDiT processes image and text tokens through a unified Multi-Modal Attention (MMA) mechanism. Specifically, image and text tokens are concatenated as $[C_T; X]$ and processed via self-attention.

Within the MMA framework, spatial information is encoded using Rotary Position Embedding (RoPE)~\cite{su2024roformer}, which applies rotational transformations to capture relative positional relationships. This is mathematically represented as:
$X_{i,j} = X_{i,j} \cdot \text{Rot}(i,j)$,
where $\text{Rot}(i,j)$ denotes the rotation matrix corresponding to position $(i,j)$ in the 2D spatial grid. The MMA mechanism $\text{MMA}([C_T; X])$ is formally defined as:
\begin{equation*}
 \text{Softmax}\left(\frac{\left([Q_T({C_T}),Q(X)]\right)\\([K_T(C_T),K(X)]^T)}{\sqrt{d}}\right)[V_T(C_T),V(X)]
\end{equation*}
where $Q, Q_T$, $K,K_T$, and $V,V_T$ represent the query, key, and value projectionmatrices applied to the position-encoded image and text tokens $X, C_T$. This formulation enables bidirectional attention across all token modalities.

\vspace{2mm}
\noindent\textbf{Efficient Local Window Attention with Window-First Permuted Token Sequence.}
Transformer-based generative models rely on self-attention, whose computational and memory costs scale quadratically with the number of tokens. For an image of size $H \times W$, the number of spatial tokens is $N = H \cdot W$, leading to $O(N^2)$ complexity. Scaling from $1024\times1024$ ($4096$ tokens\footnote{A VAE downsampling factor of 16 yields $\frac{1024}{16}^2 = 4096$ tokens, similar scaling for other resolutions.}) to $4096\times4096$ ($65536$ tokens) increases the cost by a factor of $256$, which is prohibitive even with optimized kernels such as FlashAttention~\cite{dao2022flashattention, dao2023flashattention, shah2024flashattention}.

To overcome this limitation, we partition the high-resolution latent $X \in \mathbb{R}^{H \times W}$ into non-overlapping windows ${x_i \in \mathbb{R}^{l \times l}}$, with window size $l$ bounded by the pretrained resolution (e.g., 1024). Self-attention is computed independently within each window, reducing complexity from $O(N^2)$ to $O\left(\lceil \tfrac{H}{l}\rceil \cdot \lceil \tfrac{W}{l}\rceil \cdot (l^2)^2\right) \approx O(N \cdot l^2)$. In practice, we set $l=16$ (corresponding to $256\times256$ windows) to strike a balance between accuracy and efficiency: larger windows bring little additional benefit in quality, while smaller windows underutilize GPU kernels, which are optimized for tile sizes around $128\times128$\footnote{With 16× VAE downsampling, 256×256 windows yield 16×16 tokens, creating attention matrices that efficiently utilize GPU kernels. Smaller window size produces tokens less than the granularity of 128.}. An ablation study in Section~\ref{sec:ablation} further analyzes this trade-off. With this design, scaling to $4K$ resolution ($N=65536$ tokens) reduces the attention complexity from $O(65536^2)$ to $O(65536 \cdot 16^2)$, yielding a $256\times$ reduction. To maintain boundary consistency, we also allow limited cross-window attention along adjacent regions. Moreover, because the per-window computation cost remains constant, total runtime scales nearly linearly with image resolution, making the generation process both efficient and memory-friendly via integration with the tiling strategy in FlashAttention~\cite{dao2022flashattention,dao2023flashattention}.

A key aspect of our approach is the \textbf{window-first token permutation}: we reorder the flattened image token sequence in a window-first fashion (instead of the default raster-scan order) so that each block in the attention map corresponds directly to a $256\times256$ spatial window in the original image. This ensures that every attention window operates within a fixed RoPE positional range of $256\times256$ which is within the relative range which the model was pretrained regardless of the image resolution, thereby enabling ultra-high-resolution generation without extrapolating beyond learned positional encodings. 

\vspace{2mm}
\noindent\textbf{Maintaining Global Semantics via Low-Resolution Guidance.} Partitioning a high-resolution image into local windows risks creating discontinuities and losing global semantic coherence. The key insight is that global long-range dependencies primarily influence the overall structure and layout of the generated image, but have minimal effect on fine local details. We generate a low-resolution guidance image $X_{lr} \in \mathbb{R}^{h \times w}$ to provide global context. The corresponding position indices are denoted as ${(m,n)}$, where $m\in\{0,1,\dots,h-1\}$ and $n\in\{0,1,\dots,w-1\}$. We define the scaling ratio as $\rho = \frac{H}{h}$ (empirically set to 4 for an optimal balance of performance and efficiency). We then scale the low-resolution image's position indices by this ratio, mapping them to $(\tilde{m},\tilde{n}) = (\rho \cdot m, \rho \cdot n)$. This effectively projects $X_{lr}$ to the same spatial scale as the high-resolution image $X$. Each token $X_{lr}[\tilde{m},\tilde{n}]$ acts as an anchor point, providing contextual information. 
The resulting attention structure is as follows: each high-resolution window attends to its own local and adjacent tokens together with the corresponding scaled region in $X_{lr}$, while $X_{lr}$ tokens attend globally among themselves. In concrete terms, we formulate the attention mask $M$ by allowing the following interactions: text tokens attend to themselves and to $X$; tokens in X attend locally and both text and $X_{lr}$; and tokens in $X_{lr}$ attend to each other. This design ensures semantic consistency across spatially distant regions.
The framework naturally supports recursive generation: the high-resolution output can serve as low-resolution guidance for an even higher resolution, enabling arbitrarily large-scale synthesis with stable memory and computation.

\vspace{2mm}
\noindent\textbf{Parameter-Efficient Joint Denoising.} To integrate the low-resolution guidance $X_{lr}$, we concatenate it with the standard text–image sequence to form $[C_T, X, X_{lr}]$. Since $X_{lr}$ is still an image modality, we can reuse the pretrained VAE and DiT blocks for its processing. However, scaling the positional indices of $X_{lr}$ alters the frequency characteristics of the RoPE embeddings. To adapt to this change, we fine-tune only the query, key, and value projections ($Q, K, V$) that process $X_{lr}$ using LoRA~\cite{hu2022lora}, yielding adapted matrices $\tilde{Q}, \tilde{K}, \tilde{V}$. Importantly, the original $Q, K, V$ are still used for the high-resolution windows $X$, ensuring that pretrained generative capabilities for local content are preserved. The unified attention mechanism $\text{MMA}([X; X_{lr}])$, simplified by omitting text tokens, is then:
\begin{equation*}
\text{Softmax}\left(\frac{\left([Q(X),\tilde{Q}(X_{lr})]\right)([K(X),\tilde{K}(X_{lr})]^T)\cdot M}{\sqrt{d}}\right)[V(X),\tilde{V}(X_{lr})],
\end{equation*}
where $X_{lr}$ are the scaled guidance tokens and $M$ is an attention mask enforcing local window and guidance-specific interactions.  

We train this framework with the standard flow matching objective, extended to include the low-resolution guidance:
\begin{equation*}
\begin{aligned}
L_{\mathrm{FM}}(\theta; C_T, X_{lr})
&= \mathbb{E}_{\substack{
t \sim \mathcal{U}(0,1),\\
X_t \sim p(X \mid t, C_T, X_{lr})
}}
\Big[
\big\| 
f_{\theta}(X_t, t, C_T, X_{lr})
- u_t(X_t; C_T, X_{lr})
\big\|^2
\Big].
\end{aligned}
\end{equation*}

In our implementation, $X_{lr}$ is generated at $256\times256$, while $X$ is trained at $1K\times1K$.
Since the model is merely adapting to a novel attention pattern guided by low-resolution inputs, rather than learning new high-resolution feature, \textbf{no native 4K data} are needed. This design yields three key advantages: (1) training is efficient, since only a small set of LoRA parameters is updated; (2) adaptation can be performed entirely on commodity-resolution data; and (3) ultra-high-resolution synthesis is achieved without the need for prohibitively expensive $4K$ training datasets.

\vspace{2mm}
\noindent\textbf{Inference Acceleration with Low-Res Guidance.}
Since the low-resolution guidance image already encodes the global structure of the target output—differing primarily in high-frequency detail—we leverage it to warm-start the high-resolution sampling process. Specifically, we first upsample the low-resolution reference to match the target resolution, apply sharpening, and add noise scaled to an intermediate timestep $t$, yielding $X_{t}^{\mathrm{ref}} = \mathrm{Interpolate}(X_{\mathrm{lr}})$. The high-resolution latent is then initialized as $X_{\mathrm{hr}}^{t} = \alpha\, X_{T} + (1 - \alpha)\, X_{t}^{\mathrm{ref}}$,
where $\alpha = \frac{t}{T}$ controls the noise ratio and $X_T$ denotes pure Gaussian noise. By injecting the low-frequency components directly, this initialization allows us to bypass the early denoising stages, substantially accelerating inference.



\section{Experiments}

\subsection{Implementation Details}

\noindent\textbf{Experimental Setup.}
We adopt FLUX.1-dev~\cite{blackforestlabs2024flux1dev} as our text-to-image backbone, utilizing the Hugging Face Diffusers library for implementation. Parameter-efficient fine-tuning is conducted using LoRA from the PEFT library, with both rank and alpha equal to 16. LoRA layers are only applied to the Query, Key, and Value projection layers across the DiT blocks. We adopt pyriology scheduler with default learning rate of 1 and weight decay of 0.01, and set bias correction and safe-guard warmup to true. During inference, we set the NTK factor of RoPE embedding to be 10 and disabled the dynamic shifting of the scheduler following~\cite{bu2025hiflow, du2024max}. Our custom attention mechanisms are implemented following the principles of FlashAttention-2~\cite{dao2023flashattention} and SageAttention~\cite{zhang2024sageattention2}. The model was trained for 20,000 steps on two NVIDIA RTX A6000 Ada GPUs (48GB VRAM each), using a per-GPU batch size of 1 with gradient accumulation over 4 steps. All subsequent experiments were conducted on this same hardware configuration. For the fine-tuning dataset, we generated 10,000 images at a resolution of $1024 \times 1024$ using the base FLUX.1-dev model, prompted by a randomly sampled collection of high-quality text descriptions. The kernel size is designed to be Q-block=128 and K-block=64 (128 for blocksparse Flash Attention) which suits the granularity supported for~\cite{zhang2024sageattention2} on NVIDIA A6000 Ada. All experiments were conducted on one NVIDIA RTX Pro 6000 Blackwell GPU (96GB VRAM).
\begin{figure}[t]
    \centering
    \includegraphics[width=\linewidth]{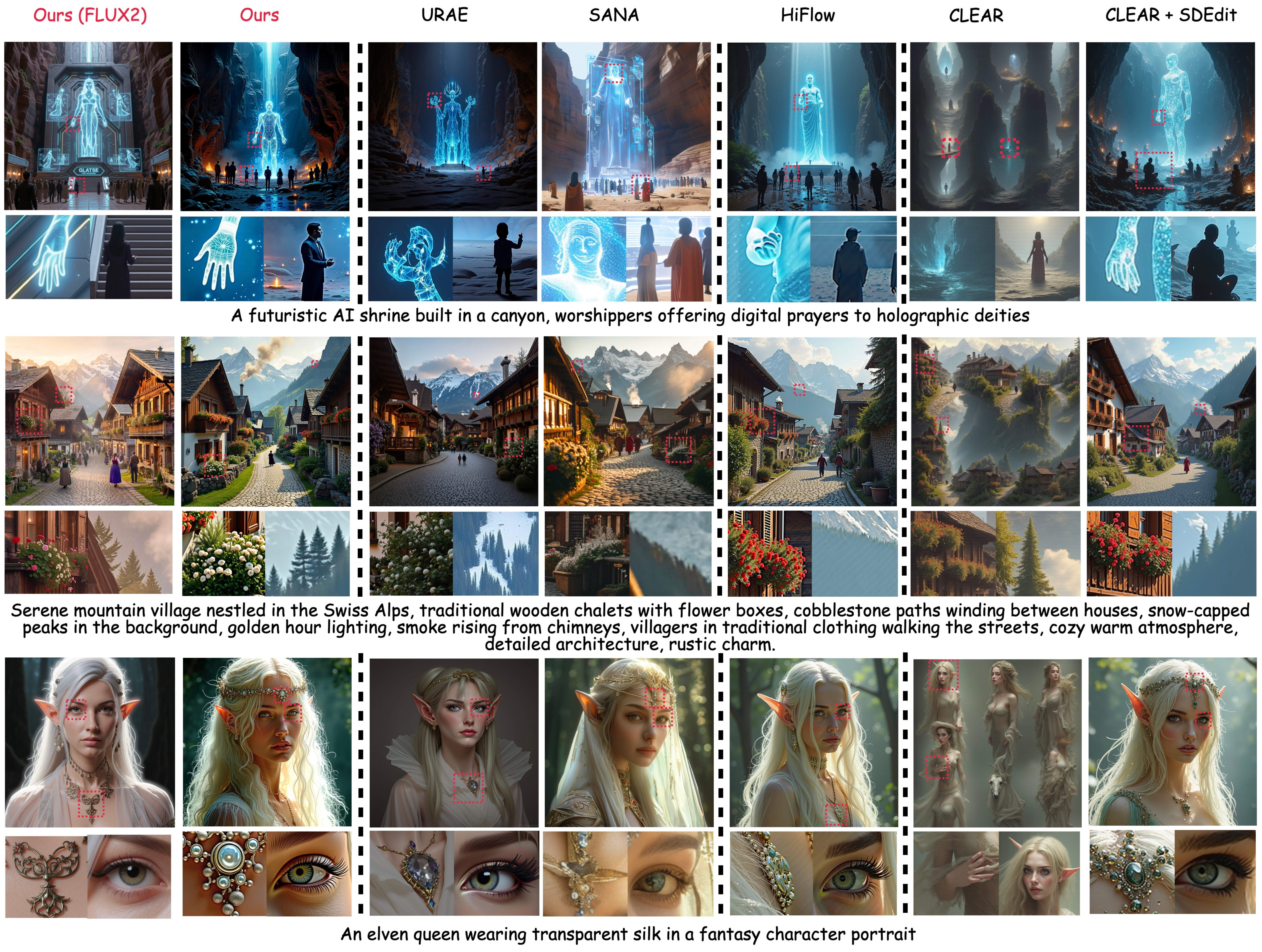}
    \vspace{-0.2in}
    \caption{4K comparison with leading baselines based on FLUX.1 (except for SANA). Zoom in to observe the fine details. \ourwork\ produces results with superior fidelity. Additional comparisons with other baselines are provided in the Appendix. The second column section shows methods finetuned on 4K data, the third column section shows training-free method, the right column section shows another efficient attention method. Our method achieves sharper results than other methods. Notably, although HiFlow produces good results but it shows diagonal artifacts. We also provide additional results for our method based on FLUX.2, which yields more natural textures.}
    \vspace{-0.2in}
    \label{fig:qualitative}
\end{figure}

\noindent\textbf{Metrics and Evaluation Protocol.}
To ensure a comprehensive and diverse evaluation, we generated a benchmark suite of 1,000 high-quality prompts across various categories using GPT-4o. We assess performance using a standard set of metrics: CLIP Score~\cite{radford2021learning} for prompt-image alignment, CLIP-IQA~\cite{wang2022exploring}, Fréchet Inception Distance (FID)~\cite{heusel2017gans}, and Inception Score (IS)~\cite{salimans2016improved} for image quality. The FID score is computed against a reference set of 10,000 real images from the LAION-High-Resolution~\cite{schuhmann2022laion} dataset. To specifically evaluate the fidelity of local details, we also report patch-based versions of these metrics, $\text{FID}_{\text{patch}}$ and $\text{IS}_{\text{patch}}$, calculated on local image patches. For all comparisons, competing training-free methods were evaluated on the same FLUX.1-dev base model, following their official implementations to ensure fairness.

\subsection{Comparison with State-of-the-Art Methods}

We compare our results with methods that can also generate 4K resolution images. These include data-intensive, training-based methods: SANA~\cite{xie2024sana}, Diffusion-4K~\cite{zhang2025diffusion}, FLUX+URAE~\cite{yu2025ultra}, UltraFLUX~\cite{ye2025ultraflux}, and UltraPixel~\cite{ren2024ultrapixel}; a super-resolution baseline: FLUX+BSRGAN~\cite{zhang2021designing}; and several training-free approaches: DemoFusion~\cite{du2024demofusion}, DiffuseHigh~\cite{kim2025diffusehigh}, I-MAX~\cite{du2024max}, and HiFlow~\cite{bu2025hiflow} and an efficient attention method CLEAR~\cite{liu2024clear} (with the 4K ability comes from SDEdit~\cite{meng2022sdedit} and the optimal setting of window size 32).

\begin{figure}[htbp]
\centering
\vspace{-0.15in}
\includegraphics[width=\textwidth]{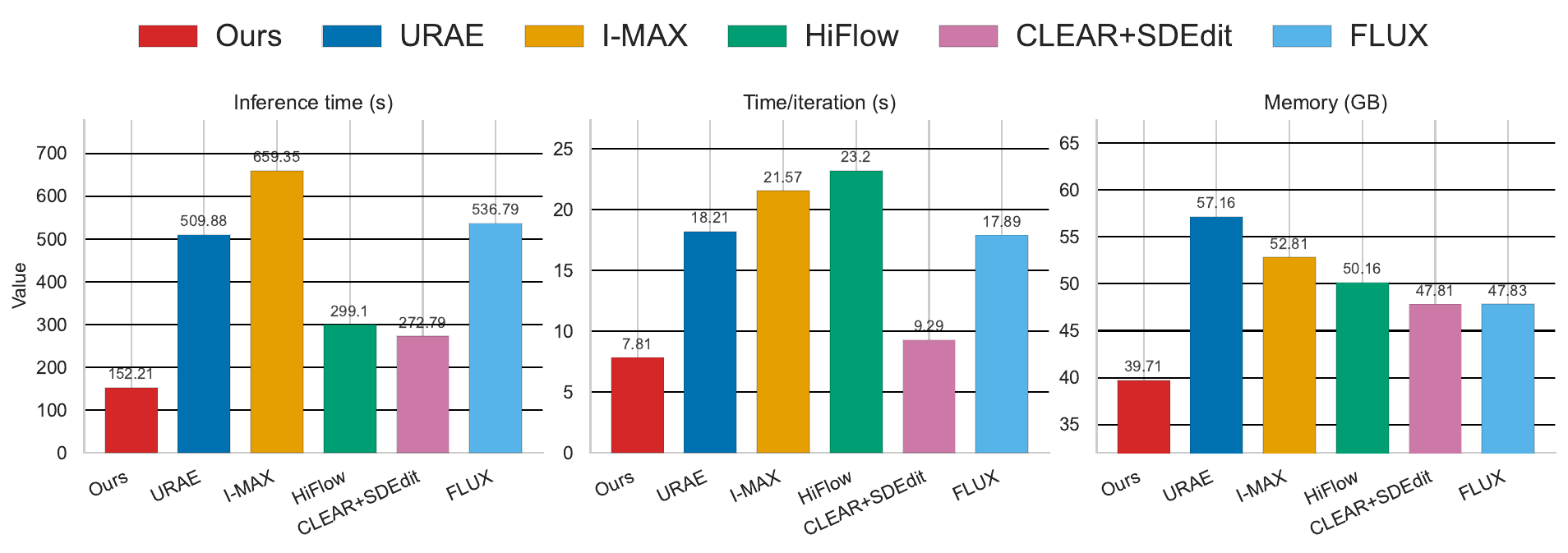}\\
\vspace{-0.15in}
\caption{Comparison of latency (sec; time to generate one 4K latents (65536 tokens)), time per iteration (sec), and memory (GB) and  across FLUX.1-dev-based 4K generation methods. Our method achieves significant speedup and memory reduction.}
\vspace{-0.15in}
\label{tab:memory_latency}
\end{figure}

\noindent\textbf{Qualitative Comparison.}
Figure~\ref{fig:teaser} presents qualitative results of \ourwork\ at various resolutions from $2K$ to $>8K$, demonstrating superior text-to-image generation with high-fidelity details and coherent global structure. A more detailed comparison at 4K is shown in Figure~\ref{fig:qualitative}, where we include a representative subset of leading baselines for clarity; comprehensive comparisons against the fourteen methods, as well as additional 2K and 4K results, are provided in the Appendix. At 4K resolution, our method renders anatomically correct hands with exceptional detail as shown in the first row, while SANA produces a blurry image, HiFlow introduces diagonal artifacts, and I-MAX fails to generate the correct number of fingers. In the second row, \ourwork\ consistently preserves textural clarity, such as tree structures, surpassing competitors while maintaining global coherence. The third row highlights highly detailed eye and jewelries generation at 4K, where our method clearly captures sharper details of eyelashes, and jewelries with fine structure. Training free methods like HiFlow~\cite{bu2025hiflow} in the fourth row frequently shows diagonal artifacts, while another efficient attention method~\cite{liu2024clear} failed to preserve global consistency as shown in the fifth columns. These examples demonstrate \ourwork’s ability to synthesize images that combine fine-grained local details with coherent large-scale composition, validating its performance across diverse high-resolution scenarios.

\begin{table}[htbp]
\vspace{-0.15in}
\centering
\caption{Quantitative comparison at 4K × 4K resolution. The best result is highlighted in bold, while the second-best result is underlined. ~\ourwork~consistently perform competitive results even compared to training-based methods. We also include our work with stronger base model FLUX2 here (in Italic) for reference.}
\label{quantitative-4k}
\begin{tabular}{l|cccccc}
\hline
\textbf{Method} 
& \textbf{$\text{CLIP}_{\text{IQA}}$} $\uparrow$
& \textbf{FID} $\downarrow$ 
& \textbf{FID$_\text{patch}$} $\downarrow$ 
& \textbf{IS} $\uparrow$ 
& \textbf{IS$_\text{patch}$} $\uparrow$ 
& \textbf{$\text{CLIP}_{\text{score}}$} $\uparrow$ \\ 
\hline
SANA             & \underline{0.4457}        & 76.31      & 74.27      & 16.68      & \textbf{14.02} & 0.3197   \\
Diffusion-4K        & 0.3012      & 121.85     & 120.59     & 14.39      & 10.77      & 0.2844      \\
UltraPixel         &  0.4421       & 77.42      & 70.94      & 16.98      & 13.26      & \textbf{0.3251} \\
URAE              & 0.4369         & \underline{67.39} & \underline{62.56} & 17.11 & 12.39 & 0.3204 \\
FLUX+BSRGAN       & 0.3897   & 71.39      & 63.45      & 17.08      & 12.87      & 0.3210      \\
UltraFlux       & 0.4371   &  75.89     &  65.12     & 17.01      &  13.24     &    0.3165   \\
\hline
DemoFusion           & 0.4392    & 74.89      & 66.37      & 16.23      & 13.02      & 0.3187      \\
DiffuseHigh           & 0.4221  & 81.54      & 73.35      & 16.15      & 13.08      & 0.3175      \\
HiDiffusion & 0.4021 & 172.46 & 217.49 & 9.43 & 8.12 & 0.3129 \\
FreCas & 0.2704 & 213.44 & 322.08 & 7.98 & 6.84 & 0.2805 \\
I-MAX          & 0.4381          & 70.33      & 65.67      & 16.50      & 12.69      & 0.3211      \\
HiFlow        & 0.4407            & 69.18      & 63.72      & \underline{17.13} & 13.43 & 0.3113 \\
\hline
CLEAR       & 0.3463            & 131.15      & 120.79      & 14.18 & 11.92 & 0.2913 \\
CLEAR+SDEdit        &   0.4352          & 88.23      & 87.19      & 16.84 & \underline{13.78} & 0.3221 \\
\hline
\textbf{Ours}  & \textbf{0.4505} & \textbf{67.03} & \textbf{61.78} & \textbf{17.21} & 13.31 & \underline{0.3231} \\
\textbf{Ours(FLUX2)\*}  & \textit{\textbf{0.4594}} & \textit{\textbf{61.38}} & \textit{\textbf{59.82}} & \textit{\textbf{18.02}} & \textit{\textbf{14.93}} & \textit{\textbf{0.3267}} \\
\hline
\end{tabular}
\vspace{-0.2in}
\end{table}

\noindent\textbf{Quantitative Comparison.}
Table~\ref{quantitative-4k} presents a quantitative comparison of our method against the nine state-of-the-art methods at 4K resolution \yuyao{where most of them are based on the same FLUX.1-dev backbone}. The results show that our method achieves highly competitive performance across all metrics, even outperforming training-based methods that rely on extensive 4K-resolution training data. Table~\ref{tab:performance_comparison_alt} demonstrates consistent performance on the GenEval~\cite{ghosh2023geneval} benchmark, confirming that \ourwork's output quality is resolution-agnostic. 

\begin{figure}[htbp]
        \centering
        \vspace{-0.2in}
        \includegraphics[width=\linewidth]{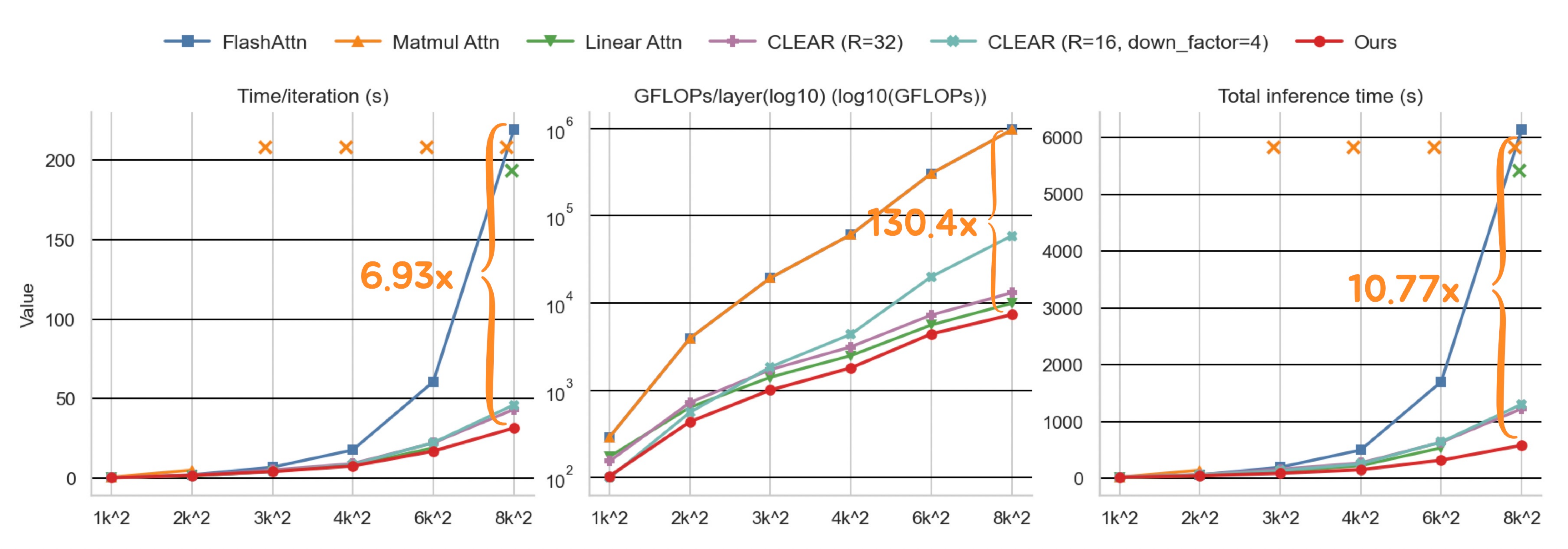}\\
        \caption{ \textbf{Efficiency comparisons across resolutions (1K to 8K).} We report per-iteration latency (\textbf{Left}), floating-point operations per attention layer (\textbf{Middle}), and total inference time (\textbf{Right}) for our method, naive matrix multiplication attention, Linear Attention, FlashAttention-2, and two linearlized attention configurations from CLEAR. At 8K resolution, our method achieves a $6.93\times$ per-iteration speedup, $130.4\times$ reduction in attention computation, and $10.77\times$ total inference speedup over FLUX with FlashAttention, while consistently outperforming CLEAR and Linear Attention across all metrics. The "X" mark means out-of-memory (OOM). Linear Attention reaches OOM at 8K resolution, while the naive one reaches OOM at 3K resolution.}
        \label{fig:scaling}
        \vspace{-0.15in}
\end{figure}

\vspace{2mm}
\noindent\textbf{Efficiency and Scalability Analysis.} The efficiency gains of our approach are summarized in Figure~\ref{tab:memory_latency} and Figure~\ref{fig:scaling}. For a fair comparison, we focus on methods built upon the FLUX.1-dev backbone without architectural modifications. Under this setting, our method achieves a $1.7\times$ to $4.3\times$ inference speedup over its counterparts (with all methods exhibiting greater than $2.3\times$ speedup per denoising iteration). Although CLEAR~\cite{liu2024clear} (augmented with SDEdit~\cite{meng2022sdedit}) attains a comparable per-iteration speed, our method surpasses it in overall inference time by substantially reducing the number of required denoising steps. In terms of memory consumption, our method is highly efficient: excluding the base model's footprint $\sim32\,GB)$, it requires only $7.3\,GB$ of additional VRAM, whereas a naive implementation with FlashAttention consumes $15.6\,GB$. This significant reduction in memory overhead enables inference on consumer-level GPUs without module offloading, and facilitates more GPU-friendly fine-tuning.

\begin{table}[t]
\centering
\caption{Performance of our method from 1K to 4K on the GenEval benchmark. Results show that output quality is maintained across resolutions despite training on 1K data, demonstrating scalability and resolution-agnostic design.}
\label{tab:performance_comparison_alt}
\begin{tabular}{|l|c|c|c|c|c|c|c|}
\hline
\textbf{Model} & \textbf{Overall} & \textbf{1 Obj} & \textbf{2 Obj} & \textbf{Counting} & \textbf{Colors} & \textbf{Position} & \textbf{Attr. Binding} \\
\hline
FLUX.1-Dev & 0.66 & 0.98 & 0.81 & 0.74 & 0.79 & 0.22 & 0.45 \\
\hline
Ours(2K) & 0.67 & 0.99 & 0.83 & 0.74 & 0.81 & 0.22 & 0.45 \\
\hline
Ours(3K) & 0.67 & 0.98 & 0.82 & 0.73 & 0.81 & 0.23 & 0.46 \\
\hline
Ours(4K) & 0.67 & 0.98 & 0.83 & 0.75 & 0.79 & 0.24 & 0.45 \\
\hline
\end{tabular}
\vspace{-0.2in}
\end{table}

Figure~\ref{fig:scaling} further illustrates the scalability of our approach. As the target resolution increases from 1K to 8K, our method achieves a $10.77\times$ reduction in latency and a $130.4\times$ reduction in floating-point computation within the attention layer compared to FlashAttention-2~\cite{dao2023flashattention}. Compared with another linearized attention mechanism~\cite{liu2024clear}, our approach remains consistently faster and more memory-efficient. We select two optimal configurations from their paper—CLEAR with a window radius of 32 and a radius of 16 with a downsample factor of 4—and as shown in the line chart, our method yields faster inference and lower computational cost in the attention layer as resolution scales, while also achieving superior generation quality as demonstrated in Table~\ref{quantitative-4k} and Figure~\ref{fig:qualitative}. We attribute these efficiency gains to two key design choices: (1)~our dense-block-style local window attention is more hardware-friendly, with window sizes that align well with GPU architectures; and (2)~we fully leverage low-resolution guidance as an initialization signal, in contrast to CLEAR, which employs low-resolution guidance solely as a starting point for SDEdit due to its limited scalability across resolutions. For linear attention, while also having relatively low GFLOPs, it requires scaling the hidden dimension with sequence length to maintain performance~\cite{han2024bridging, meng2025polaformer, zhang2402hedgehog}, and it is not able to produce meaningful results via directly substituting the softmax based attentions in pretrained models~\cite{liu2024clear}. In contrast, our method uses local window attention, which does not need to scale with resolution, enabling consistent memory and latency efficiency as resolution increases. Since linear attention requires the hidden dimension to scale with resolution, previous works would need to fine-tune on high-resolution data for each target resolution, whereas our framework achieves ultra-high-resolution synthesis using only 1K-resolution training data without any performance drop. Collectively, these results validate the scalability, efficiency, and robustness of our proposed framework.

\section{Ablation Study}\label{sec:ablation}
To validate our core design choices, we conduct a series of ablation studies analyzing the impact of window size, guidance resolution ratio, attention patterns, and token permutation strategies.

\begin{table}[htbp]
\centering
\vspace{-0.2in}
\caption{Ablation experiments investigating the effects of window size and the high-to-low resolution ratio demonstrated a negligible impact on model performance. Therefore, to prioritize computational efficiency, the configuration utilizing the smallest window size and the largest ratio was chosen for all subsequent experiments.}
\begin{tabular}{|l|c|c|c|c|}
\hline
\textbf{Method} & \textbf{ws256 $\rho$4} & \textbf{ws256 $\rho$2} & \textbf{ws512 $\rho$4} & \textbf{ws1024 $\rho$4} \\
\hline
$\text{FiD}\downarrow$ &67.23 &67.19 & 66.31& 67.83\\
$\text{FiD}_{\text{patch}}\downarrow$ &61.46 &62.28 &63.48 & 62.61\\
\hline
\end{tabular}
\vspace{-0.2in}
\label{tab:ablation-window-size}
\end{table}

\noindent\textbf{Window size and resolution ratio.} We conduct ablation study on window size (ws) and ratio between high-res and low-res image ($\rho$). We experiment with window sizes from 256, 512, and 1024, and $\rho=2,4$. As shown in Table~\ref{tab:ablation-window-size}, the variations in FiD scores across these configurations were not statistically significant. To maximize computational efficiency without compromising performance, we selected the smallest window size (ws=256) and the largest ratio ($\rho$=4). It is worth noting that smaller window sizes or larger ratios are constrained by the fixed granularity of the kernel depending on the hardware design of the GPU.

\vspace{2mm}
\noindent\textbf{Attention Pattern}  
We conducted an ablation study to evaluate the impact of different text-to-image attention mechanisms. Case 1 examines the effect of attending to low-resolution tokens $X_{\text{lr}}$, comparing (a) high-resolution tokens $X$ attending to both textual embeddings and $X_{\text{lr}}$ versus (b) attending only to textual embeddings. Case 2 investigates the effect of attending to neighboring windows, comparing (a) each window of $X$ attending to its neighbors versus (b) attending only to itself.  

\begin{figure}[htbp]
        \centering
        \vspace{-0.25in}
        \includegraphics[width=\linewidth]{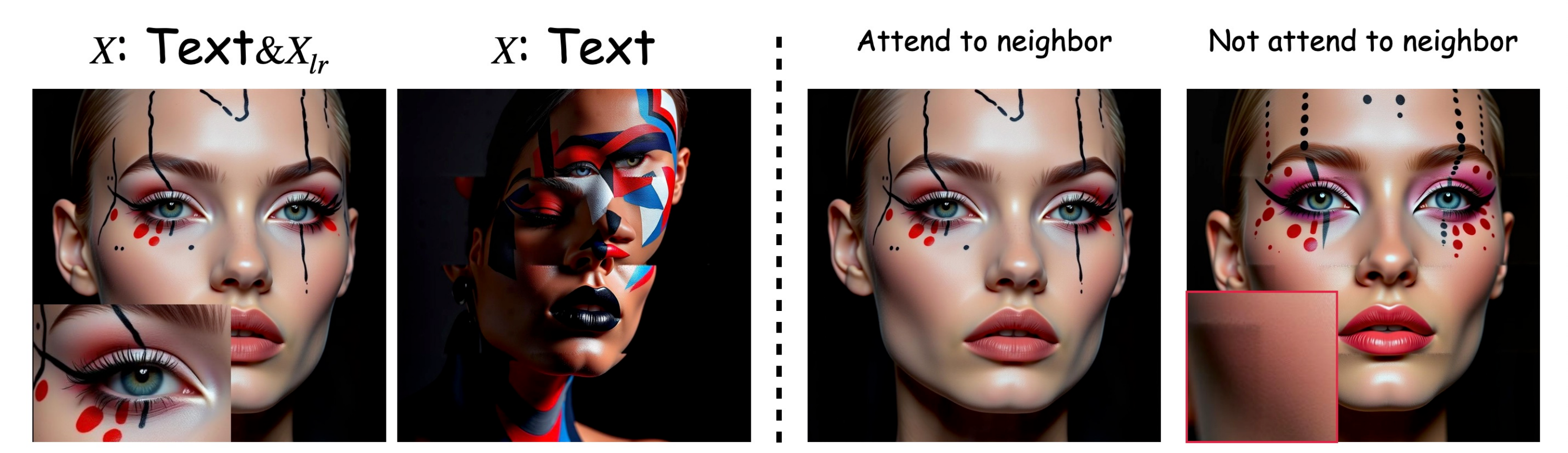}\\
        \vspace{-0.1in}
        \caption{Ablation Study on attention scale. Images from left to right corresponds to cases that 1) High-res $X$ attends to both text and $X_{lr}$. 2) $X$ attends to only Text. 3) Each window of $X$ attends to its neighbors. 4) $X$ only attends to itself. 1) and 2) demonstrates the effectiveness of our low-res guidance, and 3) and 4) illustrates that allowing each window to attend to part of the neighbors will solve the boundary issues.}
         \vspace{-0.15in}
        \label{fig:attn-scale}
    \end{figure}

As shown in Figure~\ref{fig:attn-scale}, including low-resolution tokens in Case 1 produces consistently stable results, whereas excluding them causes discontinuities and multi-face artifacts, highlighting the importance of low-resolution guidance. In Case 2, attending to neighboring windows effectively eliminates boundary artifacts, while incurring only additional computational cost of less than 2\%.

\noindent\textbf{Robustness on Different Models}
To demonstrate the robustness of~\ourwork, we conducted additional experiments on FLUX.2-series. We provide both qualitative (Figure~\ref{fig:qualitative} and Figure~\ref{fig:teaser}) and quantitative (Figure~\ref{fig:teaser}). (Table~\ref{quantitative-4k} results for FLUX.2-Klein-9B, which is a time-distilled 4 step model. We therefore achieve \textbf{sub 30-second generation} for 4K images and \textbf{sub 2-minute} for 8K images, enabling maximum efficiency in ultra-high-resolution content creation. 

\vspace{2mm}
\noindent\textbf{Content in the Supplementary Materials.} We will include (1) More 8K, 4K and 2K resolution galleries, and the PDF file for the Teasor figure. (2) a detailed quantitative 4K resolution comparison against all 14 baselines, (3) demo for our method's compatability with different styled LoRA models, robustness to model sizes and aspect ratios, (4) more analysis and ablations.


\section{Conclusion}
We present \ourwork, a novel framework enabling pre-trained diffusion transformers to generate ultra-high-resolution images \textit{without requiring additional high-resolution training data}. \ourwork\ introduces a hierarchical attention mechanism that partitions high-resolution latents into fixed-size local windows while maintaining global coherence through low-resolution guidance with scaled positional anchors. The framework comprises three key components: (i) efficient local window attention that reduces quadratic complexity to near-linear scaling; (ii) global semantic preservation via low-resolution guidance latents with position scaling; and (iii) parameter-efficient joint denoising through LoRA-based adaptations trained solely on commodity resolutions. Extensive experiments demonstrate that \ourwork\ achieves superior visual quality at 4K resolution while delivering over $10\times$ \textit{speedup} in $8K$ resolution and \textit{less memory usage} compared to dense attention baselines, establishing a practical paradigm for scalable ultra-high-resolution text-to-image generation.

\bibliographystyle{splncs04}
\bibliography{main}

@String(ICML  = {Int. Conf. Mach. Learn.})

@String(ICLR  = {Int. Conf. Learn. Represent.})

@String(AAAI  = {AAAI})

@String(ICML  = {ICML})

@String(ICLR  = {ICLR})

@inproceedings{chen2024pixartsigma,
  title={Pixart-$\sigma$: Weak-to-strong training of diffusion transformer for 4k text-to-image generation},
  author={Chen, Junsong and Ge, Chongjian and Xie, Enze and Wu, Yue and Yao, Lewei and Ren, Xiaozhe and Wang, Zhongdao and Luo, Ping and Lu, Huchuan and Li, Zhenguo},
  booktitle={European Conference on Computer Vision},
  pages={74--91},
  year={2024},
  organization={Springer}
}

@inproceedings{chen2024pixart,
  title={PIXART-$\delta$: Fast and Controllable Image Generation with Latent Consistency Models},
  author={Chen, Junsong and Luo, Simian and Xie, Enze},
  booktitle={ICML 2024 Workshop on Theoretical Foundations of Foundation Models},
    year={2024}
}

@inproceedings{chen2023pixart,
  title={PixArt-$\alpha$: Fast Training of Diffusion Transformer for Photorealistic Text-to-Image Synthesis},
  author={Chen, Junsong and Yu, Jincheng and Ge, Chongjian and Yao, Lewei and Xie, Enze and Wang, Zhongdao and Kwok, James T and Luo, Ping and Lu, Huchuan and Li, Zhenguo},
  booktitle={ICLR},
  year={2024}
}

@inproceedings{ding2021cogview,
  title={CogView: mastering text-to-image generation via transformers},
  author={Ding, Ming and Yang, Zhuoyi and Hong, Wenyi and Zheng, Wendi and Zhou, Chang and Yin, Da and Lin, Junyang and Zou, Xu and Shao, Zhou and Yang, Hongxia and others},
  booktitle={Proceedings of the 35th International Conference on Neural Information Processing Systems},
  pages={19822--19835},
  year={2021}
}

@article{blackforestlabs2024flux1dev,
      title={FLUX.1 Kontext: Flow Matching for In-Context Image Generation and Editing in Latent Space},
      author={{Black Forest Labs}},
      year={2025},
      journal={arXiv preprint arXiv:2506.15742},
}

@article{li2024playground,
  title={Playground v2. 5: Three insights towards enhancing aesthetic quality in text-to-image generation},
  author={Li, Daiqing and Kamko, Aleks and Akhgari, Ehsan and Sabet, Ali and Xu, Linmiao and Doshi, Suhail},
  journal={arXiv preprint arXiv:2402.17245},
  year={2024}
}

@inproceedings{peebles2023scalable,
  title={Scalable diffusion models with transformers},
  author={Peebles, William and Xie, Saining},
  booktitle={Proceedings of the IEEE/CVF international conference on computer vision},
  pages={4195--4205},
  year={2023}
}

@inproceedings{podell2023sdxl,
  title     = {{SDXL}: Improving Latent Diffusion Models for High-Resolution Image Synthesis},
  author    = {Podell, Dustin and English, Zion and Lacey, Kyle and Blattmann, Andreas and Dockhorn, Tim and M{\"u}ller, Jonas and Penna, Joe and Rombach, Robin},
  booktitle = {International Conference on Learning Representations},
  year      = {2024}
}

@inproceedings{rombach2022high,
  title={High-resolution image synthesis with latent diffusion models},
  author={Rombach, Robin and Blattmann, Andreas and Lorenz, Dominik and Esser, Patrick and Ommer, Bj{\"o}rn},
  booktitle={Proceedings of the IEEE/CVF conference on computer vision and pattern recognition},
  pages={10684--10695},
  year={2022}
}

@article{cai2025hidream,
  title={HiDream-I1: A High-Efficient Image Generative Foundation Model with Sparse Diffusion Transformer},
  author={Cai, Qi and Chen, Jingwen and Chen, Yang and Li, Yehao and Long, Fuchen and Pan, Yingwei and Qiu, Zhaofan and Zhang, Yiheng and Gao, Fengbin and Xu, Peihan and others},
  journal={arXiv preprint arXiv:2505.22705},
  year={2025}
}

@article{li2024hunyuan,
  title={Hunyuan-dit: A powerful multi-resolution diffusion transformer with fine-grained chinese understanding},
  author={Li, Zhimin and Zhang, Jianwei and Lin, Qin and Xiong, Jiangfeng and Long, Yanxin and Deng, Xinchi and Zhang, Yingfang and Liu, Xingchao and Huang, Minbin and Xiao, Zedong and others},
  journal={arXiv preprint arXiv:2405.08748},
  year={2024}
}

@article{xue2023raphael,
  title={Raphael: Text-to-image generation via large mixture of diffusion paths},
  author={Xue, Zeyue and Song, Guanglu and Guo, Qiushan and Liu, Boxiao and Zong, Zhuofan and Liu, Yu and Luo, Ping},
  journal={Advances in Neural Information Processing Systems},
  volume={36},
  pages={41693--41706},
  year={2023}
}

@inproceedings{xie2025sana,
  title={SANA 1.5: Efficient Scaling of Training-Time and Inference-Time Compute in Linear Diffusion Transformer},
  author={Xie, Enze and Chen, Junsong and Zhao, Yuyang and YU, Jincheng and Zhu, Ligeng and Lin, Yujun and Zhang, Zhekai and Li, Muyang and Chen, Junyu and Cai, Han and others},
  booktitle={Forty-second International Conference on Machine Learning},
  year={2025}
}

@inproceedings{xie2024sana,
  author       = {Enze Xie and
                  Junsong Chen and
                  Junyu Chen and
                  Han Cai and
                  Haotian Tang and
                  Yujun Lin and
                  Zhekai Zhang and
                  Muyang Li and
                  Ligeng Zhu and
                  Yao Lu and
                  Song Han},
  title        = {{SANA:} Efficient High-Resolution Text-to-Image Synthesis with Linear
                  Diffusion Transformers},
  booktitle    = {The Thirteenth International Conference on Learning Representations},
  year         = {2025}
}

@inproceedings{guo2024make,
  title={Make a cheap scaling: A self-cascade diffusion model for higher-resolution adaptation},
  author={Guo, Lanqing and He, Yingqing and Chen, Haoxin and Xia, Menghan and Cun, Xiaodong and Wang, Yufei and Huang, Siyu and Zhang, Yong and Wang, Xintao and Chen, Qifeng and others},
  booktitle={European conference on computer vision},
  pages={39--55},
  year={2024},
  organization={Springer}
}

@inproceedings{hoogeboom2023simple,
  title={simple diffusion: End-to-end diffusion for high resolution images},
  author={Hoogeboom, Emiel and Heek, Jonathan and Salimans, Tim},
  booktitle={International Conference on Machine Learning},
  pages={13213--13232},
  year={2023},
  organization={PMLR}
}

@inproceedings{yu2025ultra,
  title     = {Ultra-Resolution Adaptation with Ease},
  author    = {Yu, Ruonan and Liu, Songhua and Tan, Zhenxiong and Wang, Xinchao},
  booktitle   = {International Conference on Machine Learning},
  year      = {2025},
}

@article{liu2024linfusion,
  title={Linfusion: 1 gpu, 1 minute, 16k image},
  author={Liu, Songhua and Yu, Weihao and Tan, Zhenxiong and Wang, Xinchao},
  journal={arXiv preprint arXiv:2409.02097},
  year={2024}
}

@article{ren2024ultrapixel,
  title={Ultrapixel: Advancing ultra high-resolution image synthesis to new peaks},
  author={Ren, Jingjing and Li, Wenbo and Chen, Haoyu and Pei, Renjing and Shao, Bin and Guo, Yong and Peng, Long and Song, Fenglong and Zhu, Lei},
  journal={Advances in Neural Information Processing Systems},
  volume={37},
  pages={111131--111171},
  year={2024}
}

@inproceedings{xie2023difffit,
  title={Difffit: Unlocking transferability of large diffusion models via simple parameter-efficient fine-tuning},
  author={Xie, Enze and Yao, Lewei and Shi, Han and Liu, Zhili and Zhou, Daquan and Liu, Zhaoqiang and Li, Jiawei and Li, Zhenguo},
  booktitle={Proceedings of the IEEE/CVF International Conference on Computer Vision},
  pages={4230--4239},
  year={2023}
}

@inproceedings{teng2023relay,
  author       = {Jiayan Teng and
                  Wendi Zheng and
                  Ming Ding and
                  Wenyi Hong and
                  Jianqiao Wangni and
                  Zhuoyi Yang and
                  Jie Tang},
  title        = {Relay Diffusion: Unifying diffusion process across resolutions for
                  image synthesis},
  booktitle    = {The Twelfth International Conference on Learning Representations},
  year         = {2024}
}

@inproceedings{zheng2024any,
  title={Any-size-diffusion: Toward efficient text-driven synthesis for any-size hd images},
  author={Zheng, Qingping and Guo, Yuanfan and Deng, Jiankang and Han, Jianhua and Li, Ying and Xu, Songcen and Xu, Hang},
  booktitle={Proceedings of the AAAI Conference on Artificial Intelligence},
  volume={38},
  number={7},
  pages={7571--7578},
  year={2024}
}

@inproceedings{zhang2025diffusion,
  title={Diffusion-4k: Ultra-high-resolution image synthesis with latent diffusion models},
  author={Zhang, Jinjin and Huang, Qiuyu and Liu, Junjie and Guo, Xiefan and Huang, Di},
  booktitle={Proceedings of the Computer Vision and Pattern Recognition Conference},
  pages={23464--23473},
  year={2025}
}

@inproceedings{qiu2024freescale,
  title={Freescale: Unleashing the resolution of diffusion models via tuning-free scale fusion},
  author={Qiu, Haonan and Zhang, Shiwei and Wei, Yujie and Chu, Ruihang and Yuan, Hangjie and Wang, Xiang and Zhang, Yingya and Liu, Ziwei},
  booktitle={Proceedings of the IEEE/CVF International Conference on Computer Vision},
  pages={16893--16903},
  year={2025}
}

@article{du2024max,
  title={I-Max: Maximize the Resolution Potential of Pre-trained Rectified Flow Transformers with Projected Flow},
  author={Ruoyi Du and Dongyang Liu and Le Zhuo and Qi Qin and Hongsheng Li and Zhanyu Ma and Peng Gao},
  journal={arXiv preprint arXiv:2410.07536},
  year={2024},
}

@article{liu2024hiprompt,
  title   = {{HiPrompt}: Tuning-free Higher-Resolution Generation with Hierarchical {MLLM} Prompts},
  author  = {Liu, Xinyu and He, Yingqing and Guo, Lanqing and Li, Xiang and Jin, Bu and Li, Peng and Li, Yan and Chan, Chi-Min and Chen, Qifeng and Xue, Wei and Liu, Jiashi and Liu, Shengfeng},
  journal = {International Journal of Computer Vision},
  volume  = {134},
  pages   = {147},
  year    = {2026},
}

@inproceedings{wu2025megafusion,
  title={Megafusion: Extend diffusion models towards higher-resolution image generation without further tuning},
  author={Wu, Haoning and Shen, Shaocheng and Hu, Qiang and Zhang, Xiaoyun and Zhang, Ya and Wang, Yanfeng},
  booktitle={2025 IEEE/CVF Winter Conference on Applications of Computer Vision},
  pages={3944--3953},
  year={2025},
  organization={IEEE}
}

@inproceedings{shi2025resmaster,
  title={Resmaster: Mastering high-resolution image generation via structural and fine-grained guidance},
  author={Shi, Shuwei and Li, Wenbo and Zhang, Yuechen and He, Jingwen and Gong, Biao and Zheng, Yinqiang},
  booktitle={Proceedings of the AAAI Conference on Artificial Intelligence},
  volume={39},
  number={7},
  pages={6887--6895},
  year={2025}
}

@inproceedings{kim2025diffusehigh,
  title={Diffusehigh: Training-free progressive high-resolution image synthesis through structure guidance},
  author={Kim, Younghyun and Hwang, Geunmin and Zhang, Junyu and Park, Eunbyung},
  booktitle={Proceedings of the AAAI conference on artificial intelligence},
  volume={39},
  number={4},
  pages={4338--4346},
  year={2025}
}

@inproceedings{huang2024fouriscale,
  title={Fouriscale: A frequency perspective on training-free high-resolution image synthesis},
  author={Huang, Linjiang and Fang, Rongyao and Zhang, Aiping and Song, Guanglu and Liu, Si and Liu, Yu and Li, Hongsheng},
  booktitle={European conference on computer vision},
  pages={196--212},
  year={2024},
  organization={Springer}
}

@inproceedings{bu2025hiflow,
  title     = {{HiFlow}: Training-Free High-Resolution Image Generation with Flow-Aligned Guidance},
  author    = {Bu, Jiazi and Ling, Pengyang and Zhou, Yujie and Zhang, Pan and Wu, Tong and Dong, Xiaoyi and Zang, Yuhang and Cao, Yuhang and Lin, Dahua and Wang, Jiaqi},
  booktitle = {Advances in Neural Information Processing Systems},
  year      = {2025}
}

@inproceedings{dao2022flashattention,
  title={Flash{A}ttention: Fast and Memory-Efficient Exact Attention with {IO}-Awareness},
  author={Dao, Tri and Fu, Daniel Y. and Ermon, Stefano and Rudra, Atri and R{\'e}, Christopher},
  booktitle={Advances in Neural Information Processing Systems},
  year={2022}
}

@inproceedings{dao2023flashattention,
  title={Flash{A}ttention-2: Faster Attention with Better Parallelism and Work Partitioning},
  author={Dao, Tri},
  booktitle={International Conference on Learning Representations},
  year={2024}
}

@inproceedings{shah2024flashattention,
  title={FlashAttention-3: fast and accurate attention with asynchrony and low-precision},
  author={Shah, Jay and Bikshandi, Ganesh and Zhang, Ying and Thakkar, Vijay and Ramani, Pradeep and Dao, Tri},
  booktitle={Proceedings of the 38th International Conference on Neural Information Processing Systems},
  pages={68658--68685},
  year={2024}
}

@article{deng2024attention,
  title={Attention is naturally sparse with gaussian distributed input},
  author={Deng, Yichuan and Song, Zhao and Yang, Chiwun},
  journal={CoRR},
  year={2024}
}

@article{liu2022dynamic,
  title={Dynamic sparse attention for scalable transformer acceleration},
  author={Liu, Liu and Qu, Zheng and Chen, Zhaodong and Tu, Fengbin and Ding, Yufei and Xie, Yuan},
  journal={IEEE Transactions on Computers},
  volume={71},
  number={12},
  pages={3165--3178},
  year={2022},
  publisher={IEEE}
}

@article{beltagy2020longformer,
  title={Longformer: The long-document transformer},
  author={Beltagy, Iz and Peters, Matthew E and Cohan, Arman},
  journal={arXiv preprint arXiv:2004.05150},
  year={2020}
}

@inproceedings{liu2021swin,
  title={Swin transformer: Hierarchical vision transformer using shifted windows},
  author={Liu, Ze and Lin, Yutong and Cao, Yue and Hu, Han and Wei, Yixuan and Zhang, Zheng and Lin, Stephen and Guo, Baining},
  booktitle={Proceedings of the IEEE/CVF international conference on computer vision},
  pages={10012--10022},
  year={2021}
}

@article{han2024bridging,
  title={Bridging the divide: Reconsidering softmax and linear attention},
  author={Han, Dongchen and Pu, Yifan and Xia, Zhuofan and Han, Yizeng and Pan, Xuran and Li, Xiu and Lu, Jiwen and Song, Shiji and Huang, Gao},
  journal={Advances in Neural Information Processing Systems},
  volume={37},
  pages={79221--79245},
  year={2024}
}

@inproceedings{
lai2025flexprefill,
title={FlexPrefill: A Context-Aware Sparse Attention Mechanism for Efficient Long-Sequence Inference},
author={Xunhao Lai and Jianqiao Lu and Yao Luo and Yiyuan Ma and Xun Zhou},
booktitle={The Thirteenth International Conference on Learning Representations},
year={2025},
}

@inproceedings{zhang2025spargeattn,
  title={Spargeattn: Accurate sparse attention accelerating any model inference},
  author={Zhang, Jintao and Xiang, Chendong and Huang, Haofeng and Wei, Jia and Xi, Haocheng and Zhu, Jun and Chen, Jianfei},
  booktitle={International Conference on Machine Learning},
  year={2025}
}

@article{yang2025sparse,
  title={Sparse VideoGen2: Accelerate Video Generation with Sparse Attention via Semantic-Aware Permutation},
  author={Yang, Shuo and Xi, Haocheng and Zhao, Yilong and Li, Muyang and Zhang, Jintao and Cai, Han and Lin, Yujun and Li, Xiuyu and Xu, Chenfeng and Peng, Kelly and others},
  journal={arXiv preprint arXiv:2505.18875},
  year={2025}
}

@inproceedings{xi2025sparse,
  title={Sparse videogen: Accelerating video diffusion transformers with spatial-temporal sparsity},
  author={Xi, Haocheng and Yang, Shuo and Zhao, Yilong and Xu, Chenfeng and Li, Muyang and Li, Xiuyu and Lin, Yujun and Cai, Han and Zhang, Jintao and Li, Dacheng and others},
  booktitle={Forty-second International Conference on Machine Learning},
  year={2025}
}

@article{yuan2024ditfastattn,
  title={Ditfastattn: Attention compression for diffusion transformer models},
  author={Yuan, Zhihang and Zhang, Hanling and Pu, Lu and Ning, Xuefei and Zhang, Linfeng and Zhao, Tianchen and Yan, Shengen and Dai, Guohao and Wang, Yu},
  journal={Advances in Neural Information Processing Systems},
  volume={37},
  pages={1196--1219},
  year={2024}
}

@inproceedings{zhang2025ditfastattnv2,
  title={DiTFastAttnV2: Head-wise Attention Compression for Multi-Modality Diffusion Transformers},
  author={Zhang, Hanling and Su, Rundong and Yuan, Zhihang and Chen, Pengtao and Shen, Mingzhu and Fan, Yibo and Yan, Shengen and Dai, Guohao and Wang, Yu},
  booktitle={Proceedings of the IEEE/CVF International Conference on Computer Vision},
  pages={16399--16409},
  year={2025}
}

@inproceedings{xu2025xattention,
  title     = {XAttention: Block Sparse Attention with Antidiagonal Scoring},
  author    = {Xu, Ruyi and Xiao, Guangxuan and Huang, Haofeng and Guo, Junxian and Han, Song},
  booktitle = {Proceedings of the 42nd International Conference on Machine Learning},
  year      = {2025}
}

@inproceedings{
meng2025polaformer,
title={PolaFormer: Polarity-aware Linear Attention for Vision Transformers},
author={Weikang Meng and Yadan Luo and Xin Li and Dongmei Jiang and Zheng Zhang},
booktitle={The Thirteenth International Conference on Learning Representations},
year={2025},
}

@inproceedings{zhang2025layercraft,
  title     = {{LayerCraft}: Enhancing Text-to-Image Generation with CoT Reasoning and Layered Object Integration},
  author    = {Zhang, Yuyao and Li, Jinghao and Tai, Yu-Wing},
  booktitle = {Advances in Neural Information Processing Systems},
  year      = {2025}
}

@article{gao2025seedream,
  title={Seedream 3.0 technical report},
  author={Gao, Yu and Gong, Lixue and Guo, Qiushan and Hou, Xiaoxia and Lai, Zhichao and Li, Fanshi and Li, Liang and Lian, Xiaochen and Liao, Chao and Liu, Liyang and others},
  journal={arXiv preprint arXiv:2504.11346},
  year={2025}
}

@article{su2024roformer,
  title={Roformer: Enhanced transformer with rotary position embedding},
  author={Su, Jianlin and Ahmed, Murtadha and Lu, Yu and Pan, Shengfeng and Bo, Wen and Liu, Yunfeng},
  journal={Neurocomputing},
  volume={568},
  pages={127063},
  year={2024},
  publisher={Elsevier}
}

@inproceedings{hu2022lora,
  title     = {{LoRA}: Low-Rank Adaptation of Large Language Models},
  author    = {Hu, Edward J. and Shen, Yelong and Wallis, Phillip and Allen-Zhu, Zeyuan and Li, Yuanzhi and Wang, Shean and Wang, Lu and Chen, Weizhu},
  booktitle = {International Conference on Learning Representations},
  year      = {2022}
}

@inproceedings{zhang2025sageattention,
  title={SageAttention: Accurate 8-Bit Attention for Plug-and-play Inference Acceleration}, 
  author={Zhang, Jintao and Wei, Jia and Zhang, Pengle and Zhu, Jun and Chen, Jianfei},
  booktitle={International Conference on Learning Representations},
  year={2025}
}

@inproceedings{zhang2024sageattention2,
  title={Sageattention2: Efficient attention with thorough outlier smoothing and per-thread int4 quantization},
  author={Zhang, Jintao and Huang, Haofeng and Zhang, Pengle and Wei, Jia and Zhu, Jun and Chen, Jianfei},
  booktitle={International Conference on Machine Learning},
  year={2025}
}

@inproceedings{zhang2021designing,
    title={Designing a Practical Degradation Model for Deep Blind Image Super-Resolution},
    author={Zhang, Kai and Liang, Jingyun and Van Gool, Luc and Timofte, Radu},
    booktitle={IEEE International Conference on Computer Vision},
    pages={4791--4800},
    year={2021}
}

@inproceedings{du2024demofusion,
  title={DemoFusion: Democratising High-Resolution Image Generation With No \$\$\$},
  author={Du, Ruoyi and Chang, Dongliang and Hospedales, Timothy and Song, Yi-Zhe and Ma, Zhanyu},
  booktitle={Proceedings of the IEEE/CVF Conference on Computer Vision and Pattern Recognition},
  year={2024}
}

@inproceedings{radford2021learning,
  title={Learning transferable visual models from natural language supervision},
  author={Radford, Alec and Kim, Jong Wook and Hallacy, Chris and Ramesh, Aditya and Goh, Gabriel and Agarwal, Sandhini and Sastry, Girish and Askell, Amanda and Mishkin, Pamela and Clark, Jack and others},
  booktitle={International conference on machine learning},
  pages={8748--8763},
  year={2021},
  organization={PmLR}
}

@article{salimans2016improved,
  title={Improved techniques for training gans},
  author={Salimans, Tim and Goodfellow, Ian and Zaremba, Wojciech and Cheung, Vicki and Radford, Alec and Chen, Xi},
  journal={Advances in neural information processing systems},
  volume={29},
  year={2016}
}

@article{heusel2017gans,
  title={Gans trained by a two time-scale update rule converge to a local nash equilibrium},
  author={Heusel, Martin and Ramsauer, Hubert and Unterthiner, Thomas and Nessler, Bernhard and Hochreiter, Sepp},
  journal={Advances in neural information processing systems},
  volume={30},
  year={2017}
}

@article{schuhmann2022laion,
  title={Laion-5b: An open large-scale dataset for training next generation image-text models},
  author={Schuhmann, Christoph and Beaumont, Romain and Vencu, Richard and Gordon, Cade and Wightman, Ross and Cherti, Mehdi and Coombes, Theo and Katta, Aarush and Mullis, Clayton and Wortsman, Mitchell and others},
  journal={Advances in neural information processing systems},
  volume={35},
  pages={25278--25294},
  year={2022}
}

@article{ghosh2023geneval,
  title={Geneval: An object-focused framework for evaluating text-to-image alignment},
  author={Ghosh, Dhruba and Hajishirzi, Hannaneh and Schmidt, Ludwig},
  journal={Advances in Neural Information Processing Systems},
  volume={36},
  pages={52132--52152},
  year={2023}
}

@inproceedings{zhang2402hedgehog,
  title={The Hedgehog \& the Porcupine: Expressive Linear Attentions with Softmax Mimicry},
  author={Zhang, Michael and Bhatia, Kush and Kumbong, Hermann and Re, Christopher},
  booktitle={The Twelfth International Conference on Learning Representations},
year={2024}
}

@inproceedings{zhu2025dig,
  title={Dig: Scalable and efficient diffusion models with gated linear attention},
  author={Zhu, Lianghui and Huang, Zilong and Liao, Bencheng and Liew, Jun Hao and Yan, Hanshu and Feng, Jiashi and Wang, Xinggang},
  booktitle={Proceedings of the Computer Vision and Pattern Recognition Conference},
  pages={7664--7674},
  year={2025}
}

@inproceedings{zhang2025hidiffusion,
  title={HiDiffusion: Unlocking Higher-Resolution Creativity and Efficiency in Pretrained Diffusion Models},
  author={Zhang, Shen and Chen, Zhaowei and Zhao, Zhenyu and Chen, Yuhao and Tang, Yao and Liang, Jiajun},
  booktitle={European Conference on Computer Vision},
  pages={145--161},
  year={2025},
  organization={Springer}
}

@inproceedings{zhangfrecas,
  title={FreCaS: Efficient Higher-Resolution Image Generation via Frequency-aware Cascaded Sampling},
  author={Zhang, Zhengqiang and Li, Ruihuang and Zhang, Lei},
  booktitle={The Thirteenth International Conference on Learning Representations},
year={2025}
}

@inproceedings{liu2024clear,
  title     = {{CLEAR}: Conv-like Linearization Revs Pre-Trained Diffusion Transformers up},
  author    = {Liu, Songhua and Tan, Zhenxiong and Wang, Xinchao},
  booktitle = {Advances in Neural Information Processing Systems},
  year      = {2025}
}

@article{ren2025grat,
  title={Grouping First, Attending Smartly: Training-Free Acceleration for Diffusion Transformers},
  author={Ren, Sucheng and Yu, Qihang and He, Ju and Yuille, Alan and Chen, Liang-Chieh},
  journal={arXiv preprint arXiv:2505.14687},
  year={2025}
}

@inproceedings{zhang2025sta,
  title={Fast Video Generation with Sliding Tile Attention},
  author={Zhang, Peiyuan and Chen, Yongqi and Su, Runlong and Ding, Hangliang and Stoica, Ion and Liu, Zhengzhong and Zhang, Hao},
  booktitle={Forty-second International Conference on Machine Learning},
year={2025}
}

@inproceedings{wang2022exploring,
    author = {Wang, Jianyi and Chan, Kelvin CK and Loy, Chen Change},
    title = {Exploring CLIP for Assessing the Look and Feel of Images},
    booktitle = {AAAI},
    year = {2023}
}

@article{ye2025ultraflux,
      title={UltraFlux: Data-Model Co-Design for High-quality Native 4K Text-to-Image Generation across Diverse Aspect Ratios}, 
      author={Tian Ye and Song Fei and Lei Zhu},
      year={2025},
      journal={arXiv preprint arXiv:2511.18050}, 
}

@misc{flux-2-2025,
    author       = {{Black Forest Labs}},
    title        = {{FLUX.2: Frontier Visual Intelligence}},
    year         = {2025},
    howpublished = {\url{https://bfl.ai/blog/flux-2}},
    note         = {Accessed: 2026-06-28},
}

@inproceedings{
      meng2022sdedit,
      title={{SDE}dit: Guided Image Synthesis and Editing with Stochastic Differential Equations},
      author={Chenlin Meng and Yutong He and Yang Song and Jiaming Song and Jiajun Wu and Jun-Yan Zhu and Stefano Ermon},
      booktitle={International Conference on Learning Representations},
      year={2022},
}

@article{he2025ragsr,
  title={Ragsr: Regional attention guided diffusion for image super-resolution},
  author={He, Haodong and Bai, Yancheng and Lan, Rui and Duan, Xu and Sun, Lei and Chu, Xiangxiang and Xia, Gui-Song},
  journal={arXiv preprint arXiv:2508.16158},
  year={2025}
}

@inproceedings{dong2025can,
  title={Can We Achieve Efficient Diffusion without Self-Attention? Distilling Self-Attention into Convolutions},
  author={Dong, Ziyi and Zhou, Chengxing and Deng, Weijian and Wei, Pengxu and Ji, Xiangyang and Lin, Liang},
  booktitle={Proceedings of the IEEE/CVF International Conference on Computer Vision},
  pages={17401--17410},
  year={2025}
}

@inproceedings{hu2026ultragen,
  title={UltraGen: High-Resolution Video Generation with Hierarchical Attention},
  author={Hu, Teng and Zhang, Jiangning and Su, Zihan and Yi, Ran},
  booktitle={Proceedings of the AAAI Conference on Artificial Intelligence},
  volume={40},
  number={6},
  pages={4923--4931},
  year={2026}
}

@inproceedings{kwon2026reviving,
  title={Reviving ConvNeXt for Efficient Convolutional Diffusion Models},
  author={Kwon, Taesung and Bianchi, Lorenzo and Wittke, Lennart and Watine, Felix and Carrara, Fabio and Ye, Jong Chul and Weber, Romann and Azevedo, Vinicius},
  booktitle={Proceedings of the IEEE/CVF Conference on Computer Vision and Pattern Recognition},
  pages={43675--43685},
  year={2026}
}

@inproceedings{zhang2026hieredit,
  title={HierEdit: Region-Aware Hierarchical Diffusion for Efficient High-Resolution Editing},
  author={Zhang, Yuyao and Huang-Menders, Alexander and Tai, Yu-Wing},
  booktitle={Proceedings of the IEEE/CVF Conference on Computer Vision and Pattern Recognition},
  pages={43546--43557},
  year={2026}
}
\newpage
\section*{\LARGE{Supplementary Materials}}

\section{Ultra-High-Res Gallery ($>4K$)}
\begin{figure}
    \centering
    \includegraphics[width=.8\linewidth]{Figures/8k-gallery/8k-001.jpg}
    \includegraphics[width=\linewidth]{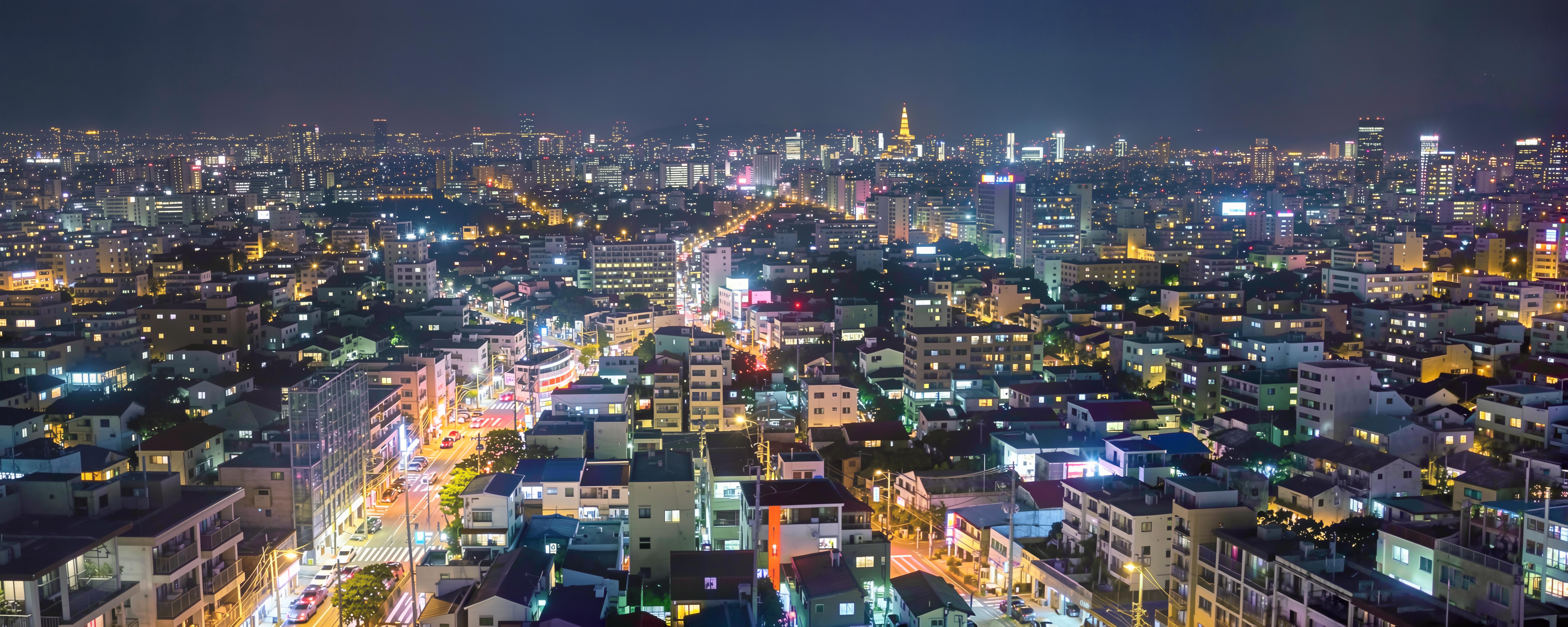}
    \caption{One $8192\times8192$ and One $4096\times10240$ images}
    \label{fig:8k}
\end{figure}

In Figure~\ref{fig:8k}, we show one $8192\times8192$ and one $4096\times10240$ images, more demos are in the Zip file in the supplementary materials.

\section{4K and 2K Galleries}
\begin{figure}[htbp]
    \centering
    \includegraphics[width=0.49\linewidth]{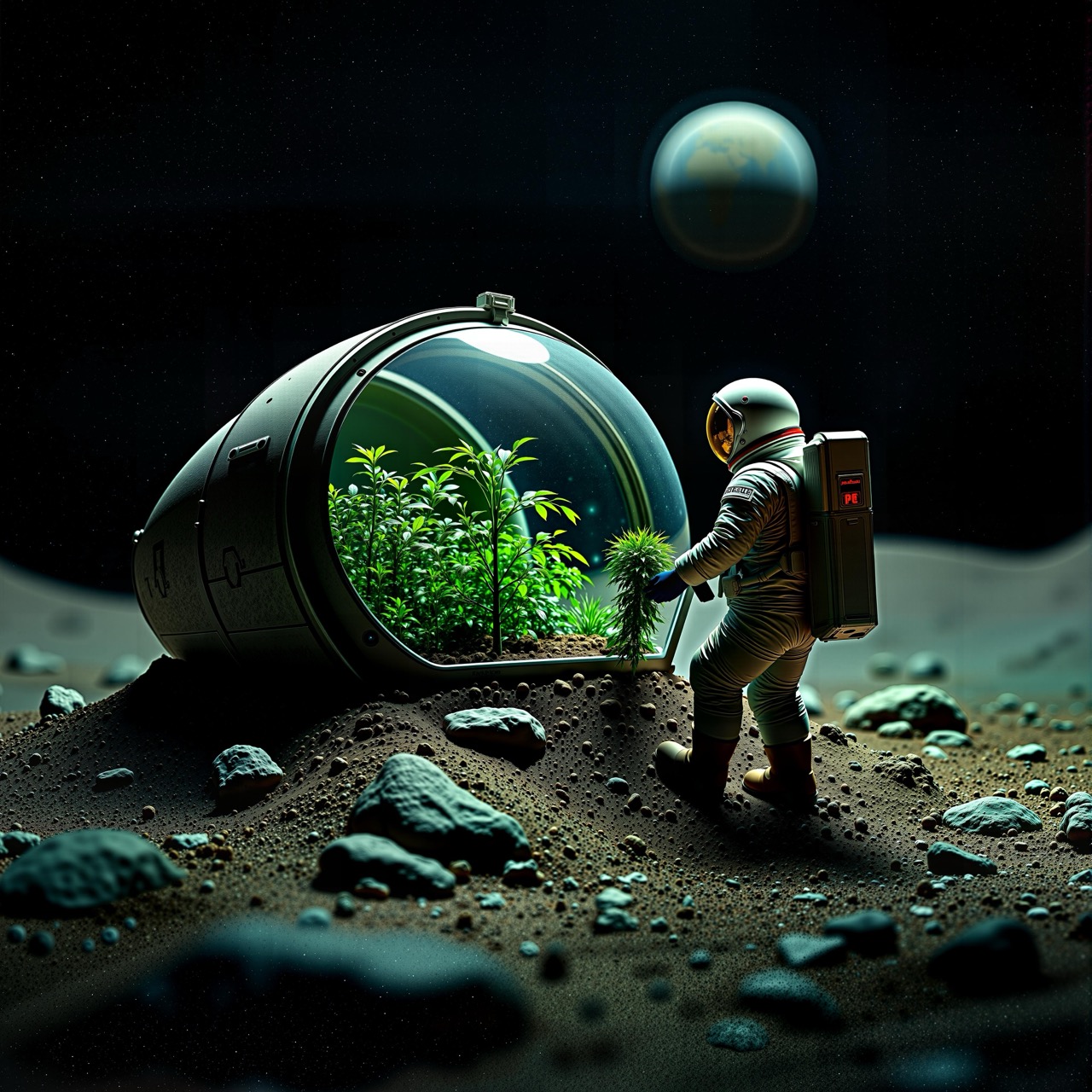}
    \includegraphics[width=0.49\linewidth]{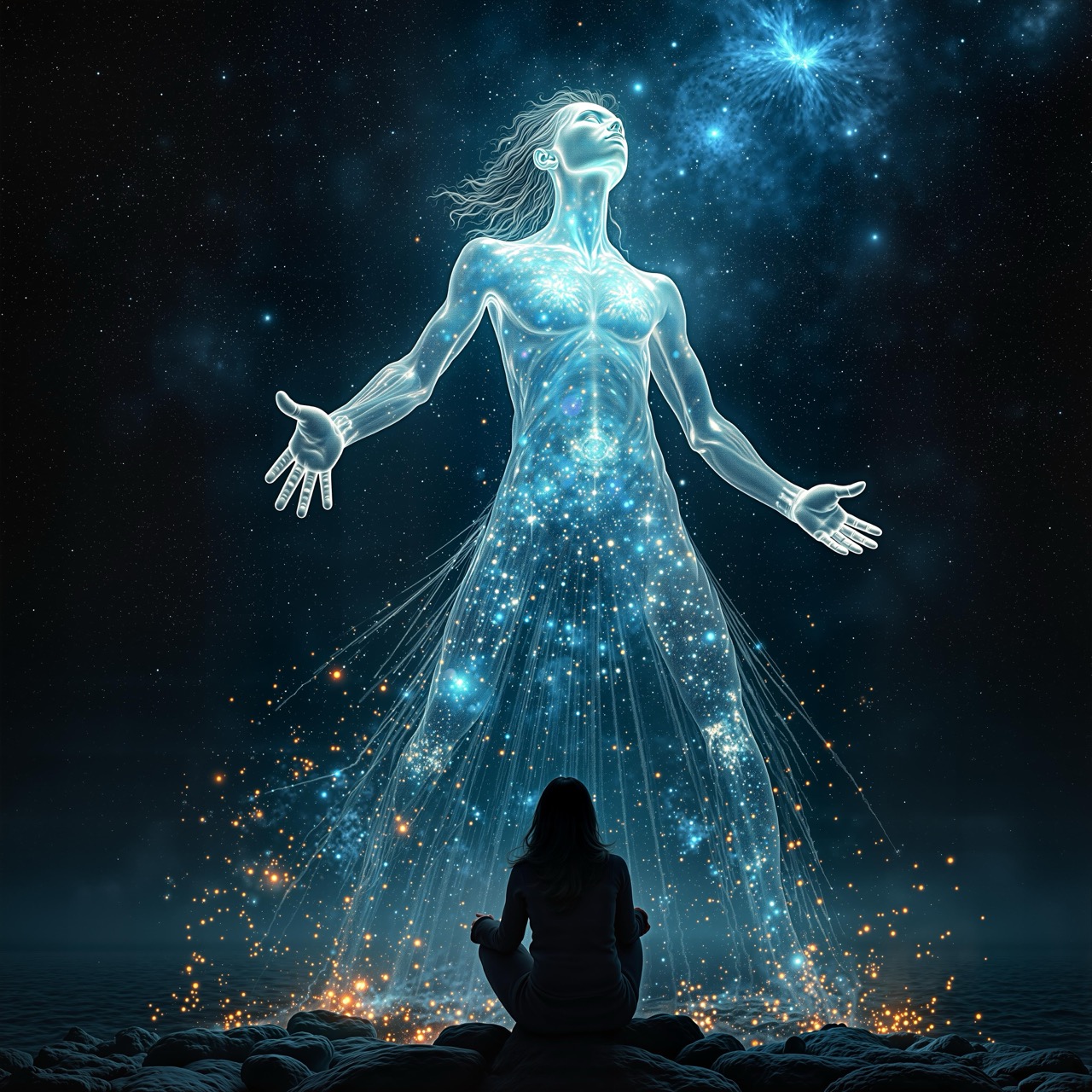}
    \includegraphics[width=0.49\linewidth]{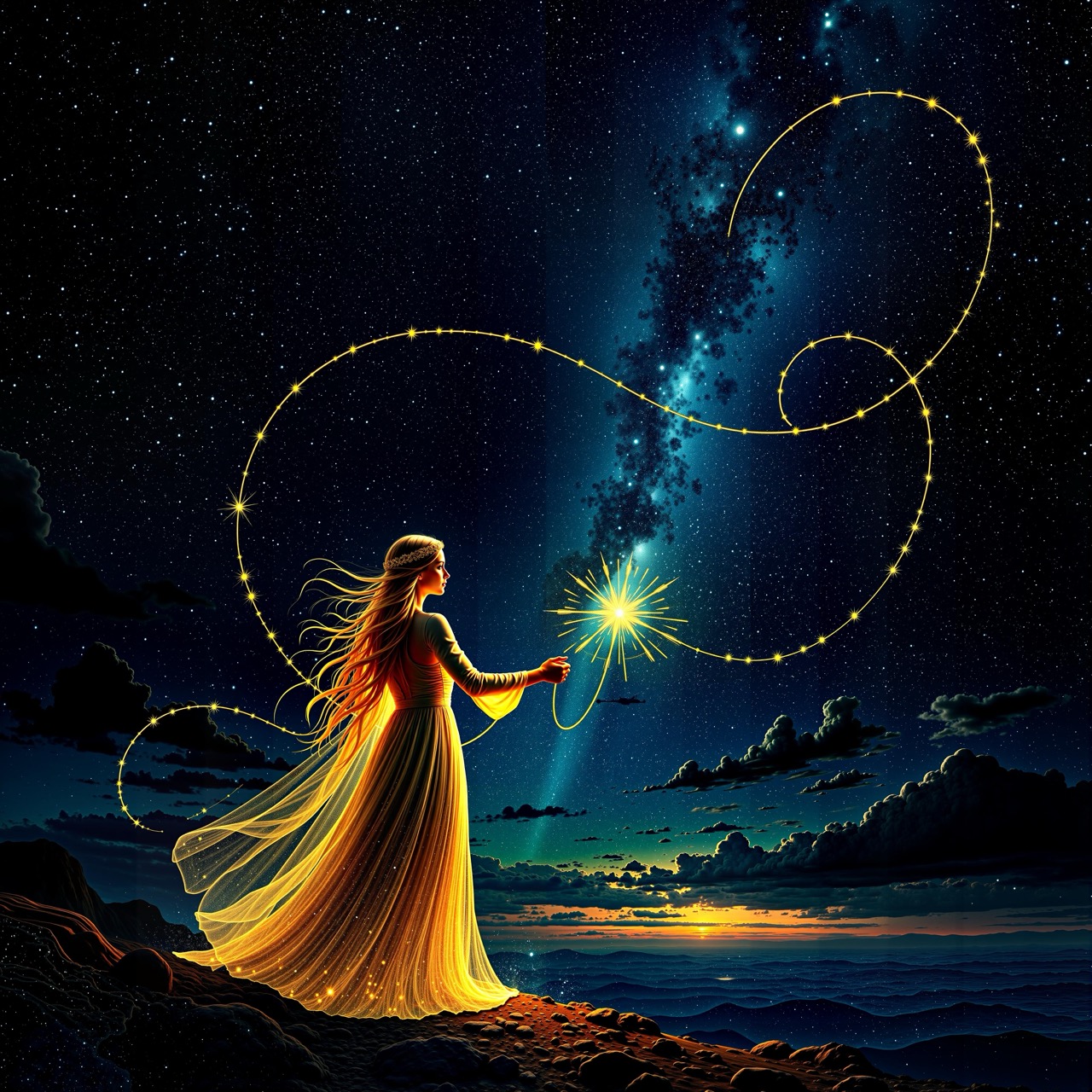}
    \includegraphics[width=0.49\linewidth]{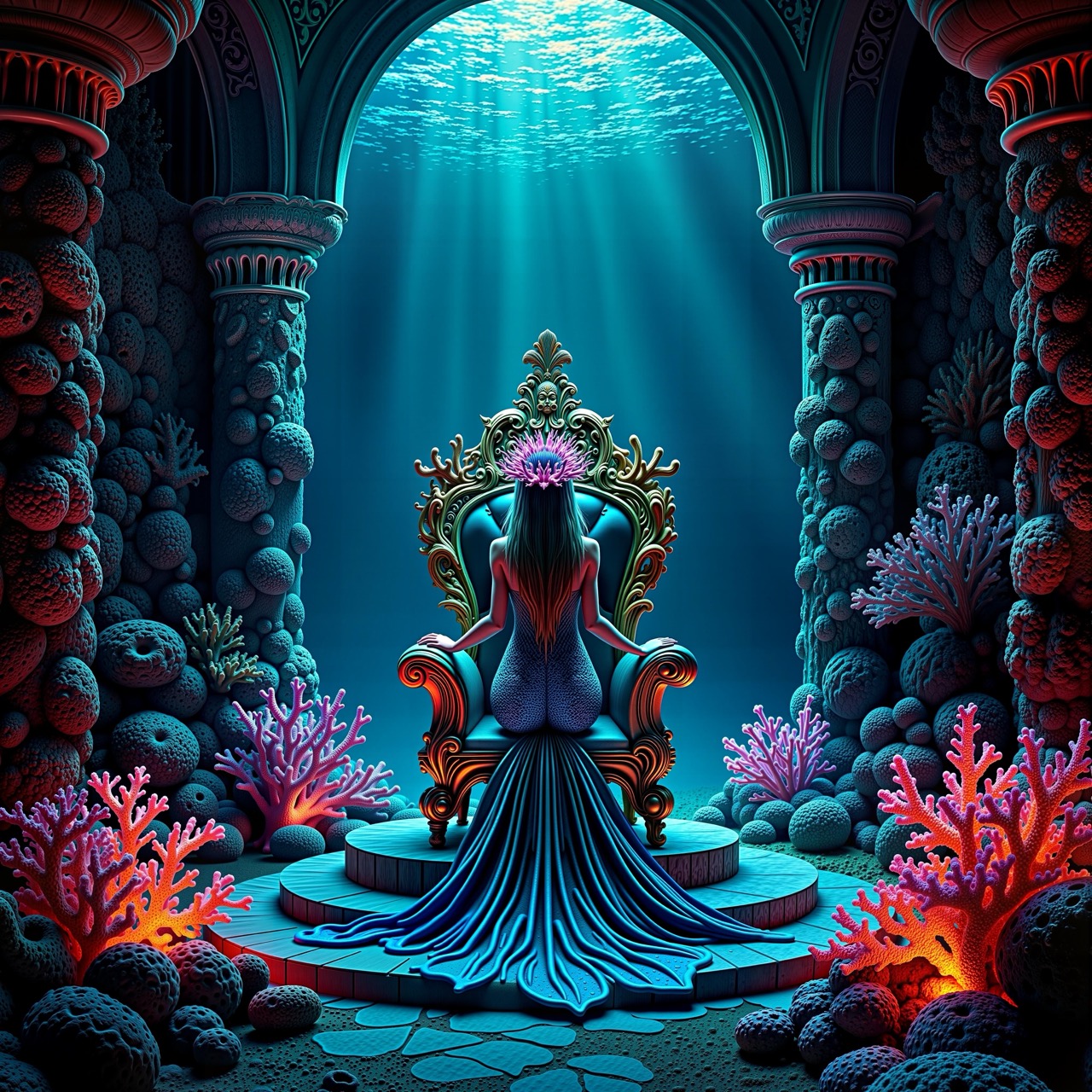}
    \includegraphics[width=0.49\linewidth]{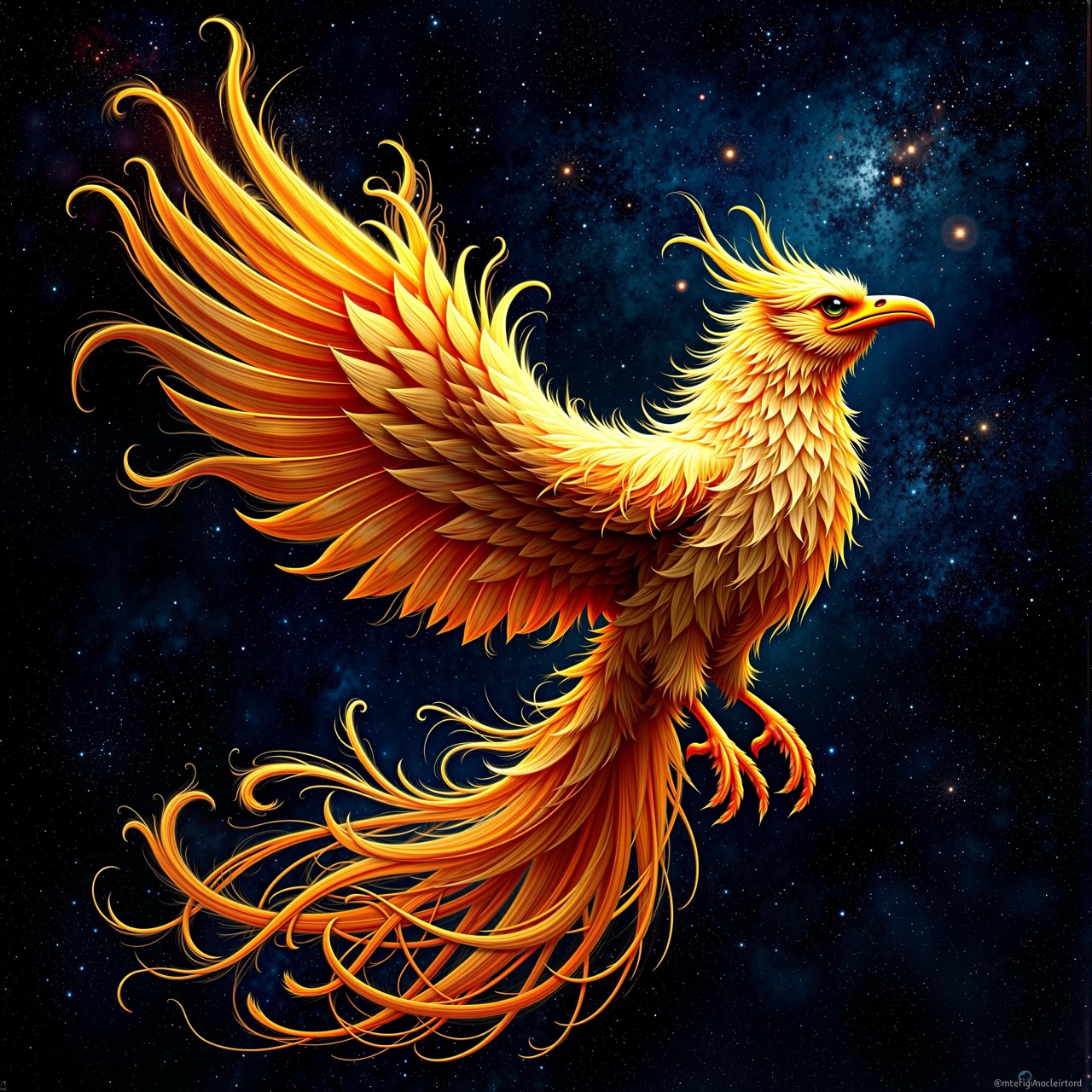}
    \includegraphics[width=0.49\linewidth]{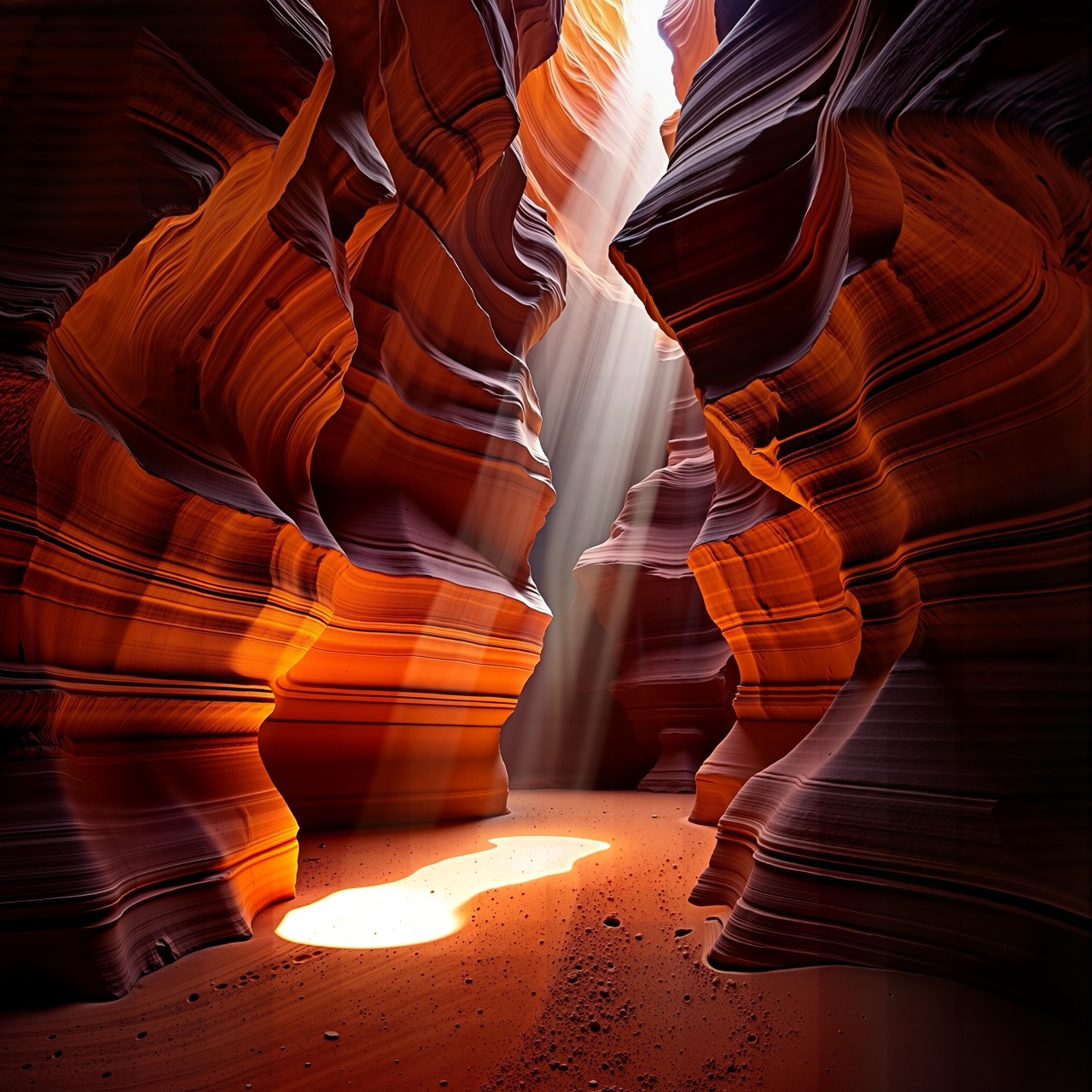}
    \caption{More 4K results}
    \label{fig:placeholder-5}
\end{figure}

\begin{figure}[htbp]
    \centering
    \includegraphics[width=0.49\linewidth]{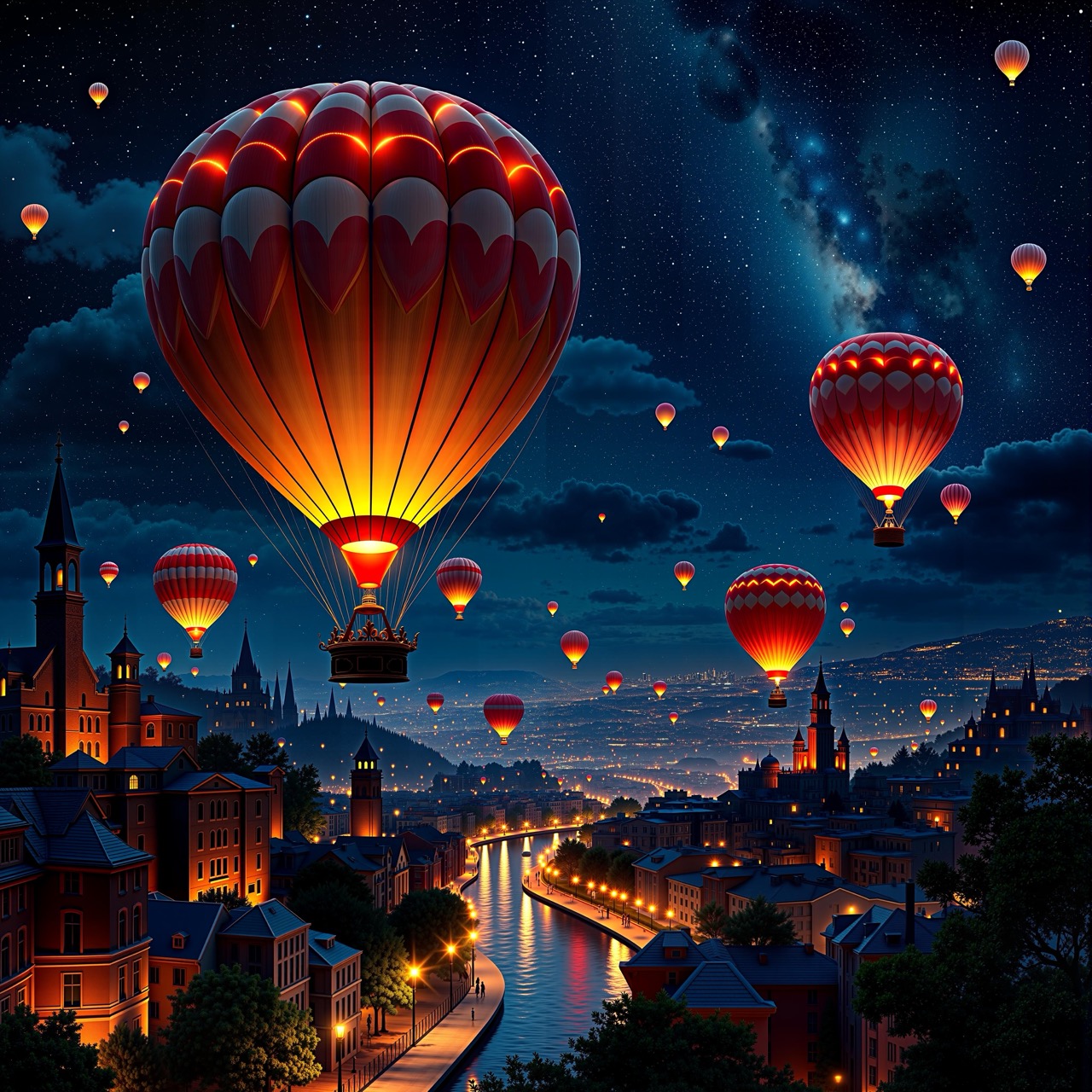}
    \includegraphics[width=0.49\linewidth]{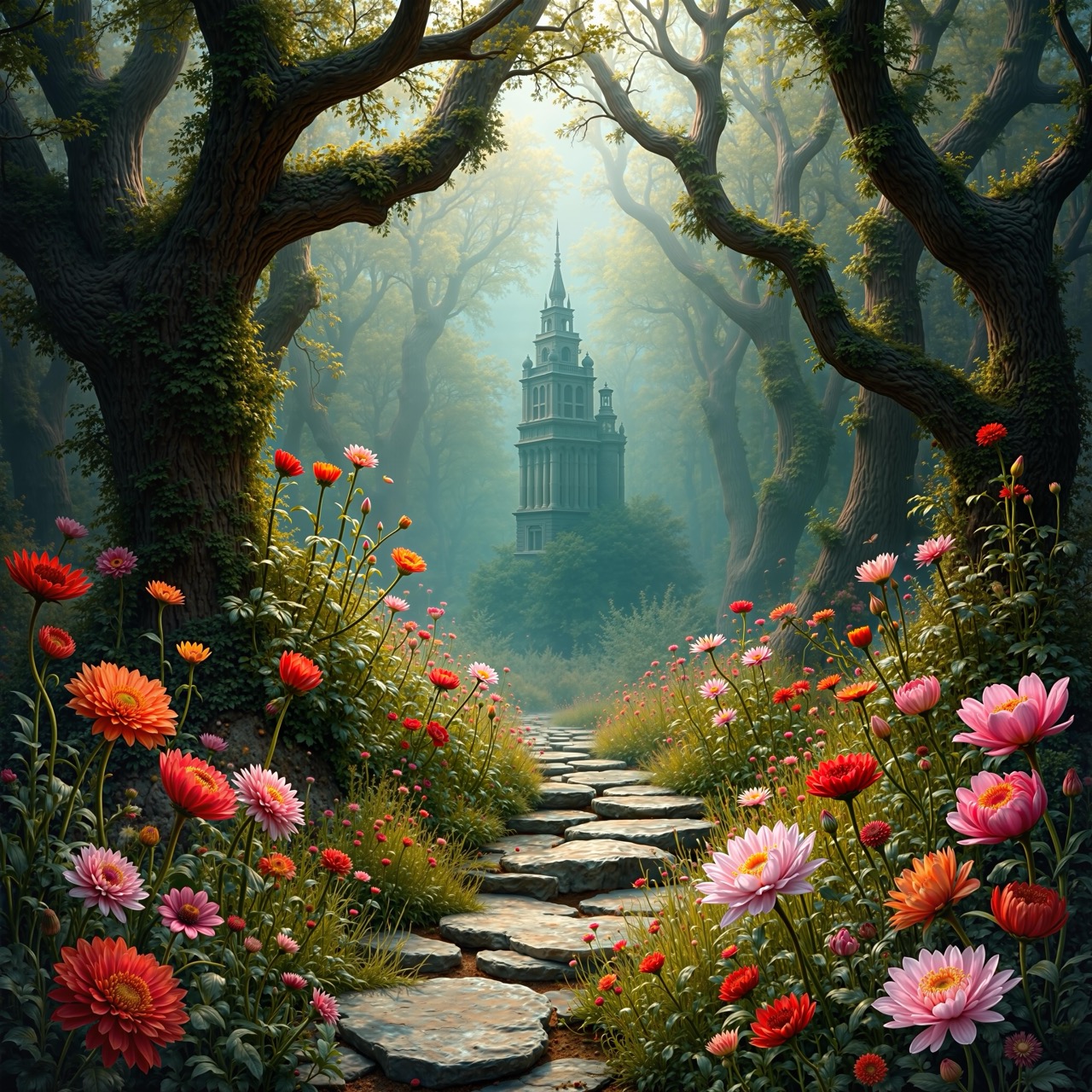}
    \includegraphics[width=0.49\linewidth]{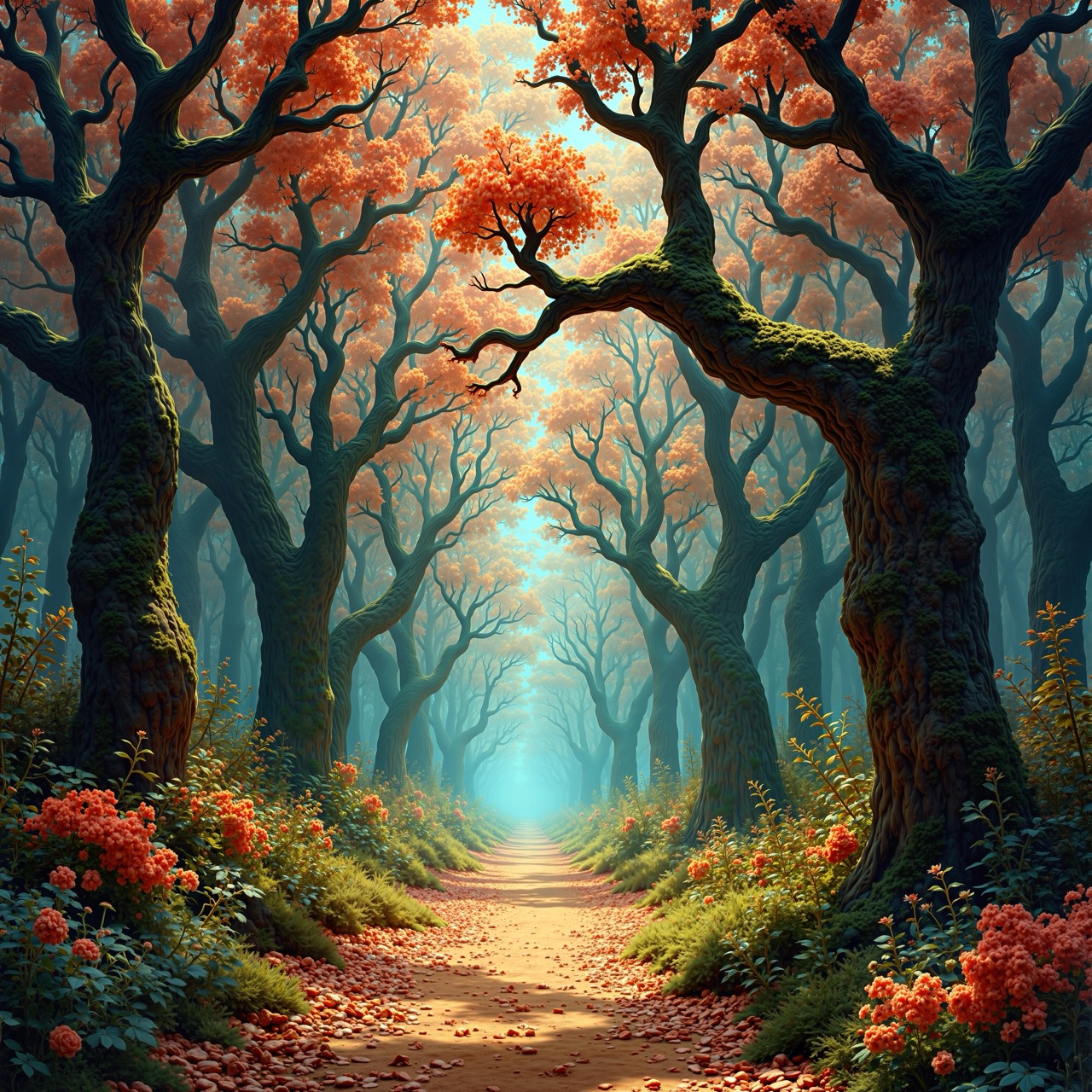}
    \includegraphics[width=0.49\linewidth]{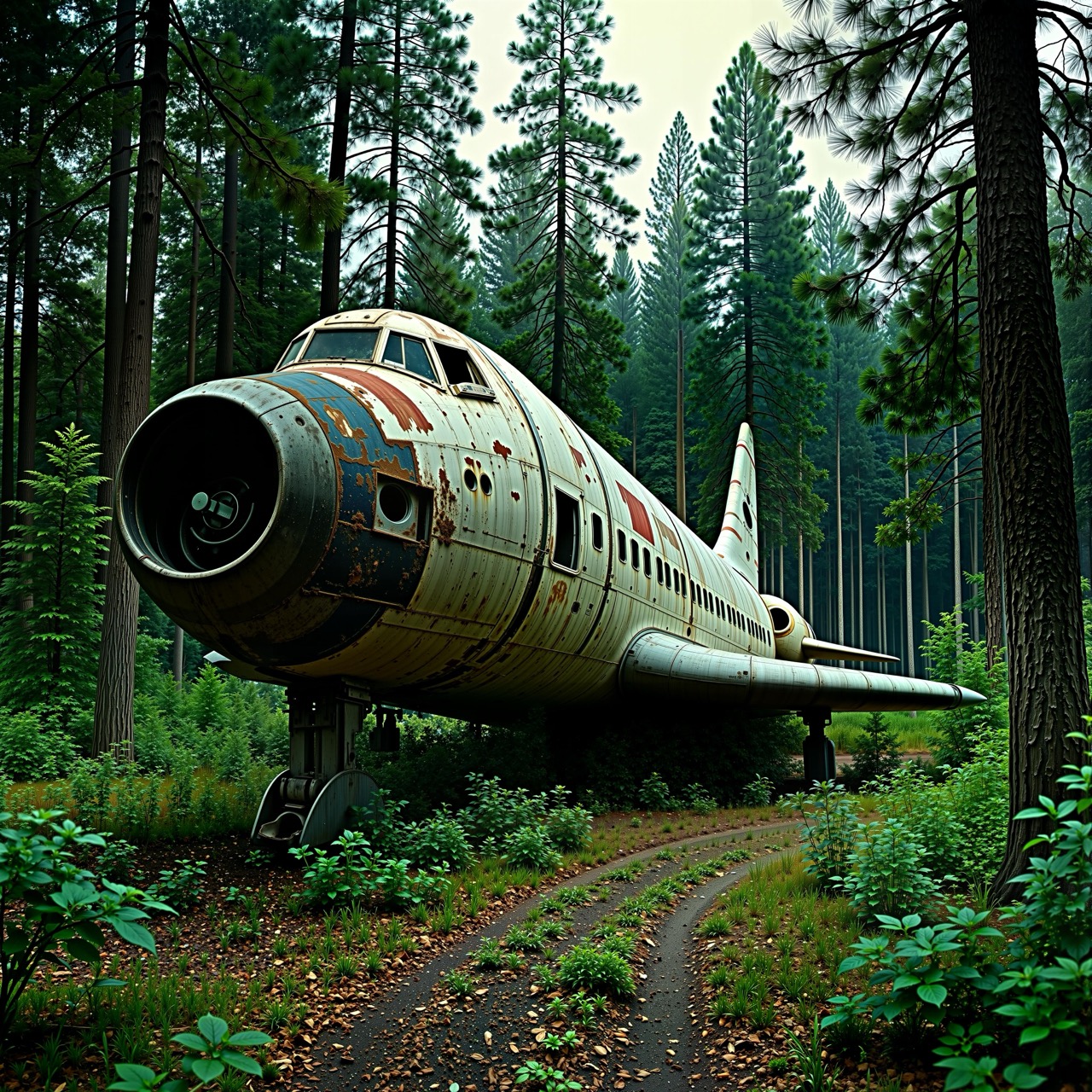}
    \includegraphics[width=0.49\linewidth]{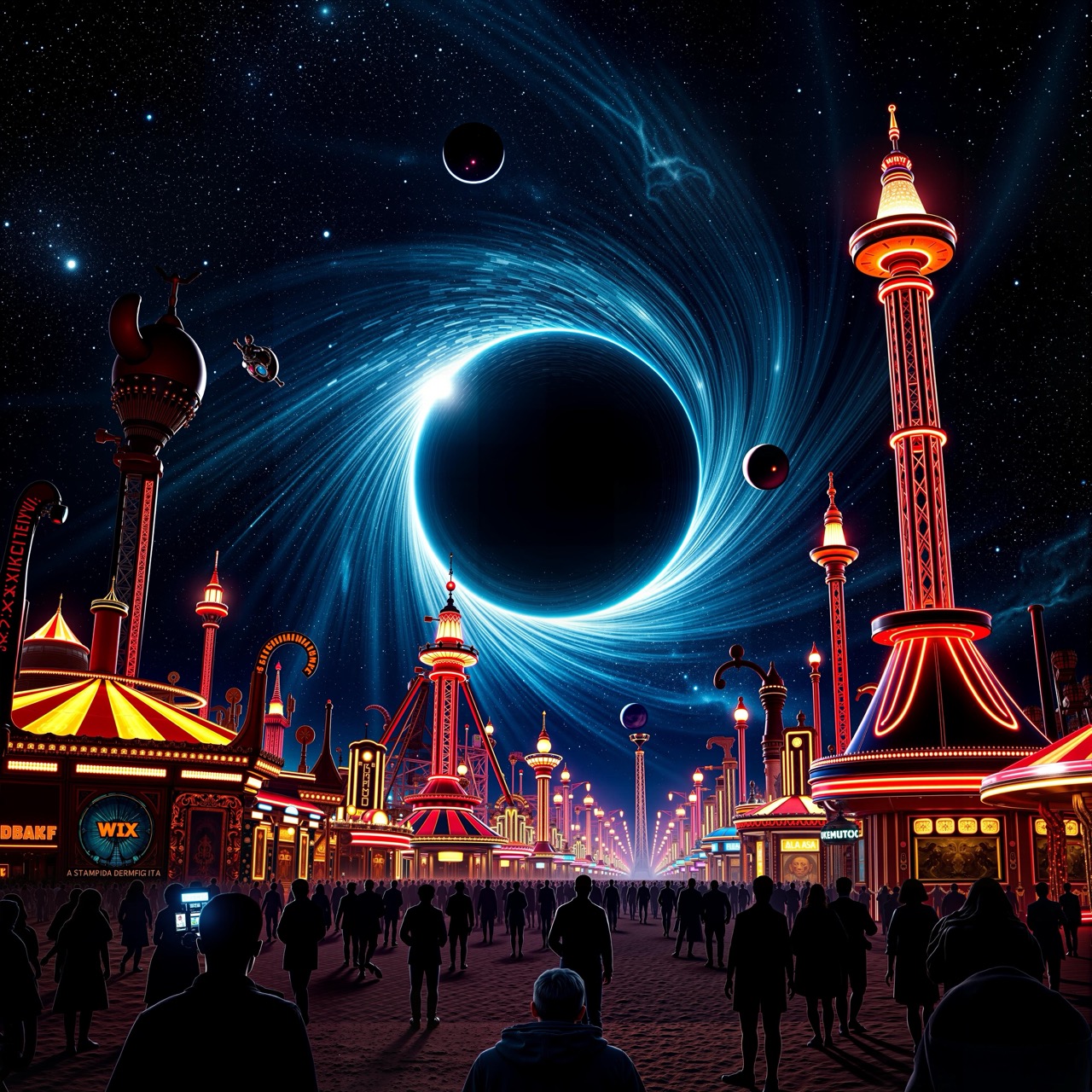}
    \includegraphics[width=0.49\linewidth]{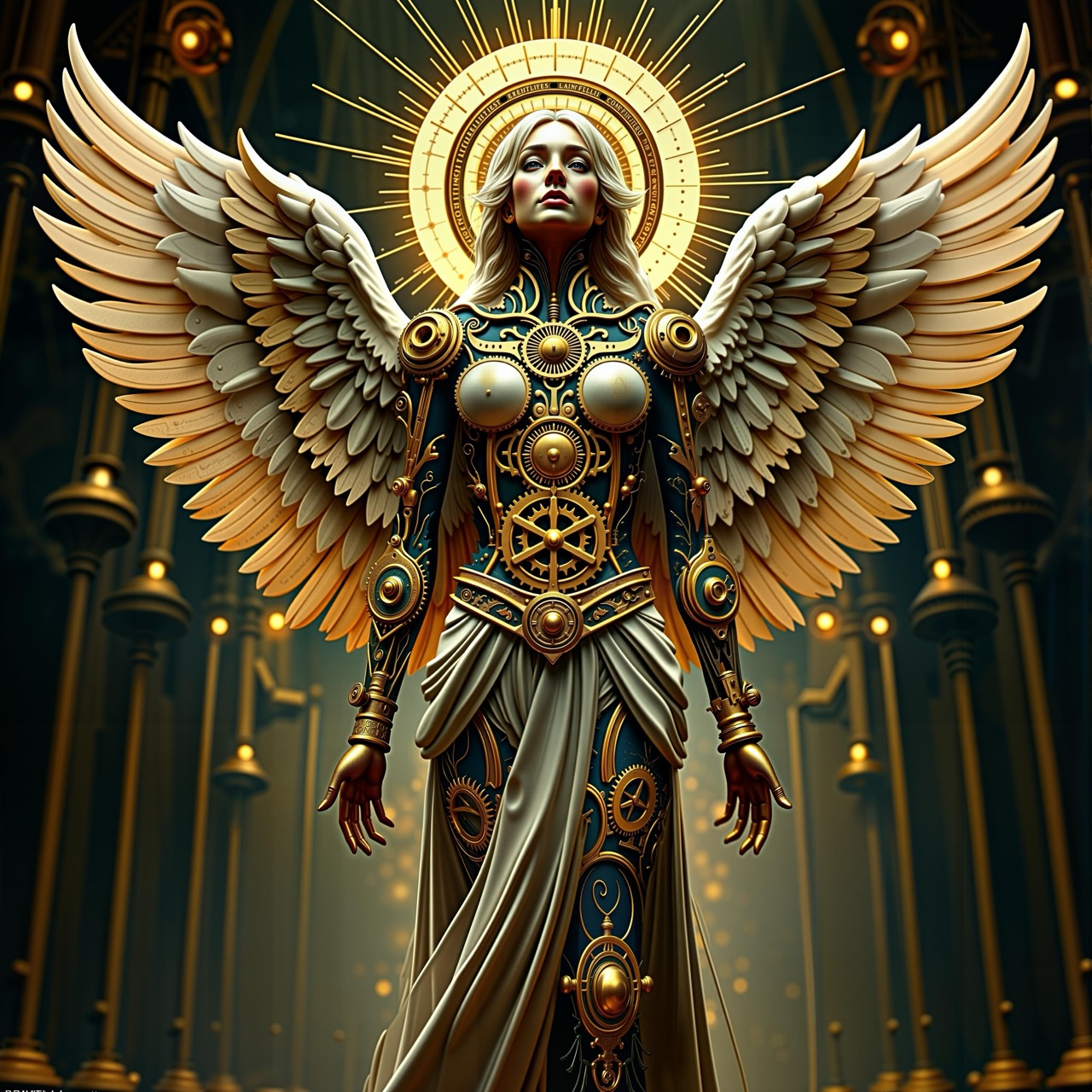}
    \caption{More 4K results}
    \label{fig:placeholder-6}
\end{figure}

\begin{figure}[htbp]
    \centering
    \includegraphics[width=0.49\linewidth]{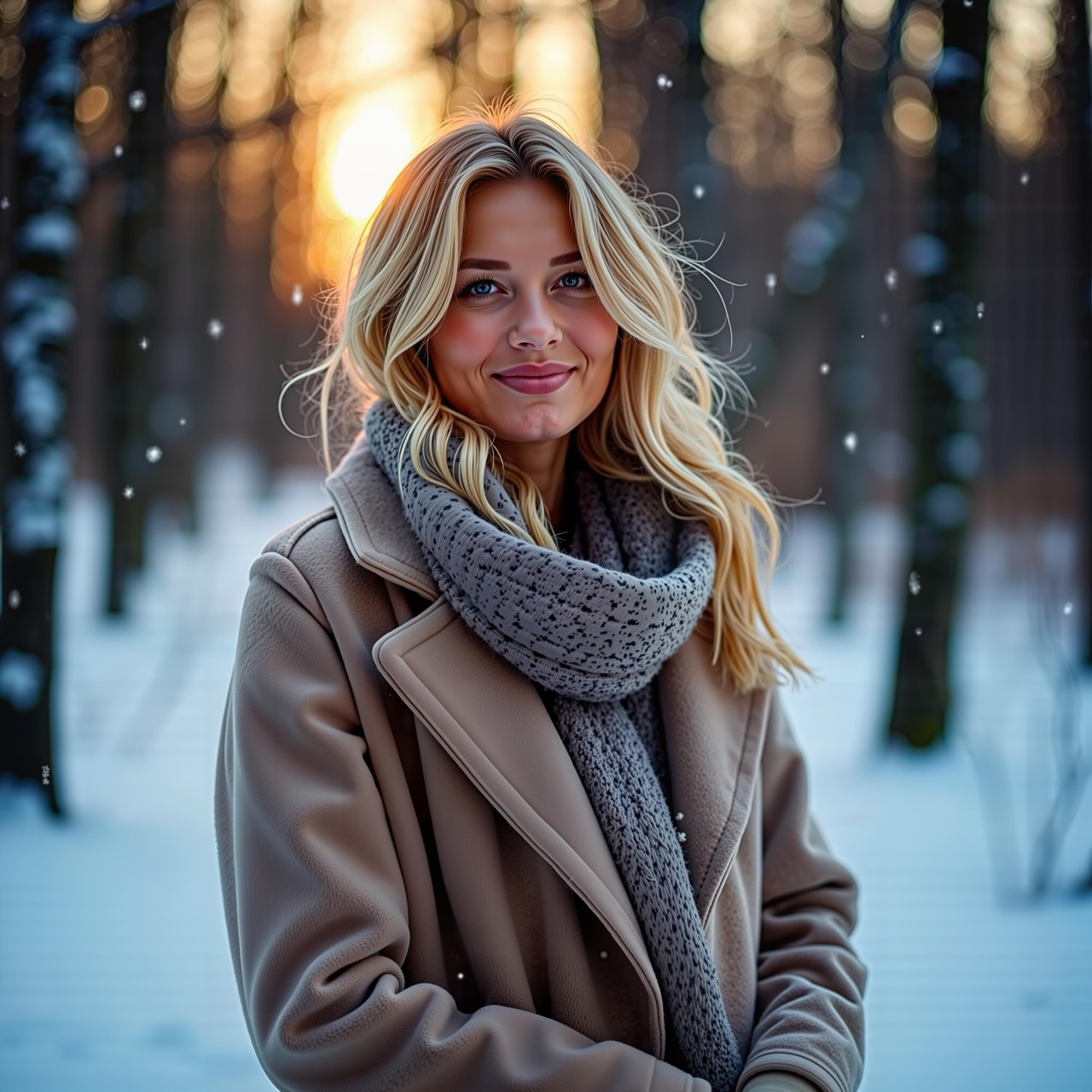}
    \includegraphics[width=0.49\linewidth]{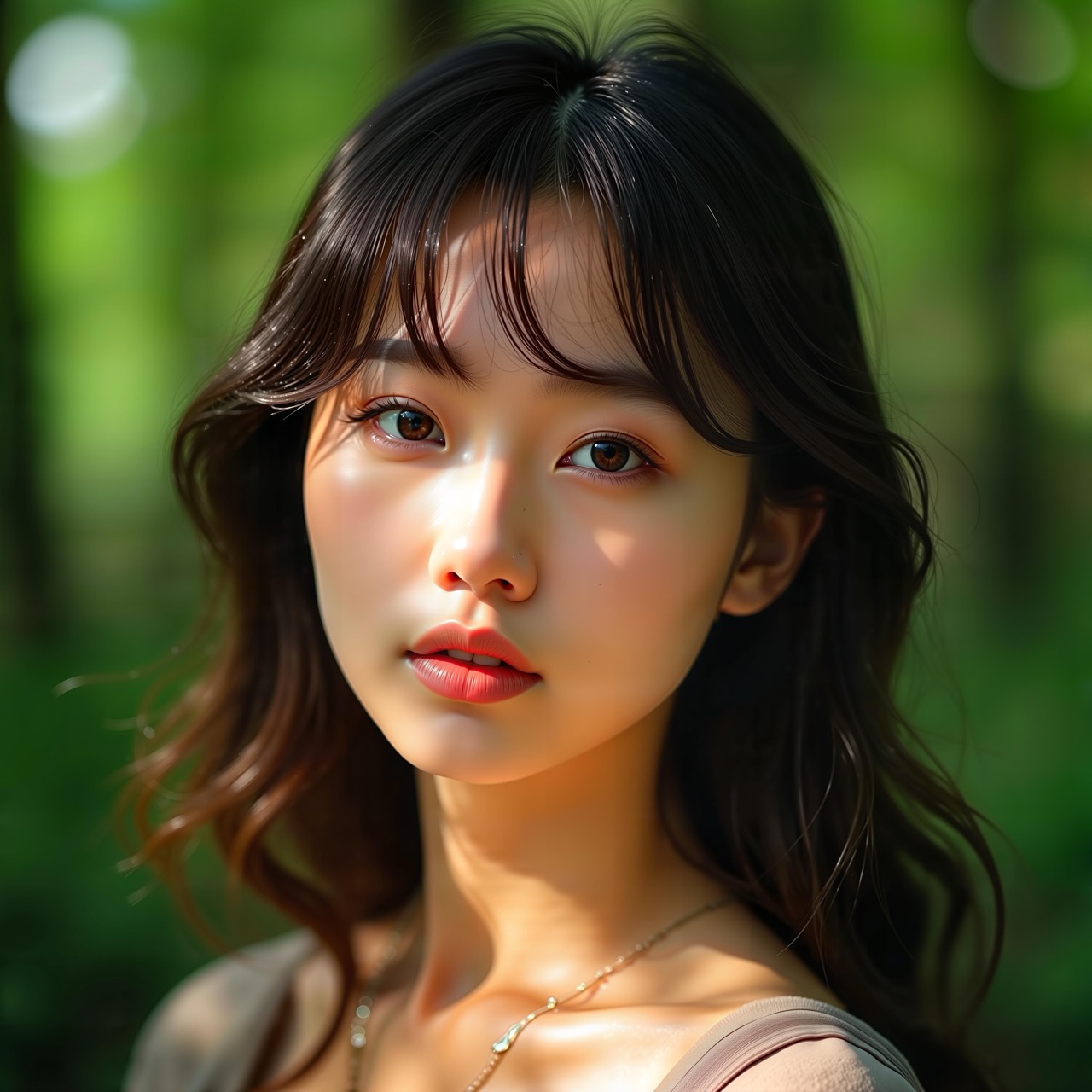}
    \includegraphics[width=0.49\linewidth]{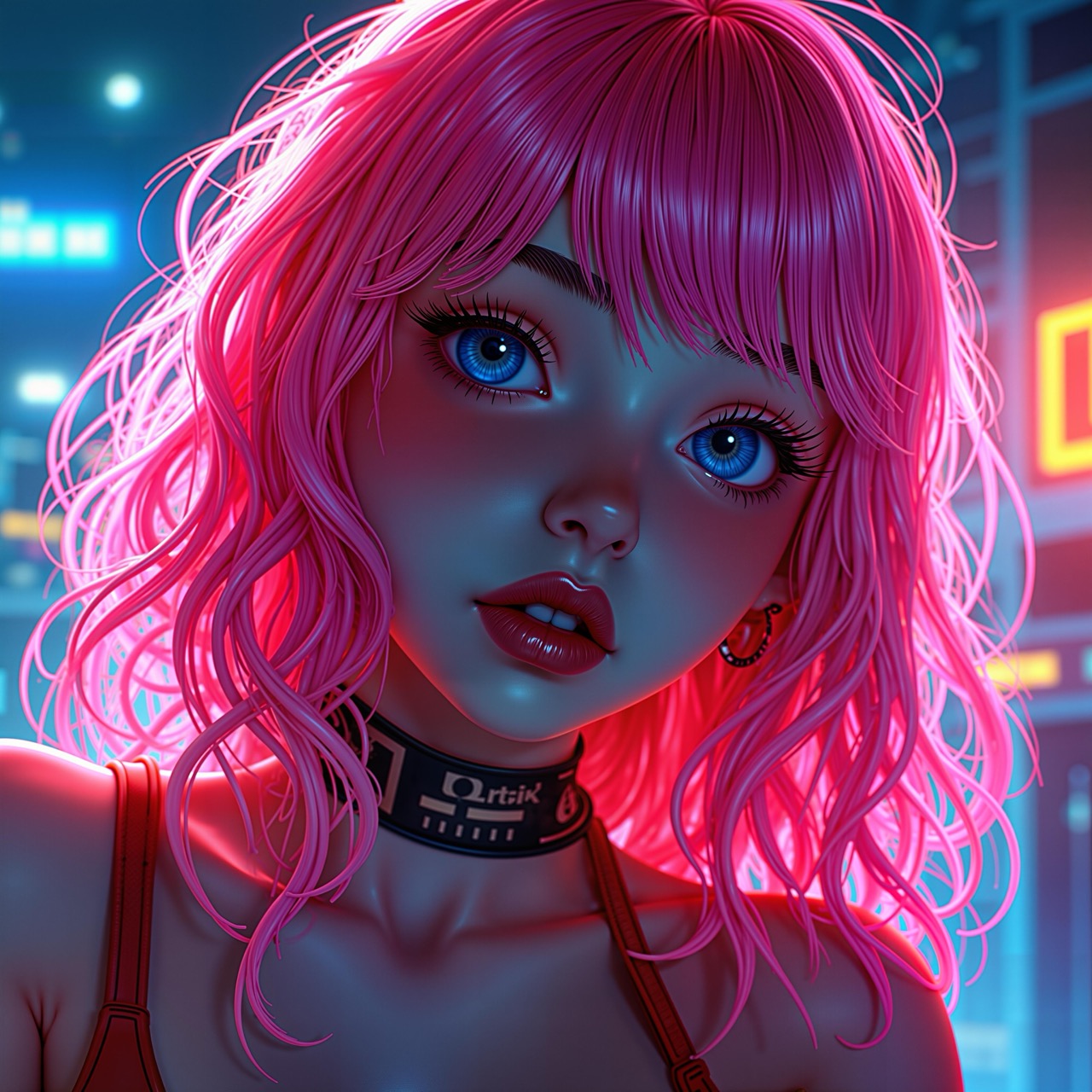}
    \includegraphics[width=0.49\linewidth]{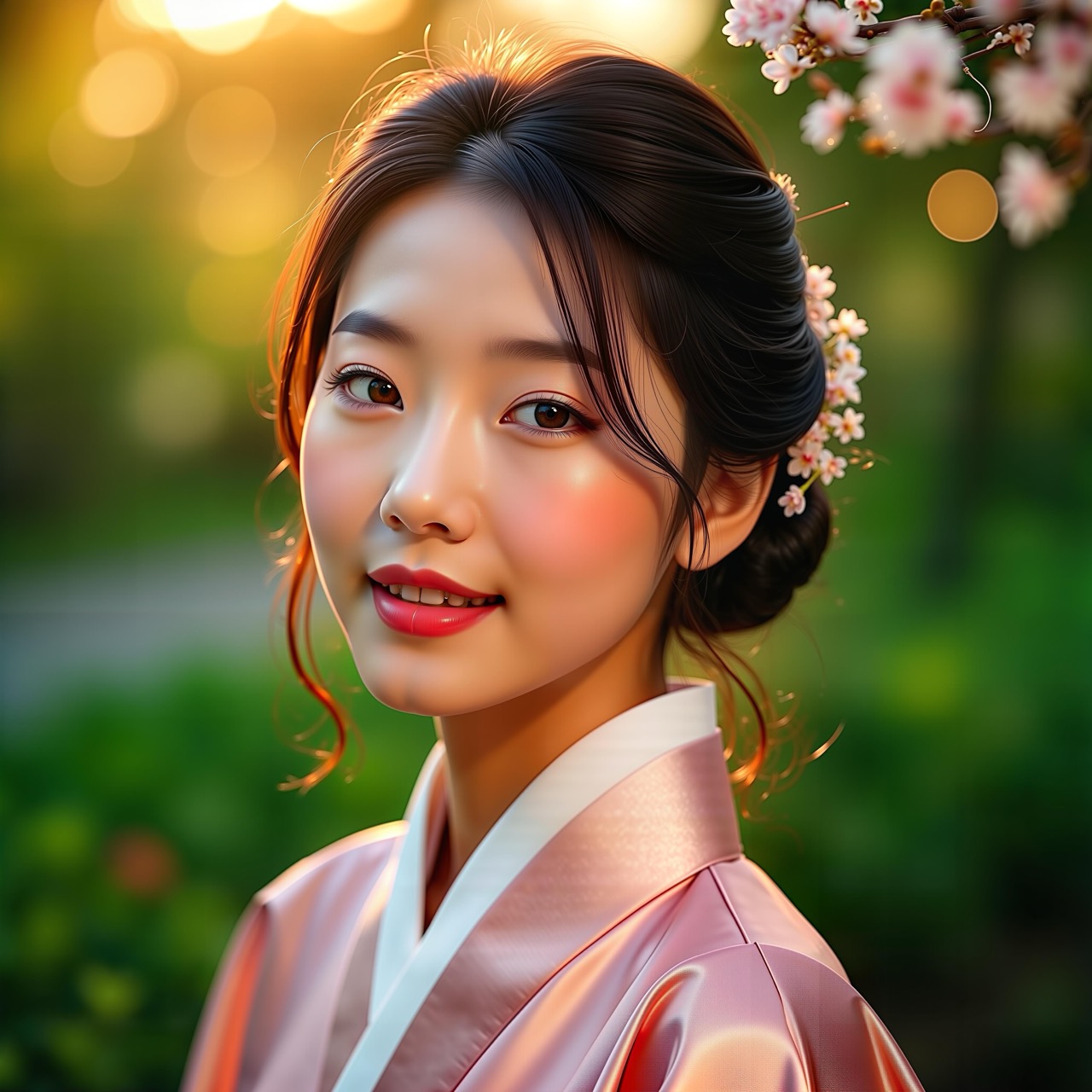}
    \caption{More 4K results}
    \label{fig:placeholder-2}
\end{figure}

\begin{figure}[htbp]
    \centering
    \includegraphics[width=0.24\linewidth]{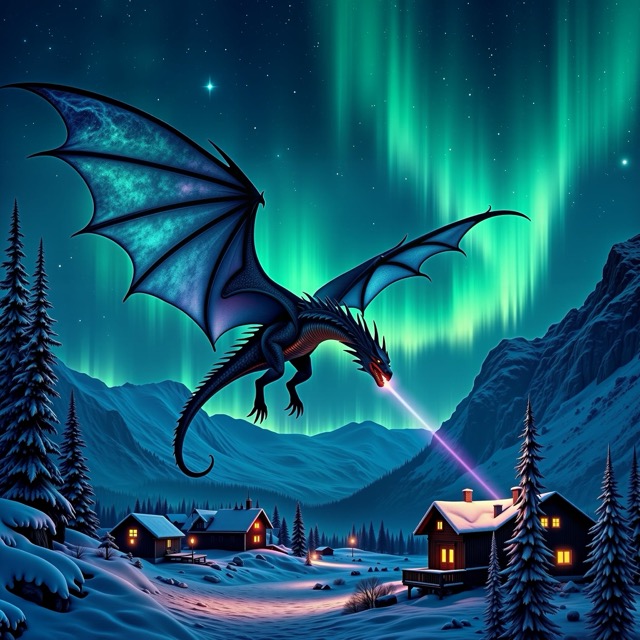}
    \includegraphics[width=0.24\linewidth]{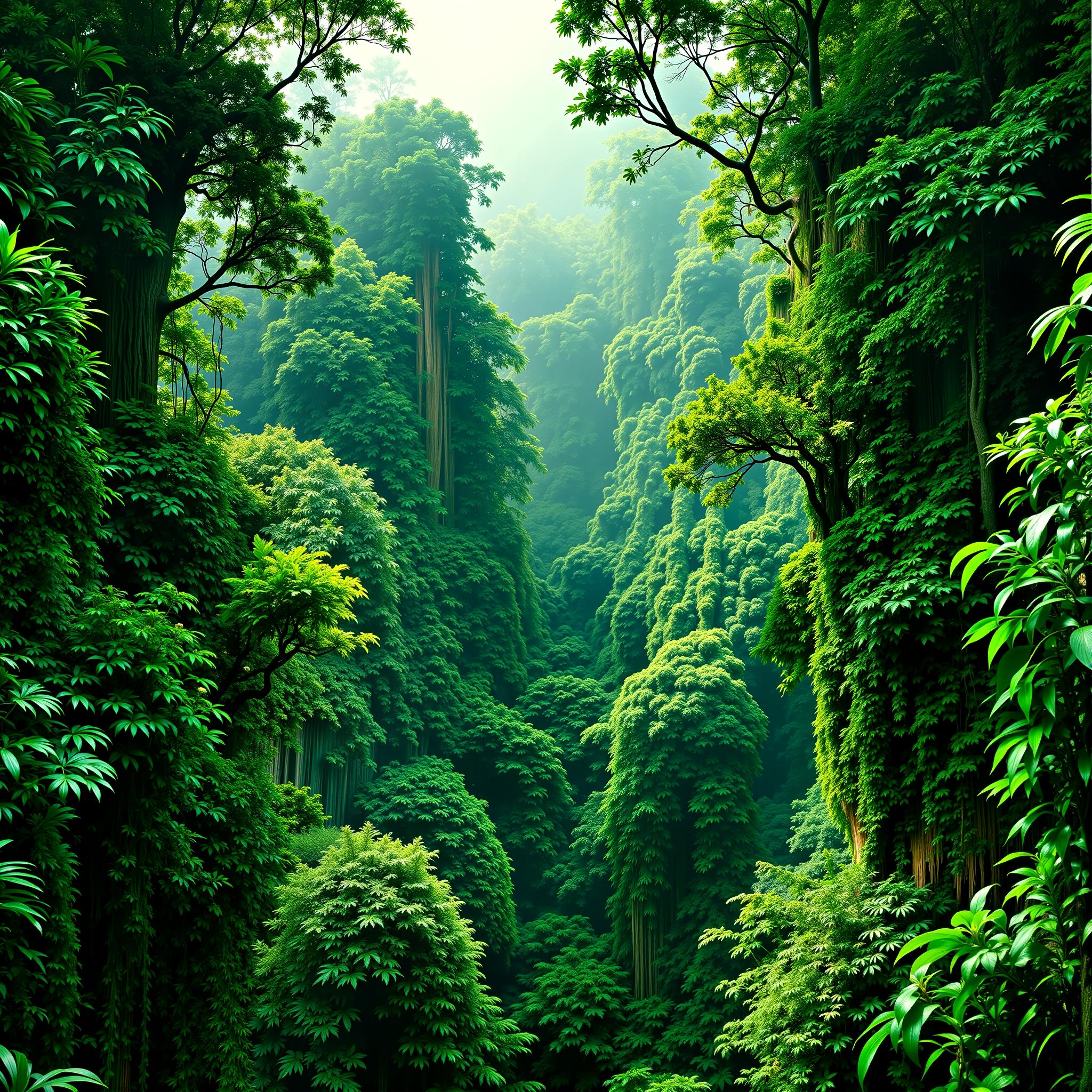}
    \includegraphics[width=0.24\linewidth]{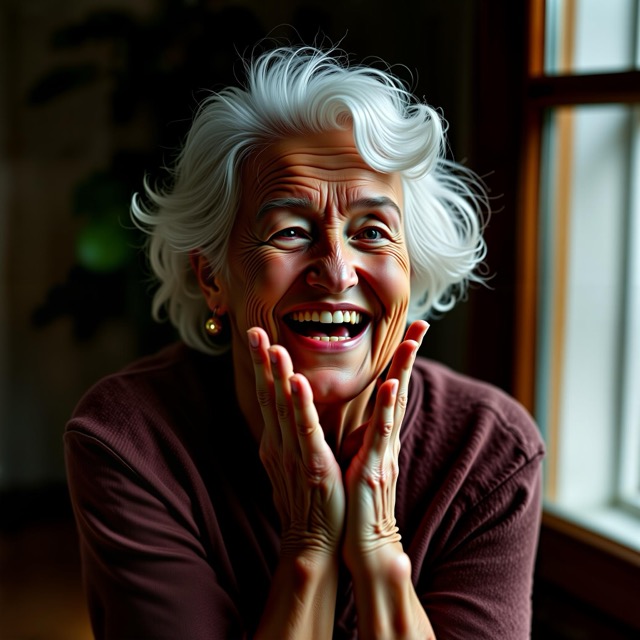}
    \includegraphics[width=0.24\linewidth]{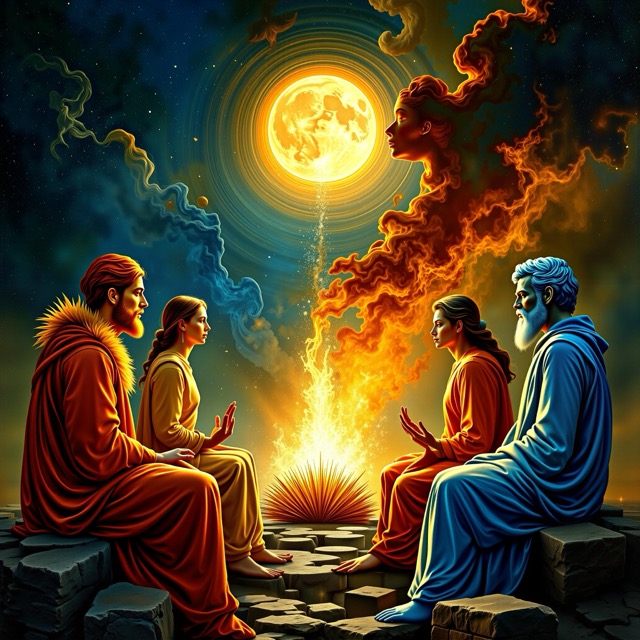}

    \includegraphics[width=0.24\linewidth]{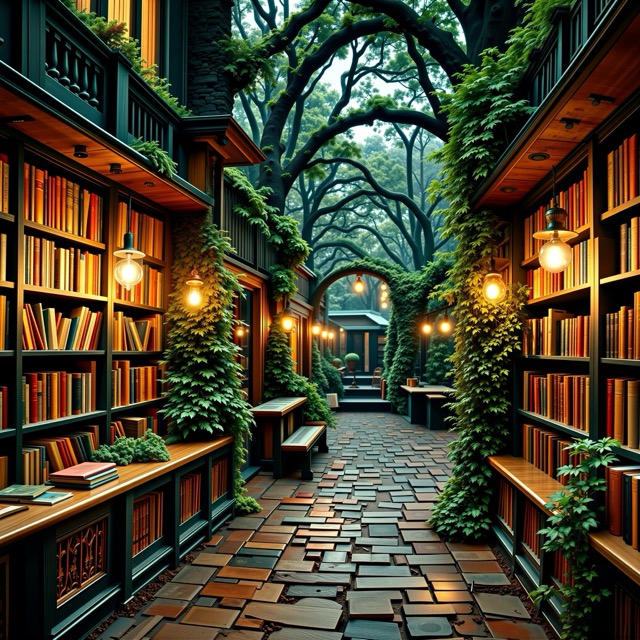}
    \includegraphics[width=0.24\linewidth]{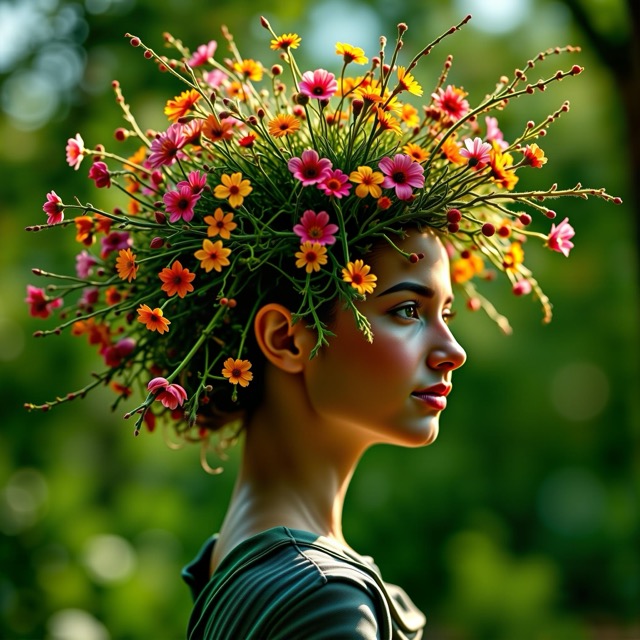}
    \includegraphics[width=0.24\linewidth]{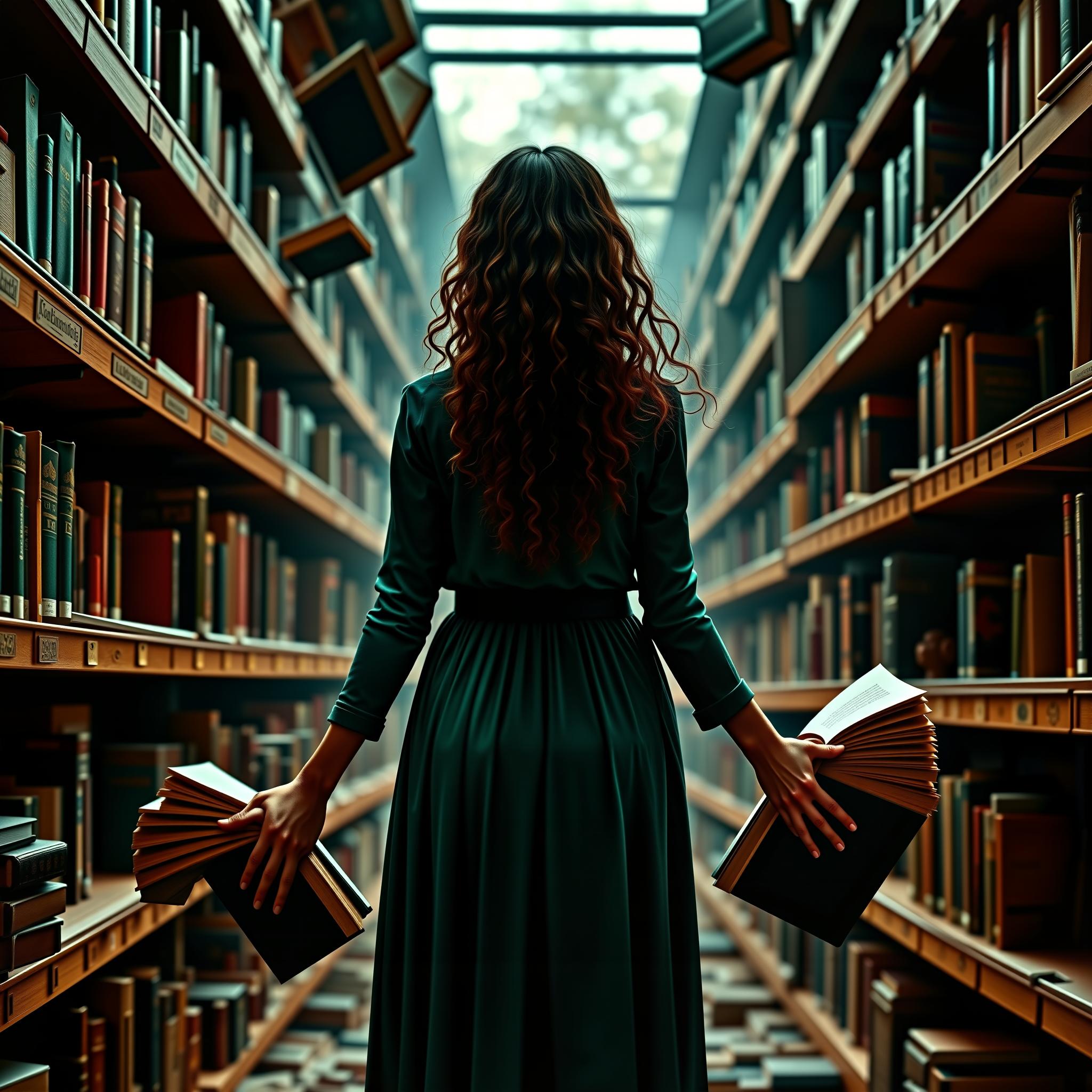}
    \includegraphics[width=0.24\linewidth]{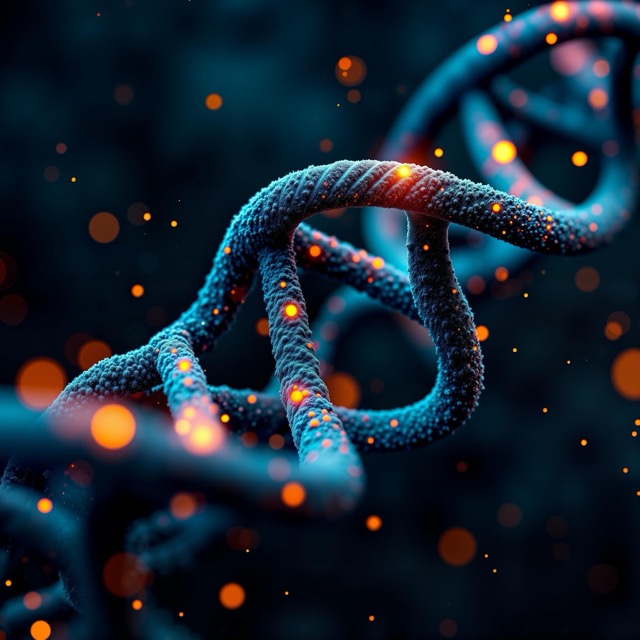}

    \includegraphics[width=0.24\linewidth]{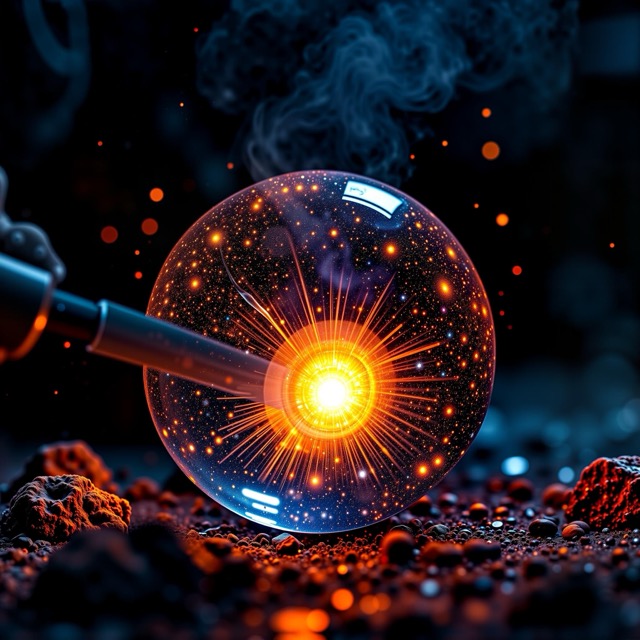}
    \includegraphics[width=0.24\linewidth]{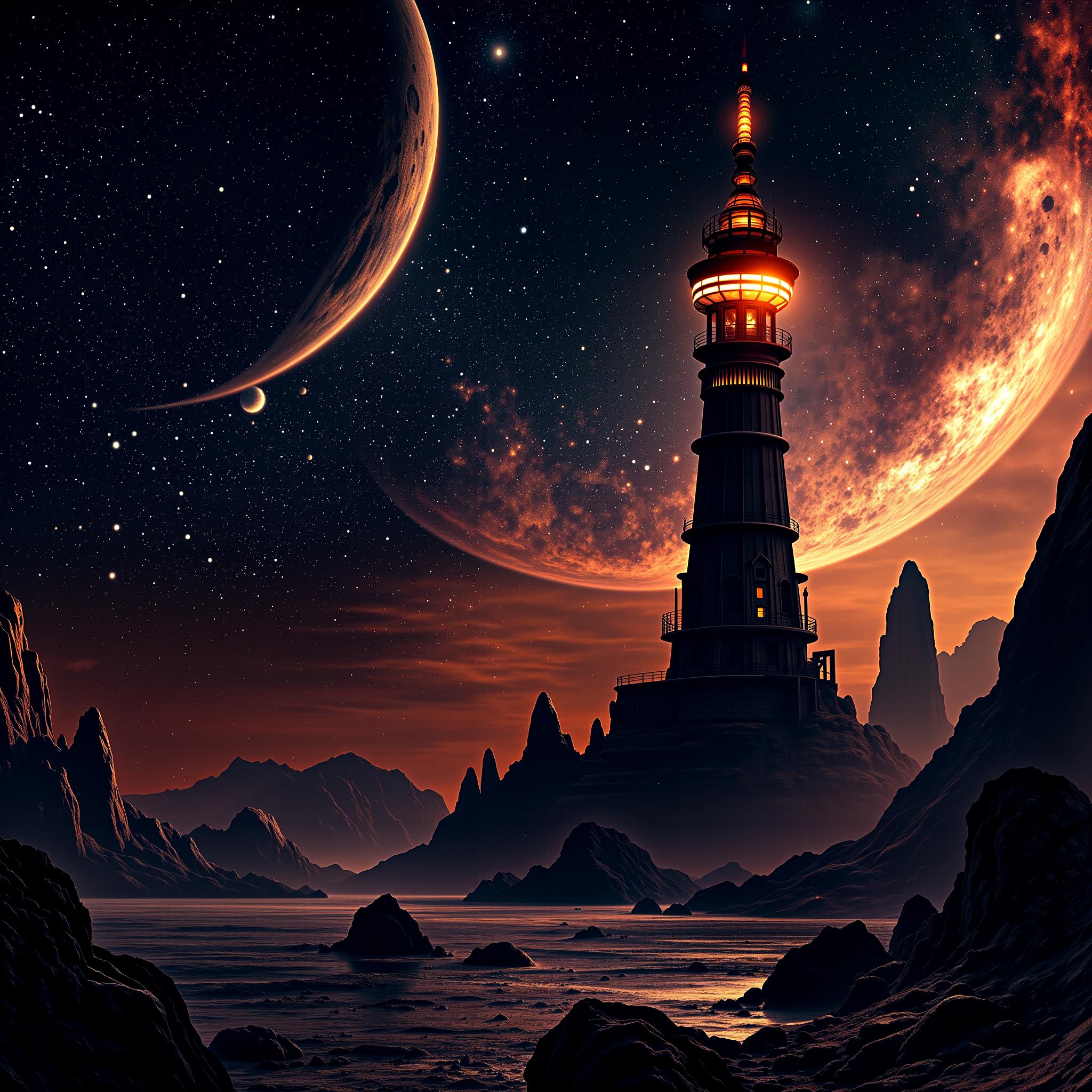}
    \includegraphics[width=0.24\linewidth]{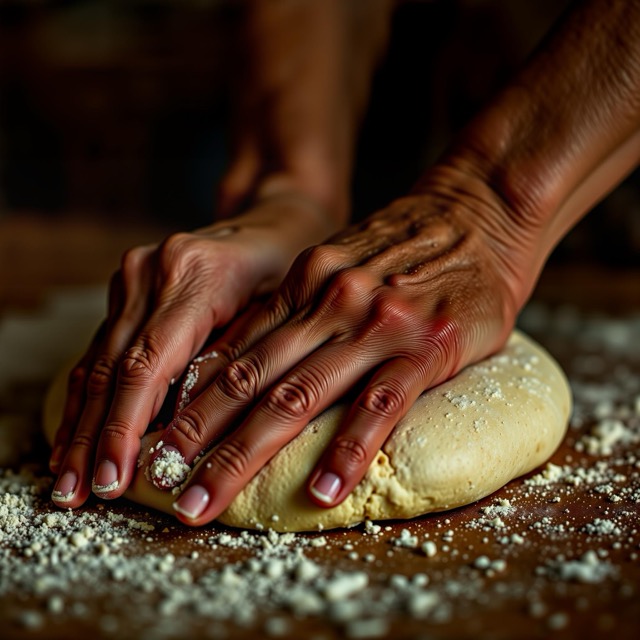}
    \includegraphics[width=0.24\linewidth]{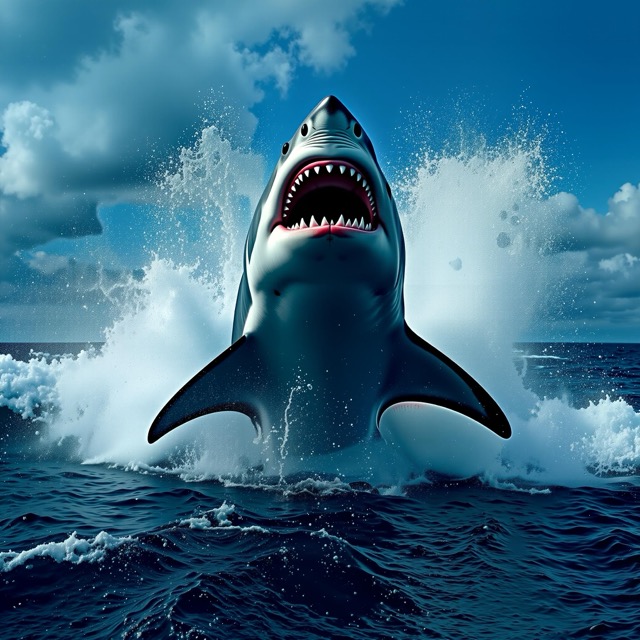}

    \includegraphics[width=0.24\linewidth]{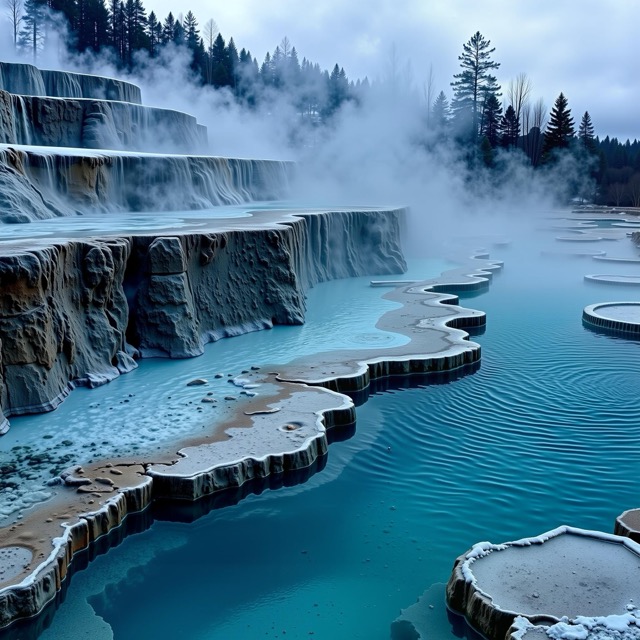}
    \includegraphics[width=0.24\linewidth]{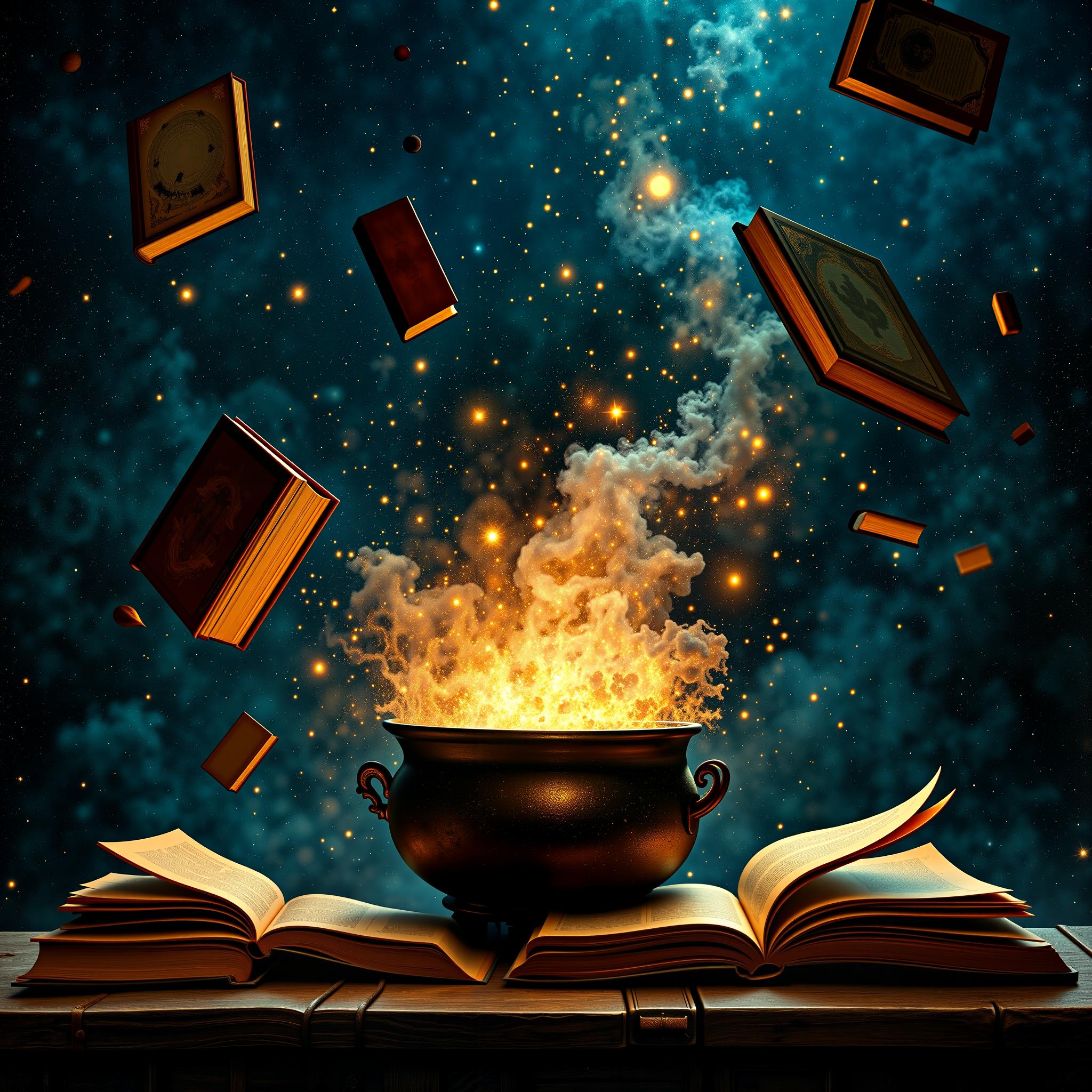}
    \includegraphics[width=0.24\linewidth]{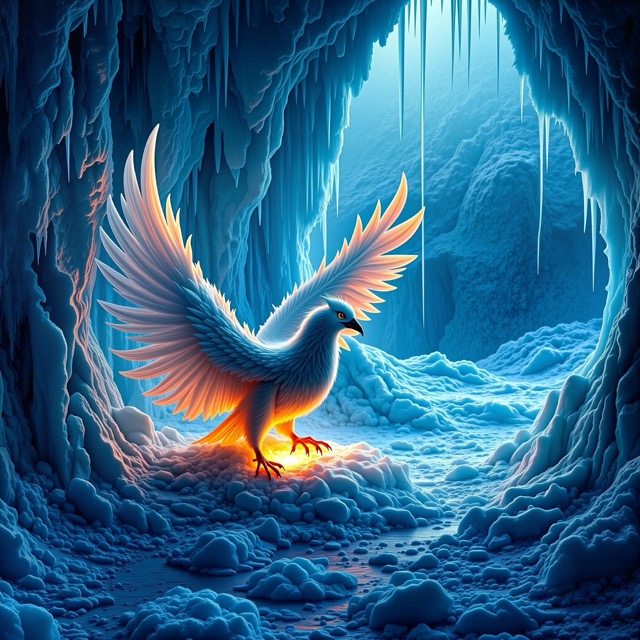}
    \includegraphics[width=0.24\linewidth]{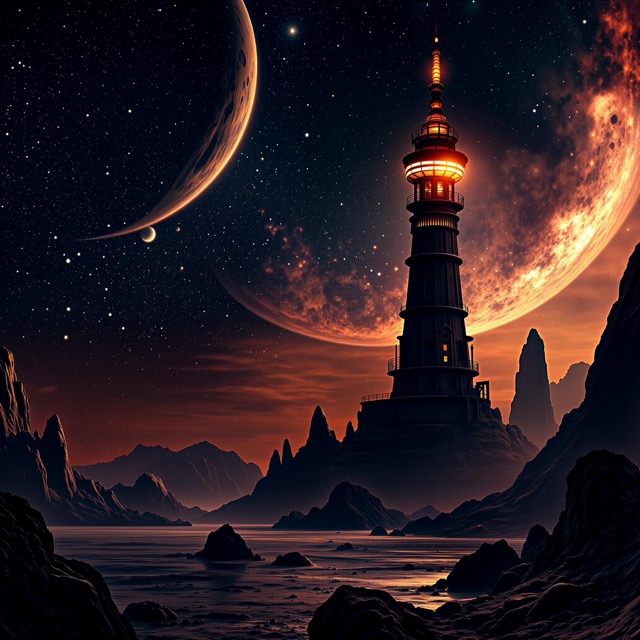}

    \includegraphics[width=0.24\linewidth]{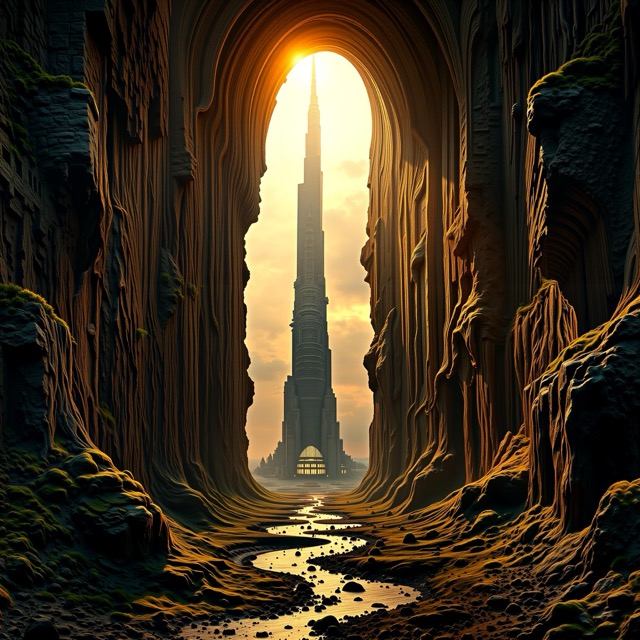}
    \includegraphics[width=0.24\linewidth]{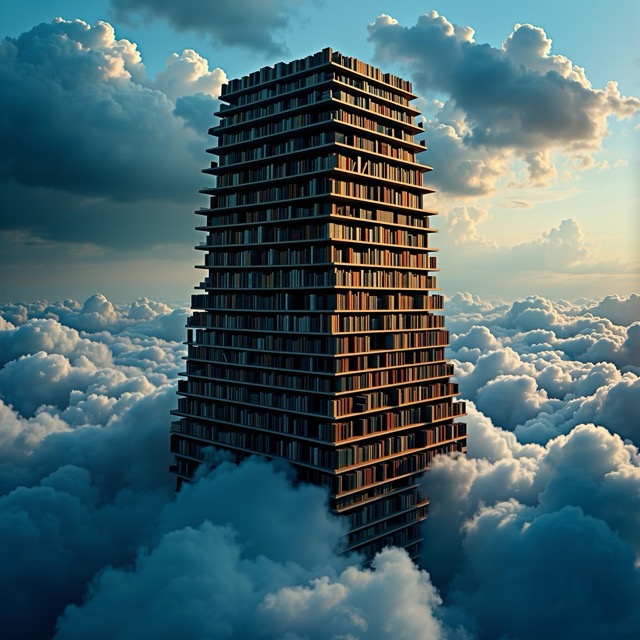}
    \includegraphics[width=0.24\linewidth]{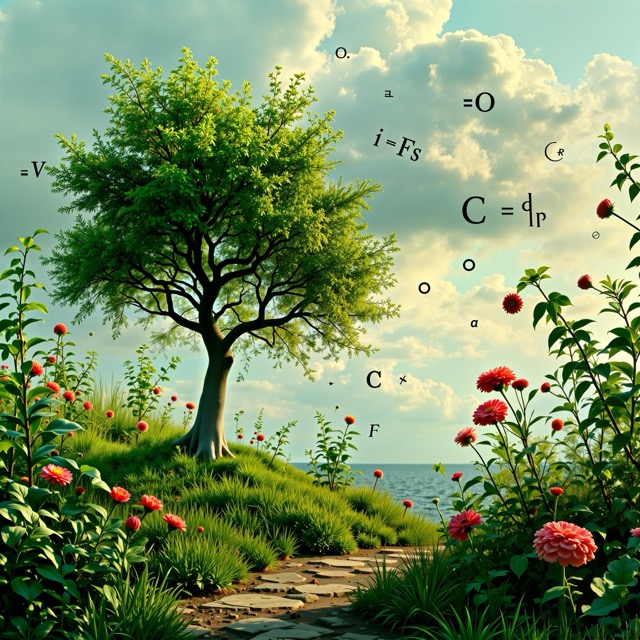}
    \includegraphics[width=0.24\linewidth]{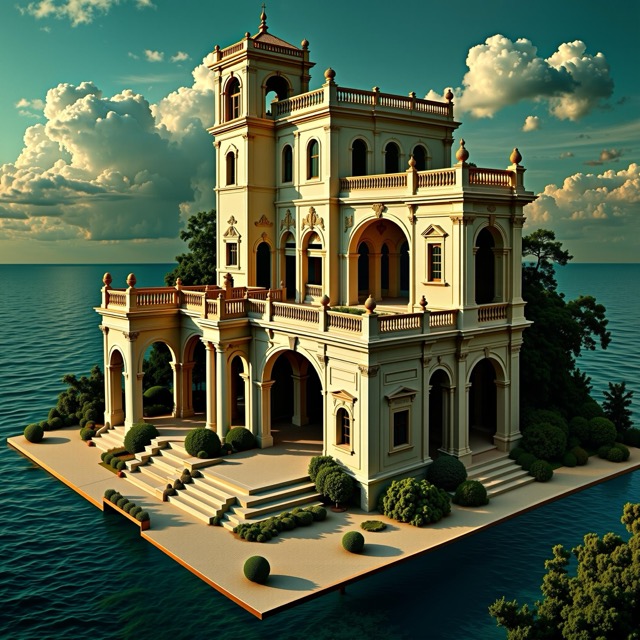}

    \includegraphics[width=0.24\linewidth]{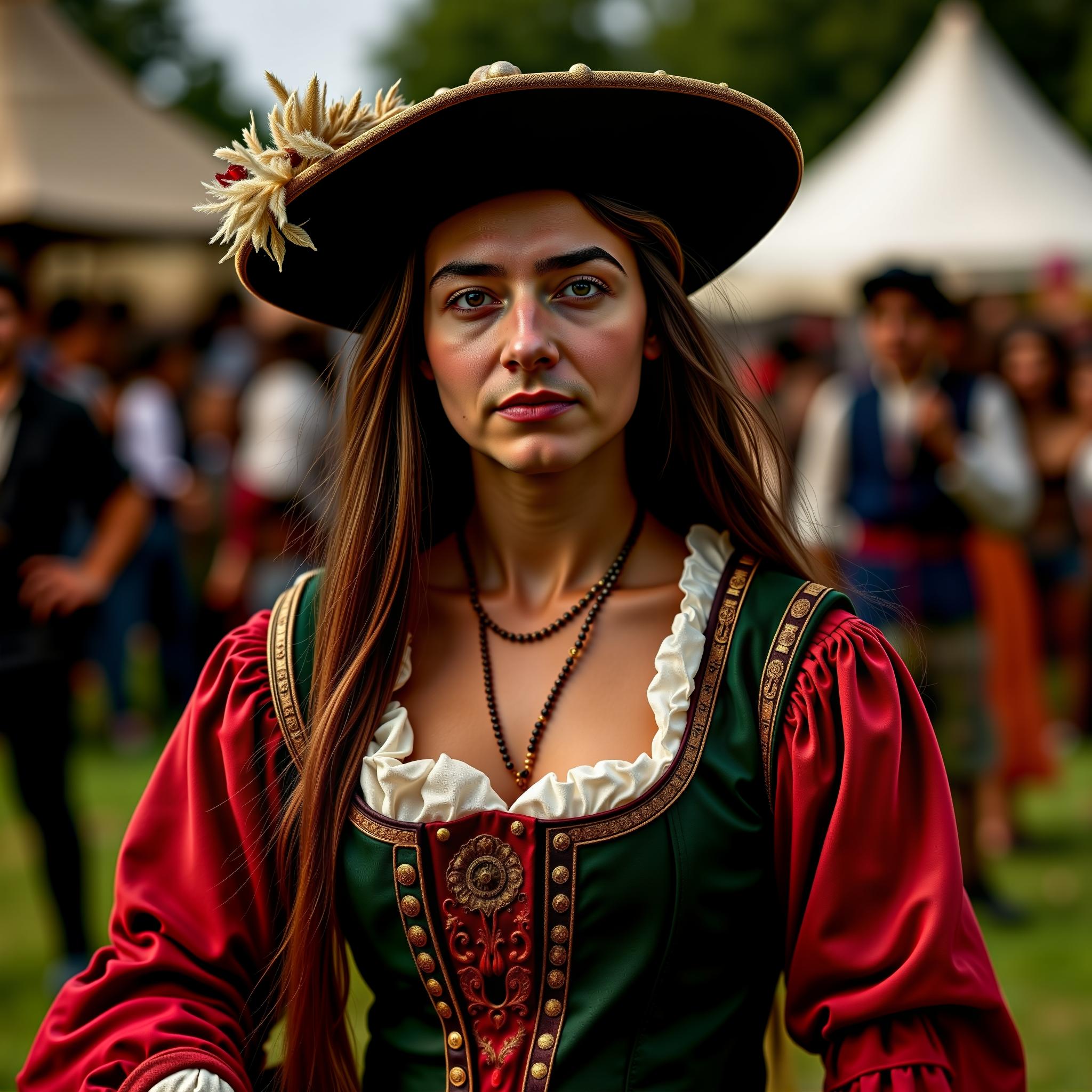}
    \includegraphics[width=0.24\linewidth]{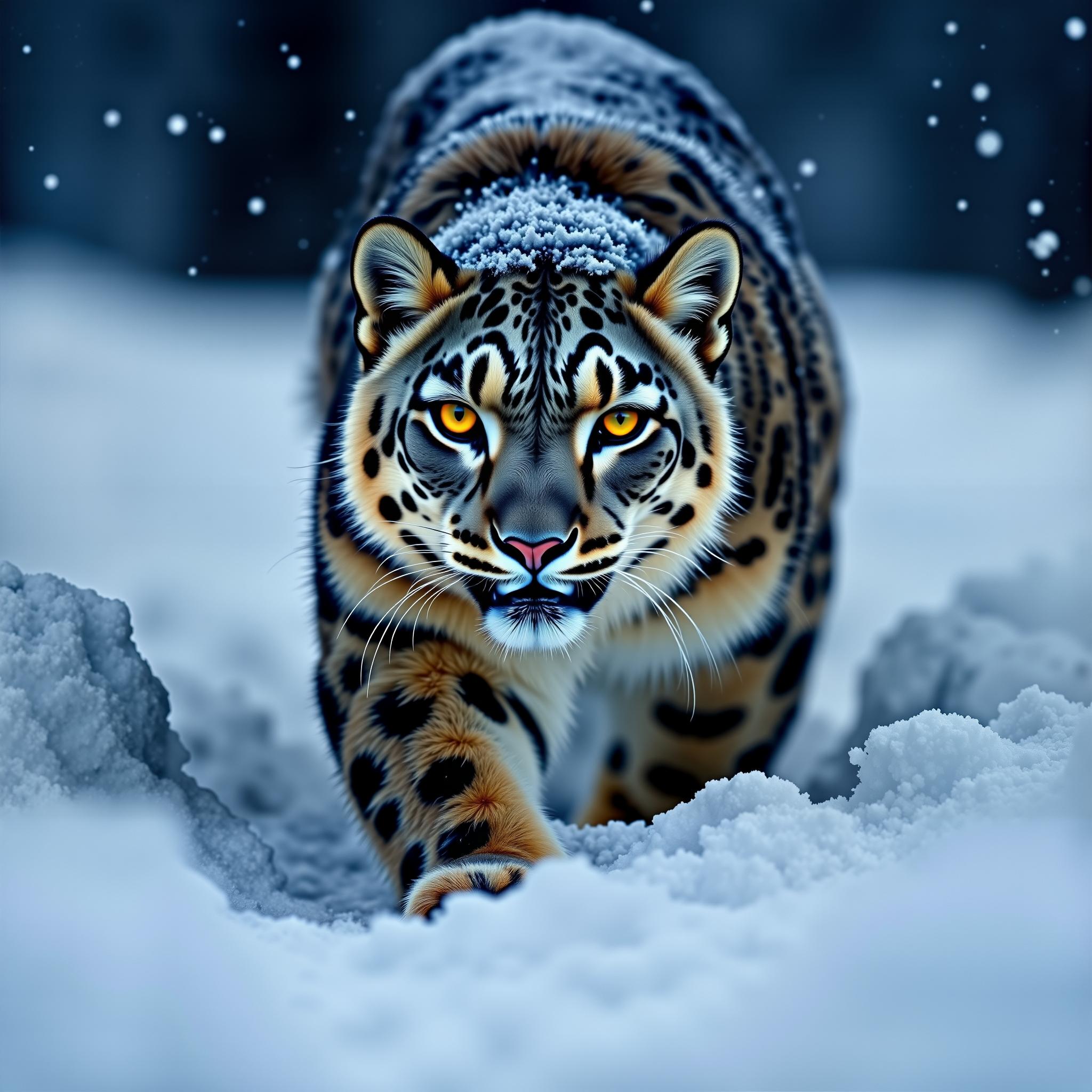} 
    \includegraphics[width=0.24\linewidth]{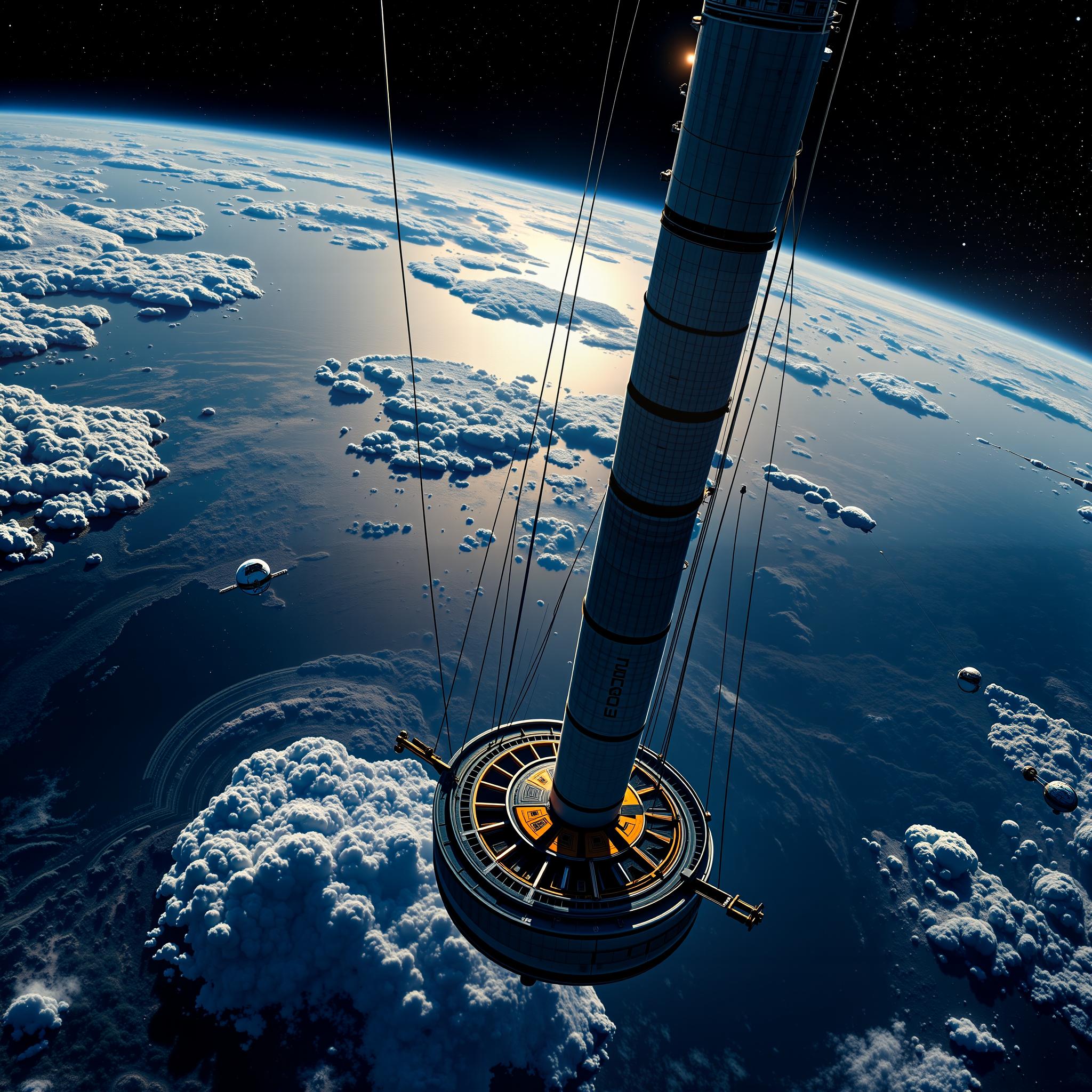}
    \includegraphics[width=0.24\linewidth]{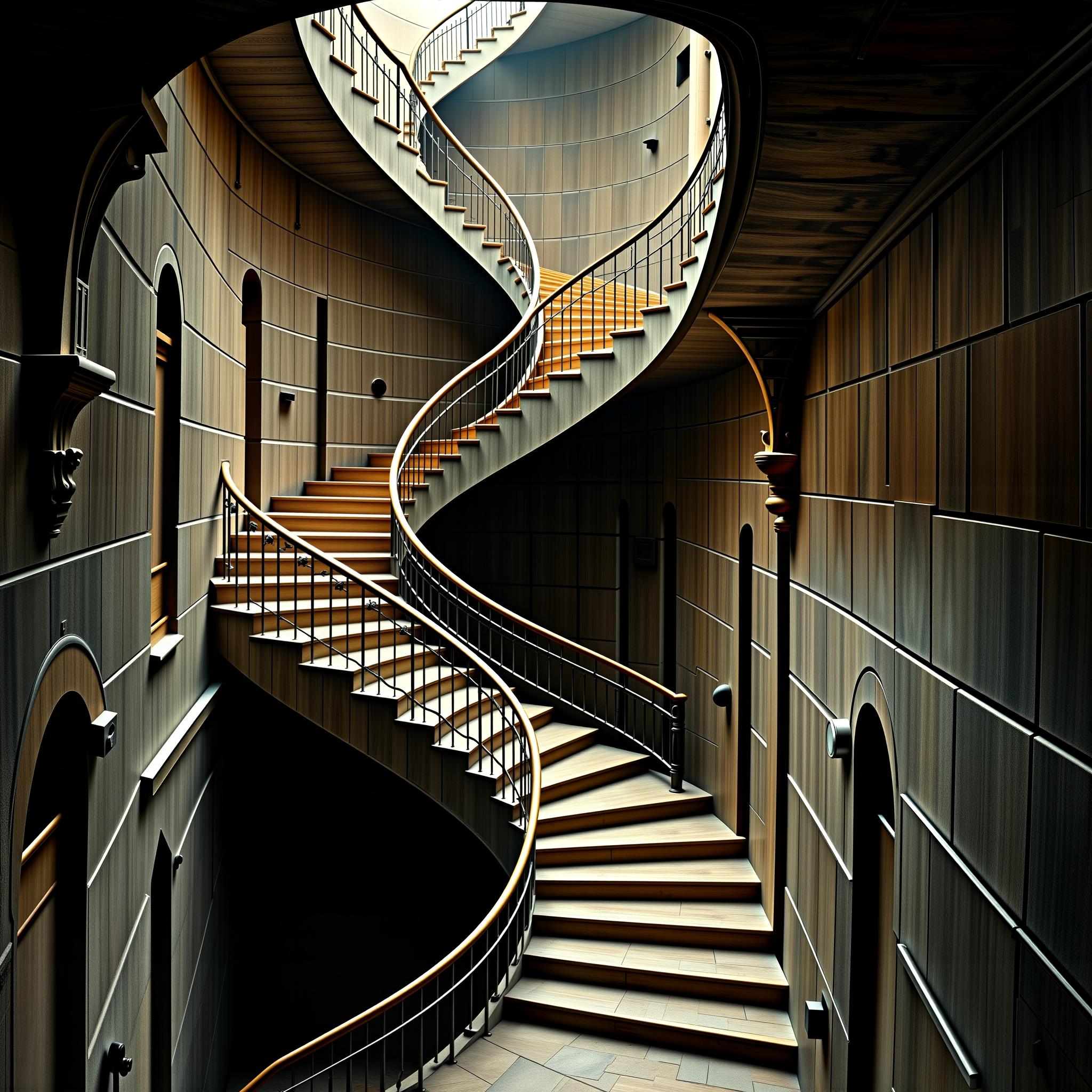}

    \caption{More 2K results.}
    \label{fig:2kgallery-part2}
\end{figure}

In Figure~\ref{fig:placeholder-2},~\ref{fig:placeholder-5},~\ref{fig:placeholder-6} and Figure~\ref{fig:2kgallery-part2} we present more 4K and 2K resolution results.  We also provide the PDF for the Teasor figure.

\begin{figure}[htbp]
    \centering
    \includegraphics[width=\linewidth]{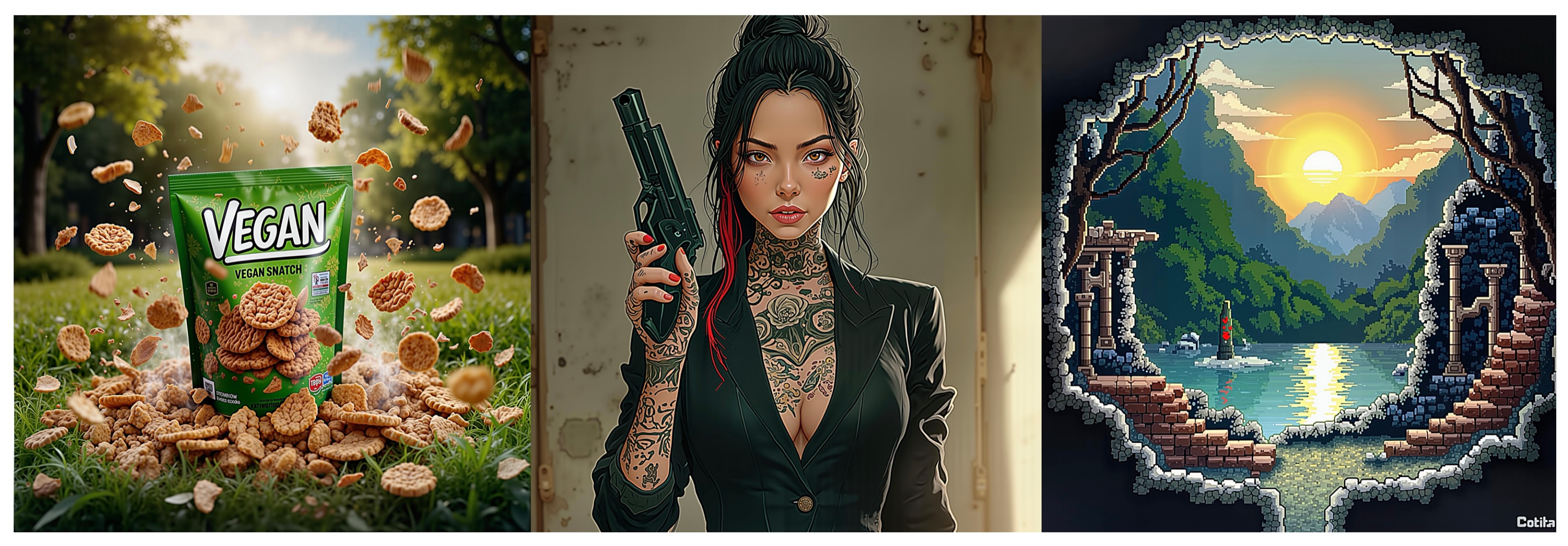}
    \caption{4K results with styled LoRAs, zoom in to see the details.}
    \label{fig:style-lora}
\end{figure}

\section{Complete Qualitative Comparison}
\begin{figure}[htbp]
    \centering
    \includegraphics[width=\linewidth]{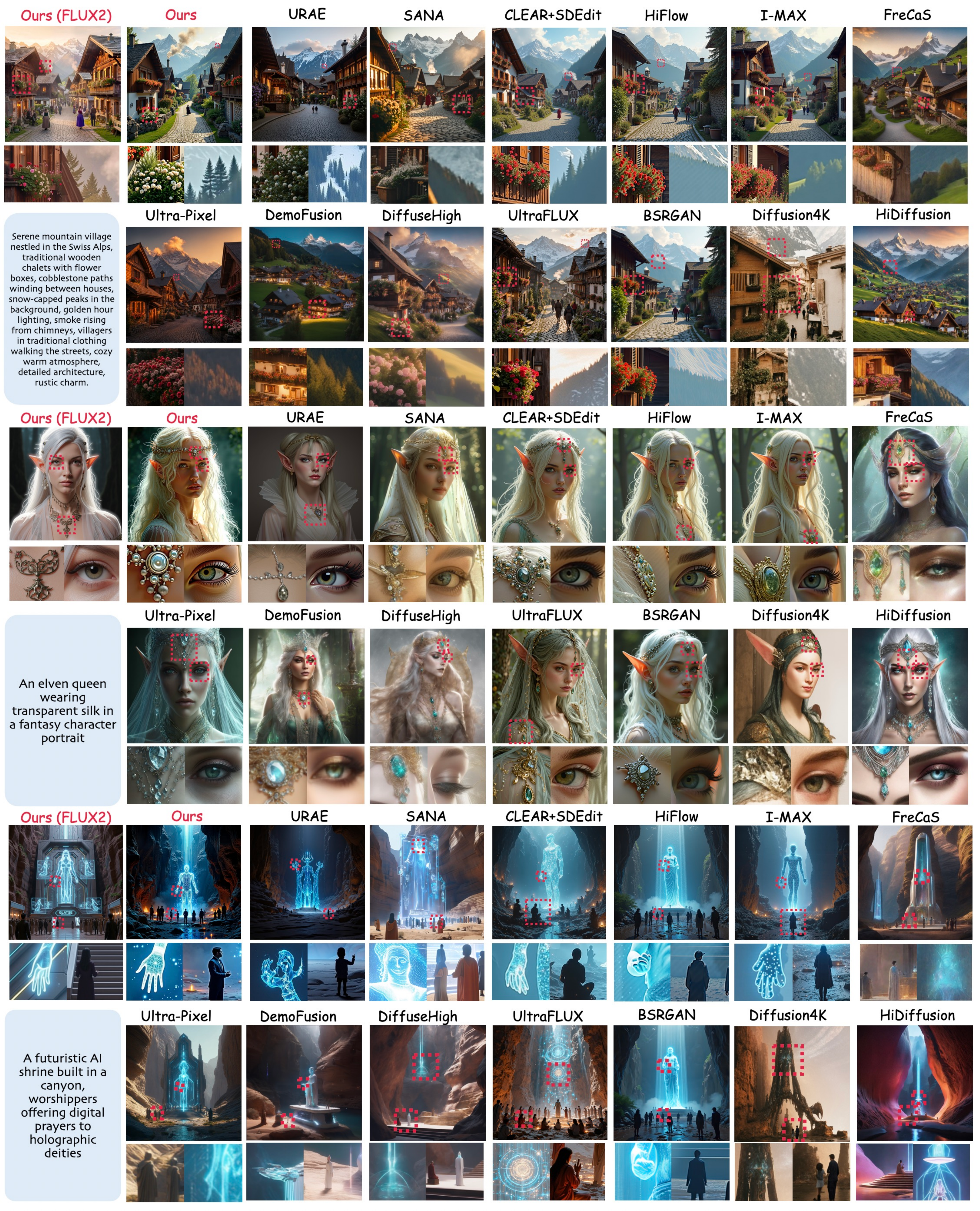}\\
    \vspace{-0.1in}
    \caption{Complete comparison against all baselines on 4K resolution. Zoom in to view the details, our method produces the best results.}
    \label{fig:appendix-qualitative}
\end{figure}

In figure~\ref{fig:appendix-qualitative}, we present the complete qualitative comparison of 4K resolution images with the quantitatively compared baselines in the main paper. We provide the PDF in the Zip file.


\section{Results with Different Styled LoRAs}
Our model is compatible with other styled LoRAs, Figure~\ref{fig:style-lora} present 4K results from 3 different LoRAs, specifically "XLabs-AI/flux-RealismLora", \\"prithivMLmods/Flux-Product-Ad-Backdrop", \\and 
"UmeAiRT/FLUX.1-dev-LoRA-Modern\_Pixel\_art".

\section{Additional Analysis and Ablations}
\subsection{LoRA training as self-distillation.}
\yuyao{Our LoRA adaptation is trained using images generated by the base model itself rather than native high-resolution data. The goal of this stage is not to learn a fundamentally new 4K high-frequency distribution, but to adapt the pretrained model to the modified attention process introduced by UltraImageGen. This self-distillation strategy keeps the adaptation data consistent with the original model distribution and preserves the same text-image alignment behavior, while allowing the model to efficiently learn how to use low-resolution anchors and local high-resolution windows during joint denoising.}

\subsection{High-frequency statistical analysis.}
\begin{figure}[h]
    \centering
    \includegraphics[width=\linewidth]{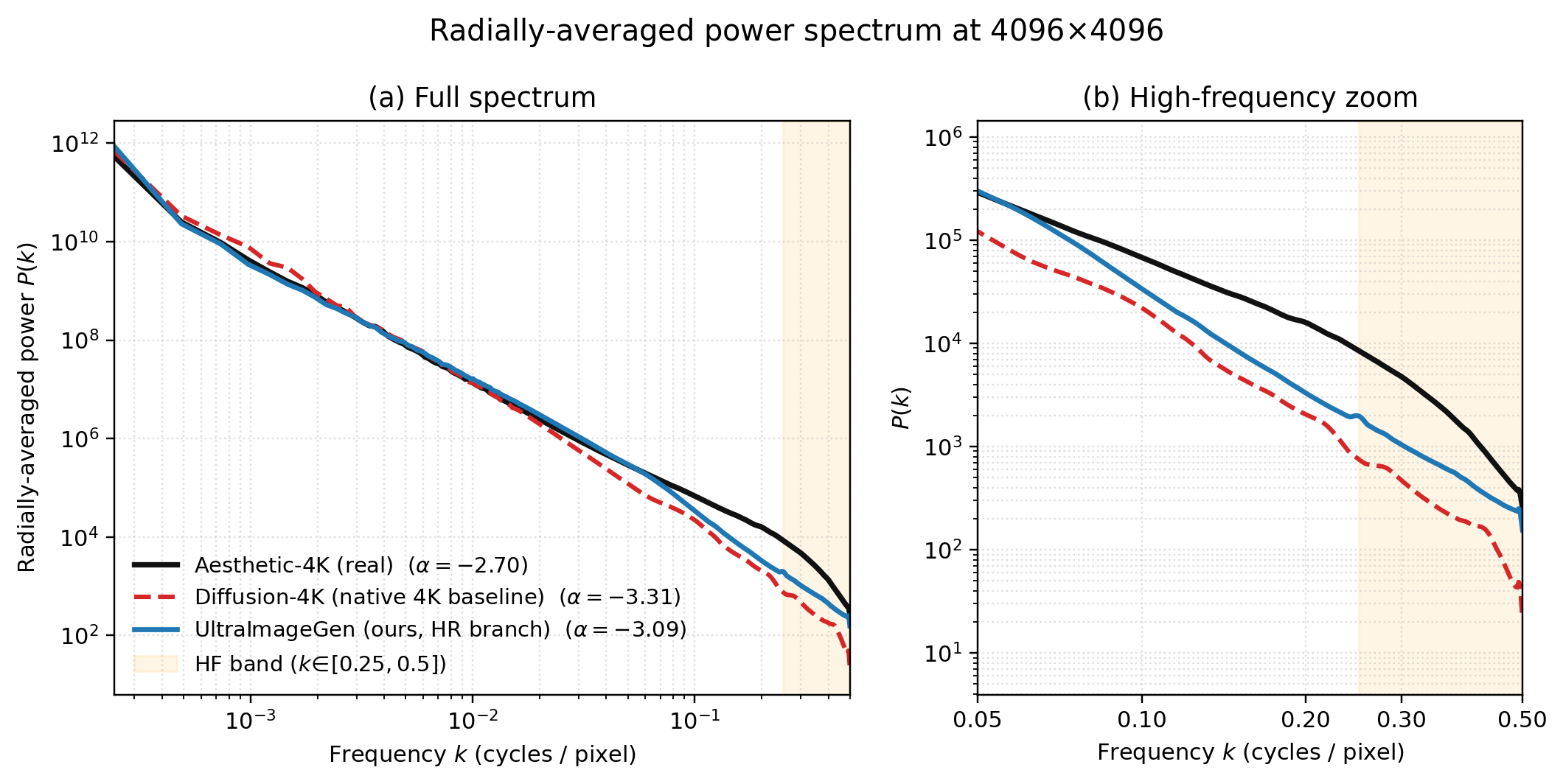}
    \caption{
    Radially averaged power spectra comparison at $4K$ resolution.
    We compare UltraImageGen with real Aesthetic-4K images and the native $4K$ baseline Diffusion-4K.
    UltraImageGen better matches the high-frequency spectrum of real images, suggesting more realistic fine-scale image statistics.
    }
    \label{fig:power_spectrum}
\end{figure}

\yuyao{To further evaluate whether UltraImageGen produces realistic fine-scale details, we compare the radially averaged power spectra of our 4K outputs with real Aesthetic-4K images and the native 4K baseline Diffusion-4K. As shown in Figure~\ref{fig:power_spectrum}, UltraImageGen more closely matches the high-frequency spectrum of real images, with a power-law slope and high-frequency energy profile closer to the real-image distribution. This suggests that our method does not merely upsample low-resolution structure, but also preserves realistic high-frequency image statistics.}

\subsection{Robustness to weaker backbones and extreme aspect ratios.}
\begin{figure}[h]
    \centering
    \includegraphics[width=\linewidth]{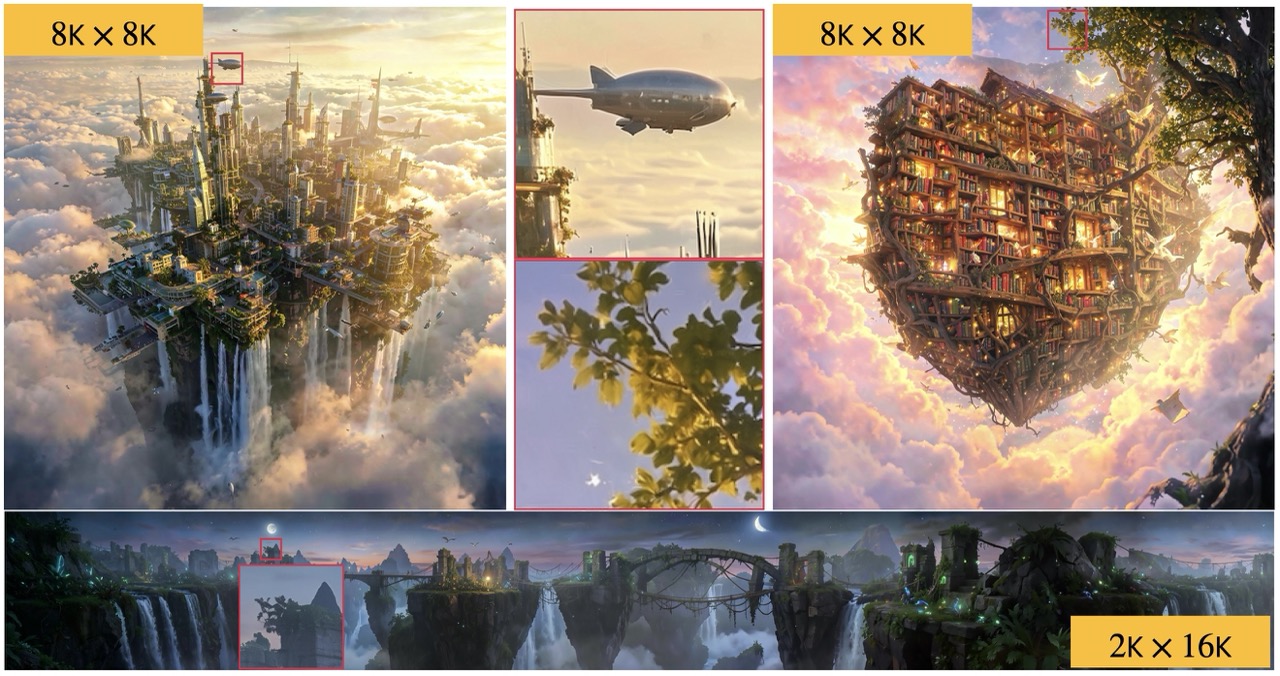}
    \caption{
    Robustness of UltraImageGen under challenging settings.
    The figure shows results on a weaker FLUX.2-Klein-4B pretrained backbone and an extreme $2048 \times 16384$ aspect ratio.
    UltraImageGen remains effective in both cases, indicating that the proposed resolution-agnostic attention formulation is not tied to a specific backbone size or square output resolution.
    }
    \label{fig:robustness}
\end{figure}

\yuyao{We further test UltraImageGen under two challenging settings: a weaker FLUX.2-Klein-4B backbone and an extreme $2048 \times 16384$ aspect ratio. As shown in Figure~\ref{fig:robustness}, UltraImageGen remains effective with the smaller pretrained model and continues to produce coherent ultra-high-resolution outputs under highly elongated aspect ratios. These results indicate that the proposed resolution-agnostic attention formulation is not tied to a specific backbone size or square-resolution setting.}
\subsection{LoRA rank ablation.}
\yuyao{We also ablate the LoRA rank used for adapting the query, key, and value projections. Table~\ref{tab:lora_rank} reports the FID scores after 10K training steps. Rank 4 underfits the modified attention behavior, while ranks 8, 16, and 64 achieve similar performance. We therefore choose rank 16 as the default setting, as it provides a good tradeoff between quality and parameter efficiency.}

\begin{table}[h]
\centering
\caption{Ablation on LoRA rank. We report FID at 10K training steps. Rank 16 achieves strong performance while maintaining parameter efficiency.}
\label{tab:lora_rank}
\begin{tabular}{c|cccc}
\toprule
LoRA rank & 4 & 8 & 16 & 64 \\
\midrule
FID $\downarrow$ & 82.41 & 76.39 & 76.08 & 76.15 \\
\bottomrule
\end{tabular}
\end{table}

\subsection{Ablation Study on Step Skipping.} 
We conduct an ablation study to investigate the optimal number of skippable denoising steps, seeking a favorable trade-off between reconstruction quality and inference speed. Specifically, we evaluate configurations in which 18, 14, 10, 8, and 4 steps are skipped during the denoising process at 4K resolution. Peak Signal-to-Noise Ratio (PSNR) is adopted as the quantitative metric, with the full-step (no skipping) output serving as the ground-truth reference. The results are summarized in Table~\ref{tab:ablation_skip}. In practice, we also find that by skipping 10 steps, we can have the most natural results as well as comparable sharpness and if too many steps are skipped like $>14$ steps, the denoising will lean towards oversmooth results as the low-resolution images. 
\begin{table}[h]
    \centering
    \vspace{-0.1in}
    \caption{Ablation study on the number of skipped denoising steps at 4K resolution. PSNR is computed against the full-step (no skipping) output.}
    \label{tab:ablation_skip}
    \begin{tabular}{lcccccc}
        \toprule
        Skipped Steps & 0 (Ref.) & 4 & 8 & 10 & 14 & 18 \\
        \midrule
        PSNR (dB) $\uparrow$ & -- & 28.78 & 28.74 & 28.70 & 27.43 & 26.43 \\
        Speedup & 1.0$\times$ & 1.17$\times$ & 1.4$\times$ & 1.56$\times$ & 2.0$\times$ & 2.8$\times$ \\
        \bottomrule
    \end{tabular}
\end{table}

\newpage

\section{Discussion on Advantages of Window Permuted Local Attention}
\subsection{Quality Preservation Across Resolutions}

A critical advantage of our approach is that \textbf{representational quality remains constant} regardless of output resolution. Since every local window operates on the same relative position range $\mathcal{R}_{\text{local}} = \{(\Delta i, \Delta j) : |\Delta i|, |\Delta j| \leq w-1\}$, and this range falls entirely within the pretrained distribution $\mathcal{D}_{\text{train}}$, each window achieves identical quality.
For an image of size $H \times W$ with local windows of size $w \times w$, and $s$ is the VAE downsampling factor
The per-unit quality across the entire image is:

\begin{equation}
\text{Quality per unit area} = \frac{Q(\mathcal{R}_{\text{local}})}{w^2 \times s^2} = \text{constant}
\end{equation}

This theoretical guarantee means that a 4K image generated by our method has the same quality as a 1K image, unlike approaches that suffer quality degradation when extrapolating beyond training distributions. The quality is \textbf{resolution-invariant} because the fundamental building blocks (local spatial relationships) remain within the learned parameter space.

\subsection{Numerical Stability Through Bounded Attention}

Local window attention maintains \textbf{numerical stability} by operating within bounded attention dimensions. While full attention at high resolutions requires increasingly high precision, our approach maintains constant precision requirements.

The information content per attention weight scales as:
\begin{itemize}
    \item \textbf{Full attention}: $-\log\left(\frac{s^2}{HW}\right) = \log(HW) - 2\log(s) \to \infty$ as resolution increases
    \item \textbf{Local attention}: $-\log\left(\frac{1}{w^2}\right) = 2\log(w) = \text{constant}$ regardless of resolution
\end{itemize}

This fundamental difference means that local attention avoids the \textbf{attention dilution problem} where each query token must distribute probability mass across tens of thousands of key tokens, leading to numerical precision issues and gradient vanishing. Instead, each query attends to only $w^2$ tokens, maintaining stable softmax distributions and reliable gradient flow.

The bounded nature of local attention ensures that the method remains \textbf{numerically robust} at any resolution, while full attention systems become increasingly unstable as they approach hardware precision limits.

\section{LLM Usage Declarations and Reproducibility statement}
We declare that Large Language Models (LLMs) were used in limited capacity during the preparation of this manuscript. Specifically, LLMs were employed for: (i) generating diverse text prompts for model evaluation and dataset creation, (ii) grammar checking and language refinement of the manuscript, and (iii) assisting in data collection procedures for experimental validation. All core technical contributions, experimental design, analysis, and conclusions presented in this work are entirely ours own. The use of LLMs did not influence the scientific methodology, results interpretation, or theoretical contributions of this research. We will release our source code upon acceptance.
\end{document}